%% file: gecco.tex
\documentclass[sigconf,authorversion]{acmart}  
\usepackage{textcomp}
\usepackage{stfloats}
\usepackage{url}
\usepackage{verbatim}
\usepackage{graphicx}

\input{1_files_boiler.tex}
\usepackage{microtype}
\definecolor{applegreen}{rgb}{0.55, 0.71, 0.0}
\definecolor{darkgreen}{rgb}{0.0, 0.5, 0.0}
\AtBeginDocument{%
  \providecommand\BibTeX{{%
    \normalfont B\kern-0.5em{\scshape i\kern-0.25em b}\kern-0.8em\TeX}}}
\usepackage{xspace}
\newcommand{\pcgnn}{PCGNN\xspace}
\newcommand{\mario}{\textit{Mario}\xspace}
\newcommand{\maze}{\textit{Maze}\xspace}

\copyrightyear{2022}
\acmYear{2022}
\setcopyright{acmlicensed}\acmConference[GECCO '22]{Genetic and Evolutionary Computation Conference}{July 9--13, 2022}{Boston, MA, USA}
\acmBooktitle{Genetic and Evolutionary Computation Conference (GECCO '22), July 9--13, 2022, Boston, MA, USA}
\acmPrice{15.00}
\acmDOI{10.1145/3512290.3528701}
\acmISBN{978-1-4503-9237-2/22/07}

\begin{document}

\title{Procedural Content Generation using Neuroevolution and Novelty Search for Diverse Video Game Levels}

\newcommand{\wits}{
  \affiliation{%
  \institution{University of the Witwatersrand}
  \department{School of Computer Science and Applied Mathematics}
  \city{Johannesburg}
  \country{South Africa}
  }
}

\author{Michael Beukman}
\orcid{0000-0002-5468-284X}
\wits{}
\email{michael.beukman1@students.wits.ac.za}

\author{Christopher W Cleghorn}
\wits{}
\email{christopher.cleghorn@wits.ac.za}

\author{Steven James}
\wits{}
\email{steven.james@wits.ac.za}


\begin{abstract}
  Procedurally generated video game content has the potential to drastically reduce the content creation budget of game developers and large studios.
  However, adoption is hindered by limitations such as slow generation, as well as low quality and diversity of content. We introduce an evolutionary search-based approach for evolving level generators using novelty search to procedurally generate diverse levels in real time, without requiring training data or detailed domain-specific knowledge.
  We test our method on two domains, and our results show an order of magnitude speedup in generation time compared to existing methods while obtaining comparable metric scores.  We further demonstrate the ability to generalise to arbitrary-sized levels without retraining.
\end{abstract}

\ccsdesc[500]{Computing methodologies~Genetic algorithms}
\keywords{Neuroevolution, Novelty Search, Procedural Content Generation}

\maketitle

\input{0_Chapter_Introduction}
\input{0_Chapter_Background}
\input{0_Chapter_GenerationMethodology}
\input{0_Chapter_Experiments}

\input{0_Chapter_Results}
\input{0_Chapter_Discussion}

\input{0_Chapter_Conclusion}
\begin{acks}
\input{0_Chapter_Acknowledgements}
\end{acks}

\bibliographystyle{ACM-Reference-Format}
\bibliography{bib}
\pagebreak
\input{0_Chapter_Appendix.tex}
\end{document}

%% file: 1_files_boiler.tex
\usepackage{float}                  
\usepackage{url}                    
\usepackage{graphicx}               
\usepackage{booktabs}
\usepackage{multirow}
\graphicspath{{../../../src/},{../../../src/}}
\usepackage{algorithm}
\usepackage{algorithmic}

\makeatletter

\newlength{\phaserulewidth}
\newcommand{\setphaserulewidth}{\setlength{\phaserulewidth}}

\newcommand{\phase}[1]{%
  \vspace{-1.25ex}
  \STATE \hspace{-2\algorithmicindent} \hrulefill
  \STATE \hspace{-2\algorithmicindent} #1
  \vspace{-1.25ex}
  \STATE \hspace{-2\algorithmicindent} \hrulefill

  }

\makeatother

\setphaserulewidth{.7pt}

\setphaserulewidth{.7pt}

\usepackage{subcaption}
\usepackage{adjustbox}
\usepackage[english]{babel} 
\hypersetup{
  colorlinks   = true,              
  urlcolor     = black,              
  linkcolor    = black,              
  citecolor    = black                
}

\addto\extrasenglish{  
    
}
\addto\extrasenglish{  
    
}
\addto\extrasenglish{  
    
}
\addto\extrasenglish{  
    
}

\addto\extrasenglish{  
    
}
\addto\extrasenglish{  
    
}

\addto\extrasenglish{  
    
}

\newcommand{\refappendix}[1]{\hyperref[#1]{Appendix}}
\makeatletter
\newcommand*{\centerfloat}{%
  \parindent \z@
  \leftskip \z@ \@plus 1fil \@minus \textwidth
  \rightskip\leftskip
  \parfillskip \z@skip}
\makeatother

\makeatletter
\newcommand\Autoref[1]{\@first@ref#1,@}
\def\@throw@dot#1.#2@{#1}
\def\@set@refname#1{
    \edef\@tmp{\getrefbykeydefault{#1}{anchor}{}}%
    \xdef\@tmp{\expandafter\@throw@dot\@tmp.@}%
    \ltx@IfUndefined{\@tmp autorefnameplural}%
         {\def\@refname{\@nameuse{\@tmp autorefname}s}}%
         {\def\@refname{\@nameuse{\@tmp autorefnameplural}}}%
}
\def\@first@ref#1,#2{%
  \ifx#2@\autoref{#1}\let\@nextref\@gobble
  \else%
    \@set@refname{#1}
    \@refname~\ref{#1}
    \let\@nextref\@next@ref
  \fi%
  \@nextref#2%
}
\def\@next@ref#1,#2{%
   \ifx#2@ and~\ref{#1}\let\@nextref\@gobble
   \else, \ref{#1}
   \fi%
   \@nextref#2%
}

\makeatother

\makeatletter
\renewcommand*{\NAT@spacechar}{~}
\makeatother

%% file: 0_Chapter_Introduction.tex
\section{Introduction} 
\label{chap:intro} 
Video games are a massive industry, with 227 million reported video game players in the United States as of 2021~\citep{esa_2022}.
Some of the main goals of video games are to keep players entertained, engaged, and challenged~\citep{pinelle2008heuristic}. 
This can be achieved by populating the game with a large amount of unique and interesting content. 
Nonetheless, most commercial video games still rely on human designers and developers to create this content~\citep{hendrikx2013procedural}.
However, this is costly, and due to tight deadlines there is always a finite amount of content that players can quickly exhaust.
This concern can be addressed through procedural content generation (PCG), where game content is algorithmically created~\citep{togelius2011searchbased}.

There are many notable games in which content and levels are procedurally generated, such as \textit{Rogue}, where the player controls a character that traverses dungeons while fighting enemies. 
Such an approach can be used to generate a near-infinite number of levels---designers need only specify the mechanics of the game and the generation method. 
This enables game developers to create engaging and fun games at a fraction of the cost compared to traditional, manual development~\citep{hendrikx2013procedural}.

Outside of game development, PCG can also be leveraged to train machine learning agents. 

For example, \citet{using_pcg_to_train_rl} train reinforcement learning (RL) agents on procedurally generated levels to improve generalisation to unseen human generated levels, while \citet{cobbe2019quantifying} demonstrate that RL agents can overfit on surprisingly large training sets, and use a large number of procedurally generated levels to overcome this.

There are many different approaches to developing PCG algorithms. These include ad hoc algorithms (used in \textit{Rogue}),  formal languages~\citep{maung2015applying_wangs}, evolutionary search-based methods~\citep{togelius2011searchbased,togelius2013procedural_challenges_generic}, exhaustive search~\citep{sturtevant2018exhaustive}, and more recent machine learning approaches~\citep{summerville2018procedural_pcgml,liu2020deep, edrl}.
Different methods also have different goals, such as providing a consistent quality of levels, or quickly generating playable \citep{pcgrl}, diverse \citep{constrained_novelty} or configurable levels~\citep{ferreira_mario}.

Despite the variety of approaches, there are gaps in the current literature. For example, some approaches generate diverse, playable levels, but the generation process is slow~\citep{constrained_novelty, searching_for_good_and_diverse_game_levels}. Some methods can generate levels relatively quickly, but they require existing training data~\citep{CPPN2GAN,volz2018evolving} or game-specific reward engineering~\citep{pcgrl}, while others are limited by the lack of diverse content~\citep{quality_diversity_pcg,summerville2018procedural_pcgml}.
The main limitation is that no one method can quickly generate diverse levels without the need for training data and game-specific knowledge.

We address this gap by training a neural network that can quickly be queried to generate levels. We avoid the need for training data by using NeuroEvolution of Augmenting Topologies~\citep{neat} to evolve this network.
To obtain diverse levels, we explicitly reward diversity in the evolutionary process by using novelty search~\citep{novelty_search}. This method assigns fitnesses based on how far an individual is from its closest neighbours, which incentivises exploration and obtains diverse behaviours. Further, we only use widely applicable general fitness functions based on novelty and solvability to evolve these networks.

We test our methods on a simple \maze game as well as \textit{Super Mario Bros}. 
Our results indicate that our method generates levels significantly faster than both a direct search-based method and an RL-based approach, without the need for  game-specific knowledge.
We also show that our method generalises to different sized levels than those trained on, while still mostly generating solvable levels.\footnote{Source code is publicly released at \url{https://github.com/Michael-Beukman/PCGNN}.} 

%% file: 0_Chapter_Background.tex
\section{Background}
\label{chapter:background}
\label{sec:bg}

Below we outline two separate approaches that have been previously used to construct PCG systems.

\subsection{Evolutionary Search}
Genetic algorithms consist of a population of individuals, each possessing a genotype, which can be thought of as the individual's genes. This genotype impacts the phenotype---the manifestation of the genotype in the problem domain.
For example, when generating platformer levels, the genotype can be a single integer vector representing the height of platforms across the $x$-axis~\citep{ferreira_mario}. The phenotype, then, is the actual level that has been generated using this specific genotype~\citep{goldberg1989genetic}.

High performing individuals are combined using \textit{crossover}, which is the process of combining two parents to form new individuals for the next generation. This new generation is also randomly mutated to facilitate exploration and prevent stagnation.
In general, ``high performing'' is quantified by the \textit{fitness function}. For example, when maximising a function $f(x)$, the fitness could simply be the function value itself.

\subsubsection{ NeuroEvolution of Augmenting Topologies (NEAT) }
NEAT is a method where a genetic algorithm optimises the structure and weights of a neural network~\citep{neat}. 

The genetic encoding in NEAT is a linear collection of either node genes, specifying the existence and type (input, hidden or output) of a node, or connection genes, which indicate a connection of a certain weight between two nodes.
Mutation can affect both the weights and the structure of the network, either by adding a connection between two nodes, inserting a node between two existing connected nodes, or perturbing a weight's value.

NEAT keeps track of the historic origin of a gene by using the \textit{innovation number}. This enables efficient crossover between different sized parents by first lining up genes with the same history in both parents. Matching genes are then inherited randomly from each parent, whereas genes that occur in only one are inherited from the parent with higher fitness.

NEAT can enable complexity to gradually increase as the search process develops, leading to later generations being more complex than previous ones.
Finally, different variations of NEAT exist, most notably HyperNEAT~\citep{hyperneat}, which evolves compositional pattern-producing networks~\citep{cppn}---a type of artificial neural network where each node can use a different activation function---that themselves generate the final network structure. This can lead to larger and more symmetric networks and smaller genomes, although HyperNEAT does not always outperform NEAT~\citep{hyperneat_vs_neat}.

\subsubsection{Novelty Search}
\label{sec:novelty}
Novelty search~\citep{novelty_search} is a different approach to designing the fitness function of a genetic algorithm. Instead of pursuing a higher objective function value, novelty search only judges individuals based on how different or novel they are compared to the current generation and an archive of previously novel individuals. Novelty is defined as the average distance between an individual and its $k$ closest neighbours in behaviour space. Distance can be defined in a domain-agnostic manner, e.g. using a vector norm like absolute difference or Euclidean distance. Domain-specific distance functions can also be used when a more appropriate measure of distance exists.     Novelty search encourages agents to pursue novel and diverse behaviours, thereby thoroughly exploring the behaviour space and resulting in diverse individuals~\citep{gomes2013evolution, lehman2011evolving}. Even though there is no explicit incentive to actually achieve the goal, novelty search can still achieve competitive results, especially in deceptive fitness landscapes when using a traditional objective can result in convergence to local minima~\citep{novelty_search}.

\subsection{Reinforcement Learning}

An alternate approach to PCG is to use reinforcement learning (RL) to learn a policy that generates new levels~\citep{pcgrl}.
Here the problem is formulated as a Markov decision process $\langle S, A, p, r \rangle$ where $S$ is the set of states, $A$ is the set of available actions, $p$ is the transition dynamics that specify how the environment changes under a given action, and $r$ is the reward function. The aim is to learn a policy $\pi: S \to A$ such that the expected sum of future rewards is maximised~\citep{sutton1998_rlbook}.

In the context of PCG, \citet{pcgrl} uses RL to generate levels for 2D tilemap-based games, experimenting with different state and action spaces. For example, in the ``turtle'' representation, the state is the current level and the coordinate of the current tile under consideration, while $A$ consists of changing the current tile to any other one or moving the agent in one of the four cardinal directions. 
The ``wide'' representation, on the other hand, also uses the current level as the state, but actions consist of a coordinate of a tile and a value to change it to. In both cases, the agent is rewarded based on the change that its action causes to the level.

\subsection{Related Work}
Most procedural level generation approaches can be classified into one of two categories.
The first is to directly search in level space. This means that each time the method is run, a new search is carried out for only one level. The genome encoding here is usually direct. Examples of this include the work by \citet{ferreira_mario} who search over integer vectors of the same length as the level, \citet{constrained_novelty, liapis_2013c} who search over a 2D tilemap representation of the levels, and \citet{Cardamone_racing} who generate racing game tracks by using a set of control points for Bezier curves as the representation.

The second is to search in generator space. This involves searching for or learning a generator of levels. This generator can be queried to generate a large number of levels by varying its input parameters. The representation used is usually more abstract---for example, the policy of a reinforcement learning agent~\citep{pcgrl} or the weights of a neural network. \citet{pplgg} use a genetic algorithm to search for parameter vectors that determine the behaviour of a non-deterministic (and thus reusable) generator.

%% file: 0_Chapter_GenerationMethodology.tex
\section{Generating Diverse Levels Quickly}
\label{chap:generation_method}
Our aim is to develop a method capable of automatic content generation with the following desiderata: the approach must a) be capable of generating levels quickly so that it can be used in real-time games; b) not require any training data, since that would require effort on the part of designers to create in the first place; c) use as little game specific information as possible, resulting in a general system; and d) should generate diverse and playable levels.

This section describes our method, Procedural Content Generation using NEAT and novelty search (PCGNN), which satisfies the above requirements, i.e.  generating diverse levels in real time without any existing training data or game-specific learning signals.

As previously mentioned, we use NEAT~\citep{neat} to evolve a neural network that generates levels. This is divided into two parts, training (evolution) and inference (generating levels).

\subsection{Generation Process}
\label{sec:method_level_gen}

\begin{figure}[h!]
    \centerfloat
    \includegraphics[width=0.9\linewidth]{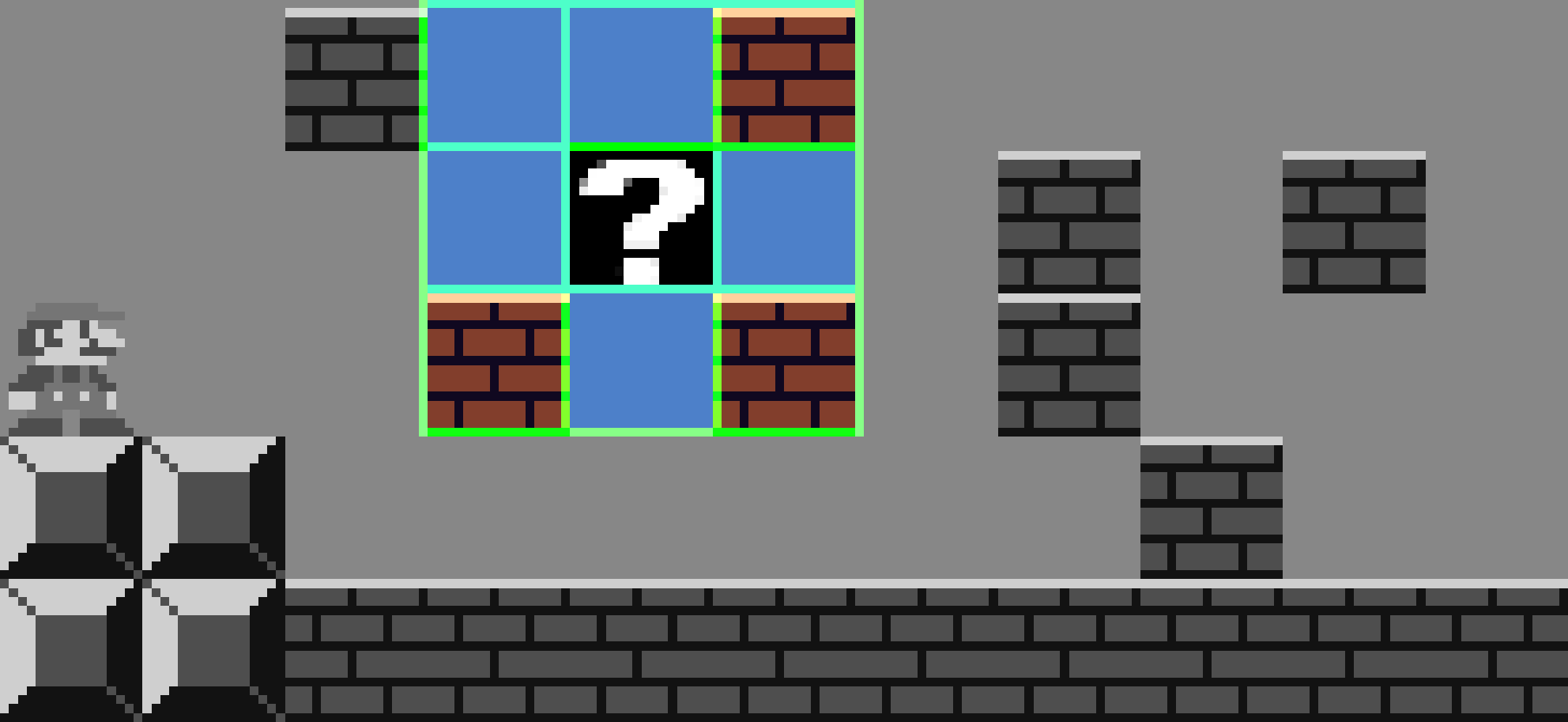}
\caption{An illustration of the level generation process on \mario. The ``?'' is the current tile to be predicted, whereas the highlighted area is the context that the algorithm uses to predict the center tile.}
\label{fig:tikz_neat_gen}
\Description[Example 2D tilemap level with a $3\times 3$ region highlighted.]{An extract of a \mario level, with a $3 \times 3$ area highlighted---which represents the observability range of the model---and a question mark in the middle of this area, indicating what tile the model should predict.}
\end{figure}

Given a neural network, we generate a level as follows.
We first generate a random 2D array of tiles, and then for each tile (e.g. the question mark in \autoref{fig:tikz_neat_gen}) we input the surrounding tiles (those highlighted in \autoref{fig:tikz_neat_gen}) into the network using either integer tile values or one-hot encoding. The output is used to predict what tile type should be placed at the original location. To predict boundary tiles, we pad the level with a row or column of $-1$ on all sides.
This method, which is similar to applying a convolution, is used instead of generating the entire level in one step, because the model is then able to generate levels of arbitrary size~\citep{volz2018evolving,CPPN2GAN, edrl}, and make locally consistent choices~\citep{hoover2015composing,summerville2018procedural_pcgml}. We input random noise into the network and perturb all of the inputs to enable the generation of multiple levels, rather than a single deterministic one. This facilitates reuse, allowing the same generator to be used multiple times. This adding of randomness is similar to inputting random noise into a GAN to generate new data~\citep{creswell2018generative}.
This process is performed sequentially, so the previous predictions become part of the level, and are used by the network as inputs when predicting adjacent tiles.

\subsection{Training Process}
\label{sec:method_train}
We use the standard NEAT algorithm~\citep{neat}, and evolve the population using fitness functions that are described in the next section.
After evolving the population for a set number of generations, we use the individual with the highest fitness as our final level generator. 
This can then be used to rapidly produce many new levels, as generation only requires the network to be queried---no learning or searching is performed at inference time.

\subsection{Fitness Functions}
\label{sec:method_fitness_functions}
To evaluate a network, we first generate $N$ levels and use these to compute the fitnesses, which we then average.
We use different fitness functions for different purposes, all of which are scaled between 0 and 1 unless otherwise stated. When using multiple fitness functions, we set the final objective as a weighted sum of these fitnesses.
This follows work by \citet{gomes2015devising}, who found that this method performs comparably to using multi-objective optimisation techniques, specifically when using the novelty metric.

\subsubsection{Solvability}
This fitness function determines if the level is solvable, using a search algorithm such as A*~\citep{astar_method}. The fitness value is 1 if the level is solvable and 0 otherwise.

\subsubsection{Novelty}
We also use the novelty metric~\citep{novelty_search} to evaluate the diversity of level generators. 
As described above, we do this by generating $N$ levels for each generator. Let $G_i$ denote generator $i$ and $L_{in}$ denote level $n$ from generator $i$. 
Then we define the distance function between two generators as 
$$D(G_i, G_j) = \frac{1}{N}\sum_{n=1}^N{d_l(L_{in}, L_{jn})},$$

where $d_l(L_{in}, L_{jn})$ is the distance between two levels. This can be computed using different methods such as Visual Diversity~\citep{constrained_novelty}, which measures the fraction of non-matching tiles (i.e. the normalised Hamming distance) or perceptual image hashing~\citep{monga2006perceptual,hadmi2012perceptual}, which gives a low value for images that look similar and a large value for those that look different. The exact function used can have a large effect on the diversity characteristics of the generated levels~\citep{searching_for_good_and_diverse_game_levels}, which we illustrate in the \refappendix{sec:appendix}.
Using the above, the novelty score of a generator, with respect to the current population and the novelty archive, is:
$$
\text{Novelty}(G_i) = \frac{1}{K}\sum_{k=1}^K { D(G_i, G_k) },
$$ where $k$ iterates over the $K$ closest neighbours of $G_i$, from either the population or the archive. Our archive of previous individuals is formed by randomly adding $\lambda$ individuals to it at every generation~\citep{gomes2015devising} instead of adding only individuals that obtained a high novelty, as in the original work~\citep{novelty_search}. This is generally preferable, as it leads to more robust parameters and uniform exploration~\citep{gomes2015devising}.

\subsubsection{Intra-Generator Novelty}
The above novelty metric rewards novelty between different generators, but the final generator that we use is a single neural network. Thus it is useful to reward generators that can generate multiple diverse levels. To this end, we also use the intra-generator novelty metric, which is similar to the above, but it simply measures the novelty between the $N$ levels generated by one generator, and sets the intra-novelty of this generator to be the average novelty of its $N$ levels. We do not use an archive of previous individuals when computing this value.

\subsubsection{Other}
Additional fitness functions can be used to generate levels with specific properties (e.g. longer paths or more connected regions~\citep{pcgrl}), and other feasibility criteria that do not affect solvability, such as not placing enemies in mid-air. 
These could be used to inject some expert knowledge into the generation or to enforce constraints, but we do not use these in our experiments.
\subsection{Summary}
Our method uses a learned level generator in the form of a neural network to be able to quickly generate levels after a one-time, offline training period. We evolve this network using NEAT so as to not require any training data, and use general fitness functions, namely solvability and novelty, to provide game-independent learning signals that also incentivise exploration.
\autoref{alg:nov_neat_training} shows pseudocode for the training procedure.
\input{1_files_algo_method_pseudocode}

%% file: 1_files_algo_method_pseudocode.tex
\begin{algorithm}
    \caption{\pcgnn Training Procedure}\label{alg:nov_neat_training}
    \begin{algorithmic}

        \REQUIRE $G \geq 1$ \COMMENT{Number of generations}
        \REQUIRE $N \geq 1$ \COMMENT{Number of levels to generate per network}
        \REQUIRE $K \geq 1$ \COMMENT{Number of neighbours for novelty}
        \REQUIRE $w_f, w_n > 0$ \COMMENT{Weighting for standard fitness and novelty}
        
        \STATE $Pop \gets $ random initial population of networks
        \FORALL {generations $g \in \{1,...,G\}$} 
        \phase{Generate Levels + Calculate Simple Fitnesses}
            \FORALL {$net_i \in Pop$}

                \STATE Generate $N$ levels, $L_{i1}, ..., L_{iN}$ from network $net_i$
                
                \COMMENT{Fitness is e.g. Solvability and/or Intra-Novelty}

                $Fitness(net_i) \gets mean(Fitness(L_{ij})), \forall j$ 
            \ENDFOR
            
            \STATE Set $Distance(i, i) = \infty, \forall i$
            \FORALL {$net_i \in Pop$}
                \FORALL {$net_j \in Pop, j > i$}
                    \STATE \COMMENT {$d_l$ could be e.g. Visual Diversity}

                    \STATE $Distance(i, j) \gets \frac{1}{N} \sum_{n=1}^N d_l(L_{in}, L_{jn})$ 
                    \STATE $Distance(j, i) \gets Distance(i, j)$
                \ENDFOR
            \ENDFOR
            \phase{Calculate Novelty Fitness}
            \FORALL {$net_i \in Pop$}

                \STATE Sort $Distance(i, \cdot)$ ascendingly
                \STATE $Novelty \gets \frac{1}{K} \sum_{k=1}^K Distance(i, k)$ 

                \STATE \COMMENT{Linear combination of different fitnesses}
                \STATE $Fitness(net_i) \gets w_f \cdot Fitness(net_i) + w_n \cdot Novelty$ 

            \ENDFOR
            \STATE Update population using calculated fitnesses.
        \ENDFOR
        \STATE Return best individual network.
    \end{algorithmic}
\end{algorithm}

%% file: 0_Chapter_Experiments.tex
\section{Experiments}
\label{chap:experiments} 
Here we detail our experimental setup, the baselines we compare against, and the metrics we use to evaluate the levels.

We consider two 2D tilemap games: firstly, a \maze game consisting of ``wall'' and ``empty'' tiles, where the objective is to find a path between the top left and bottom right and, secondly, a simplified \textit{Super Mario Bros.}, without powerups and only Goombas as enemies.

\subsection{Baselines}
\label{sec:baselines}
We compare \pcgnn to several baselines. The first is a genetic algorithm-based approach due to \citet{ferreira_mario}, which was originally developed for \mario. In this method, there is a population for each tile type in the level (e.g. ground, enemies, etc.), and each of these is evolved independently using a genetic algorithm with an entropy ~\citep{shannon1948mathematical} or sparseness-based fitness function~\citep{cook_evolve}. For the \maze, we simply use the 2D grid directly as the genotype, using partial solvability and entropy as the fitness functions.

Partial solvability is a less sparse version of the solvability fitness function described previously, directly applicable for the \maze, and it returns $\frac{1}{3}(\vmathbb{1}_{start} + \vmathbb{1}_{end} + \vmathbb{1}_{connected})$, where $\vmathbb{1}_x = 1 \text{ if } x$ is true; $0$ otherwise, $start, end$ mean that the starting and ending tiles are empty, and $connected$ is true when the starting and ending tiles are connected by a path of empty tiles.  This is to give potentially more guidance to the algorithm than the sparse solvability described above, and was found to perform better for the direct genetic algorithm specifically.
Entropy is a fitness function where we split the levels into non-overlapping chunks, calculate the entropy~\citep{shannon1948mathematical} of each chunk, and return the average. The fitness of an individual is then calculated as the distance to the ``desired entropy'', which is specified by the user~\citep{ferreira_mario}. Calculating the entropy of a chunk works as follows:
for each tile type $t$ (e.g. wall and empty for the \maze game), we count the number of tiles that have that value. From here we can construct a probability distribution and calculate the entropy. For example, let $m$ be the number of tiles in each chunk, then $P_t = \frac{\text{Count}(t)}{m}$ is the probability of tile $t$, and the entropy for that chunk is defined as 
$$H(P) = -\sum_{t=1}^{\text{n}}{ \log_2(P_t) P_t },$$ where $n$ is the number of unique tile types. We normalise the above by dividing by $\log_2(n)$ when $n>2$.

We also compare how this method performs with the novelty metric as a fitness function, similar to \citet{constrained_novelty}, but without using the two population approach. This method is henceforth referred to as DirectGA.

Our second baseline is Procedural Content Generation via Reinforcement Learning (PCGRL)~\citep{pcgrl}, where the level generation process is modelled as a reinforcement learning problem. For the \maze we use reward functions that incentivise solvable levels with path lengths in a certain range, and for \mario we reward solvability of the level, feasibly placed enemies as well as having the number of enemies in a certain range. We use the original implementation,\footnote{\url{https://github.com/amidos2006/gym-pcgrl}} and only consider the ``turtle'' and ``wide'' representations, as these performed the best in the original work.

The first baseline was chosen to compare against an evolutionary approach that directly searches for levels, as opposed to our method searching in generator space. PCGRL was chosen to specifically compare against another method that learns a level generator and should have a fast generation time.

We do not consider CPPN2GAN~\citep{CPPN2GAN} or the approach by \citet{volz2018evolving} as baselines, since these methods require existing levels as training data. Similarly, EDRL~\citep{edrl} uses a GAN-based chunk generator trained on existing \mario levels, though in principle any parameterised generator could work.

\subsection{Metrics}
\label{sec:metrics}
To determine the characteristics of the levels that we generate, we use different metrics described below.

\subsubsection{Solvability}
Solvability is a relatively simple metric: we determine if the level is solvable using a breadth-first search or A* agent to traverse the level from the starting state to the goal state. 
\subsubsection{Generation Time}
Since one of our goals is to generate levels in real time, we also measure the time it takes each method to generate one level, after (possibly) performing offline training.
\subsubsection{Difficulty}
We use the leniency~\citep{smith2010launchpad_leniency,shaker2012evolving}, and A* difficulty~\citep{beukman_metrics} metrics to evaluate the levels based on difficulty. The A* metric measures the number of nodes unnecessarily expanded by the A* algorithm, while leniency calculates how forgiving each obstacle is to a player's mistakes, and averages this across the entire level. 
For the \maze, this measures the fraction of dead ends in the level~\citep{beukman_metrics}.
\subsubsection{Diversity}
We use the compression distance~\citep{li2004similarity,shaker2012evolving} metric, particularly the Normal variant as described by~\citet{beukman_metrics}, which measures diversity by how much space is saved when compressing two strings (levels) together versus separately. We also use the A* diversity metric~\citep{beukman_metrics}, which compares the trajectories of an A* agent on pairs of levels---levels that are solved in different ways are marked as diverse

\subsection{Implementation details}
We use a standard implementation~\citep{neat_python_cite} of NEAT to perform the evolutionary process. For a game with $n$ tiles, the network outputs a single tile type at each step. This is effectively an $n$-class classification problem. For the \maze, since there are only two tiles, we use one neuron and a threshold: if the neuron's activation is larger than this threshold, it is a wall, otherwise an empty space. For \mario, the network has $n$ output neurons, and the one with the largest activation is chosen as the tile. Each tile is represented as either a binary number (for the \maze) or a one-hot encoded vector for \mario.

\subsubsection{Agents}
For \mario, we use the updated Mario-AI framework\footnote{\url{https://github.com/amidos2006/Mario-AI-Framework}} to import our generated levels and use the A* agent by Robin Baumgarten\footnote{\url{https://github.com/amidos2006/Mario-AI-Framework/tree/master/src/agents/robinBaumgarten}}~\citep{togelius20102009_with_baumgarten, togelius2013mario} to evaluate them.

Since a level might be solvable in multiple different ways, and the A* agent sometimes performs differently on the same level of \mario (due to the relative complexity of the simulation), we run the diversity and difficulty metrics 5 times on each level (for both games) and average the results. For the solvability metric, we also run the agent five times, and if it solved the level in any of these five runs, we label it as solvable.
The solvability fitness function uses a less complex but much faster simulation,\footnote{\url{https://github.com/amidos2006/gym-pcgrl/blob/master/gym_pcgrl/envs/probs/smb/engine.py}} and we only use the full fidelity simulation when evaluating the levels after training and generation. Conversely, the \maze game is simpler, and so we use a standard implementation of A*.
\subsection{Experimental Setup}
We compare our method of level generation against the aforementioned baselines using the metrics discussed above. 
For all generation methods, we perform a hyperparameter search and in all cases report the best result obtained.

All level generation experiments are run over five different random seeds with the average and standard deviation reported. Each of these five runs consist of generating 100 different levels, and the metrics of the 100 levels are averaged to obtain a single value for each seed. Since the diversity metrics calculate scores for each pair of levels, we follow \citet{horn2014comparative} and measure the diversity between a group of $N$ levels by calculating the average of the $\frac{N(N-1)}{2}$ pairwise diversity scores.

When reporting metrics, we only use the solvable levels and average over their values to minimise the effect that unsolvable levels have on the metrics. Since the fraction of solvable levels is mostly high, we still average over many levels. We also use a random baseline but this rarely generates solvable levels. Thus, we consider all levels \textit{only} for this baseline.

In the next sections, we use the following names to refer to the specific baselines. See the \refappendix{sec:appendix} for detailed hyperparameters.
\begin{description}
    \item[\pcgnn] This is our NEAT and novelty search method.
    \item[DirectGA] Only applicable to \mario , this is the genetic algorithm with the same parameters as in the original work~\citep{ferreira_mario}.
    \item[DirectGA+] The DirectGA method (for both \maze and \mario), where we perform a hyperparameter search to obtain levels with high solvability, breaking ties based on other metrics, like compression distance.
    \item[DirectGA (Novelty)] This uses the DirectGA method, but adds in novelty as a fitness function, with uniform weights to all fitness components. For \mario, we only consider population sizes and numbers of generations larger than 50, to ensure that the novelty search actually takes effect.
    \item[PCGRL (Wide/Turtle)] Using PCGRL with the ``wide'' or  ``turtle'' representations respectively~\citep{pcgrl}.
    \item[Random] Tiles are selected uniformly at random.
\end{description}
For PCGRL, we attempt to train the method for the same 100 million timesteps as stated by \citet{pcgrl}, but the \mario training process was slower than the \maze (possibly due to the much larger level sizes), and we only managed to perform about 12 million steps for ``wide'' and 8 million for ``turtle'' in 3 days, which are the models we use to report results here.
When performing inference for \mario, we limit the maximum number of steps per level to 10 000, as without this the PCGRL model sometimes becomes stuck in a loop. 

Finally, for both DirectGA approaches, each level is generated from a separate evolution process, starting from a unique initial population. The individual with the highest fitness after $G$ generations is selected as the level. This is repeated 100 times per seed.
\subsubsection{Statistical Approach}
For the following sections, we use the notation $\mu_{\text{b}}(M)$ to refer to the mean of metric $M$ when using method $b$, and $\sigma_{\text{b}}(M)$ denotes the standard deviation. All results in the following tables are of the form $\mu \text{ } (\sigma)$. 

Our statistical analysis procedure is as follows. We first performed the Kruskal–Wallis test~\citep{kruskal1952use} to determine whether a statistical significant difference exists at all. Then, we performed pairwise Mann-Whitney U tests~\citep{mann_whitney_test} between \pcgnn and each baseline's result. This was done because for most of the metrics tested, at least one method failed a Shapiro-Wilk normality test~\citep{shaphiro1965analysis} with $p < 0.05$. Finally, since using multiple pairwise tests increases the risk of making a type I error~\citep{jafari2019and,dropout_statistical_analysis}, we subsequently perform Bonferroni Error correction~\citep{bonferroni1936teoria}. If we find a statistically significant result, we also calculate the Cohen's d value~\citep{cohen2013statistical}, $d = \frac{\mu_{\text{pcgnn}}(M) - \mu_{b}(M)}{\sigma_{\text{pcgnn}(M)}}$ to measure how large this difference is. 
In the tables shown in the next section, we use \textbf{bold} to denote a statistically significant result ($p < 0.05$) and $^{\dagger}$ to denote a large effect size ($| d | > 0.8$).

All results where we report time were run on similar hardware (details in \refappendix{sec:appendix}), with minimal other processes running to enable a fair comparison.

%% file: 0_Chapter_Results.tex
\section{Results}
\label{chap:results}

We compare \pcgnn against the baselines on generation time in \autoref{sec:exp_time} and on solvability, diversity and difficulty in 
\Autoref{sec:exp_solv,sec:exp_div,sec:exp_diff}, respectively. We finally consider generalisability in \autoref{sec:exp_gene}. Example levels and \pcgnn's fitness curves are shown in the \refappendix{sec:appendix}.
\subsection{Generation Time}
\label{sec:exp_time}
Since one of our goals is to generate levels in a fast, real-time fashion, we compare the generation time of our method to our baselines under the null hypothesis $$H_0: \mu_{\text{pcgnn}}(t) \geq \mu_{b}(t)$$ i.e. that our method is comparable or slower at generating levels than the baselines.

Since \pcgnn has a two stage process of evolving the generator and then using the generator to generate levels, we split the total time taken for this into \textit{training} and \textit{testing}. Training is the time required to evolve the generator, which happens once. Generation then refers to querying this generator. The DirectGA approaches generate levels as needed, and so have 0 training time.

We use a one-sided Mann-Whitney U test for generation time and a two-sided test for training time. \autoref{tab:neat_vs_baselines_gen_time} presents the results.

\begin{table}
\caption{Train and generation times. Lower is better.}
    \label{tab:neat_vs_baselines_gen_time}
    \centerfloat
    \begin{adjustbox}{width=1\linewidth}
        \input{998_tab_src_results_v400_methods_generation_time.tex}

    \end{adjustbox}
\end{table}

We thus reject the null hypothesis that our method's generation time is comparable or slower than our baselines (except for the Random method, which does not perform any computation). We see a large effect size, and at least an order of magnitude improvement in generation speed. For training time, we find a statistically significant difference compared to all other methods (using a two-sided test). This is to be expected, since the DirectGA method requires no training time and PCGRL requires substantially more.

PCGRL has quite a large variance in generation times, and we hypothesise that this is because it generates a level until a specified reward threshold is met, instead of iterating for a fixed number of iterations like \pcgnn and DirectGA.

We do not parallelise the fitness function calculations for either DirectGA or \pcgnn, although doing so could improve performance, notably generation time for DirectGA. 
This would largely depend on the number of cores the generating machine has, and many cores are often not a given on a user's machine~\citep{steam_hw_survey}. Similarly, we only take the top individual after performing the evolution---it would be more efficient to take more individuals from the final generation, but they might lack in diversity or feasibility~\citep{quality_diversity_pcg}.

\subsection{Solvability}
\label{sec:exp_solv}
Since we can generate levels quickly, we next investigate the quality of these levels, starting with solvability. Here we compare solvability scores between our method and the baselines, with the null hypothesis that our method has the same mean solvability as the other methods. \autoref{tab:neat_vs_baselines_solv} shows these results.

\begin{table}[H]
    \centerfloat
    \caption{Solvability. Higher is better.}
    \label{tab:neat_vs_baselines_solv}
    \begin{adjustbox}{width=1\linewidth}
        \input{998_tab_src_results_v400_methods_solvability.tex}

    \end{adjustbox}
\end{table}

We find no statistically significant difference between our solvability, the solvability of PCGRL and that of DirectGA+ for the \maze. For \mario, we find no statistically significant difference between \pcgnn's solvability and that of the baselines.

PCGRL's solvability on \mario is much lower than on \maze, possibly due to training for fewer timesteps, and increasing the training budget could improve this.
Our solvability is thus perfect for \maze, and still very high on \mario.
The above result, coupled with our fast generation times, indicates that we can generate a larger number of solvable levels than the other methods in the same amount of time.

\subsection{Diversity}
We now compare the diversity of the generated levels using the compression distance and A* diversity metrics. The null hypothesis is that the diversity of \pcgnn is comparable to the baselines.
\label{sec:exp_div}
\begin{table} 
    \centerfloat
    \caption{Diversity metrics. Higher is better.}
    \label{tab:neat_vs_baselines_cd}
    \begin{adjustbox}{width=1\linewidth}
        \input{998_tab_src_results_v400_methods_cd.tex}

    \end{adjustbox}
\end{table}
Results are shown in \autoref{tab:neat_vs_baselines_cd}, where we see that \pcgnn's A* diversity is comparable to DirectGA+ for \mario, but rather different from both PCGRL representations and quite different to all methods for \maze. Because of the high variance in \pcgnn's compression distance for \mario, we find no statistically significant difference between our method and the baselines.

The A* diversity metric evaluates DirectGA's \mario levels (which are relatively flat without jumps or platforms) as quite similar, since  the same rough trajectory solves most levels.
PCGRL's levels, on the other hand, require substantially different trajectories in general, indicating that the levels cannot all be solved using the same path.
\pcgnn is somewhere in between, indicating that different trajectories and strategies are required, but the difference, on average, is not as large as PCGRL.

We also note that using novelty for the DirectGA does not improve the diversity metrics (compared to DirectGA+), potentially indicating a mismatch between what novelty rewards and what the metrics measure.

\subsection{Difficulty}
\label{sec:exp_diff}

Next we investigate the difficulty of our levels, as measured by the leniency~\citep{smith2010launchpad_leniency, shaker2012evolving} and A* difficulty~\citep{beukman_metrics} metrics. These metrics both attempt to measure the abstract notion of ``difficulty'', but they do so in different ways. Further, levels with higher leniency correspond to levels with lower A* difficulty and vice versa. We again use the null hypothesis that our method has the same mean difficulty as our baselines.

Results in \autoref{tab:neat_vs_baselines_leniency} show that for the \maze, our method generates quite lenient levels, and this is confirmed by the lower value of the  A* difficulty metric. For \mario, our levels have quite low leniency, but this varies drastically. We therefore only find a statistically significant difference between \pcgnn and the DirectGA, which generates very flat levels and thus has relatively high leniency.

\subsection{Generalisability}
\label{sec:exp_gene}

This experiment asks the question: ``after we have learnt enough to generate one level of size $Y$, how long does it take to generate a level of size $X \neq Y$?'' We hypothesise that \pcgnn will be able to generate levels of arbitrary size quickly, without any retraining, as we use a size-agnostic generation method. We test this claim by taking an evolved generator (that was trained only to generate levels of size $14 \times 14$), and using it to generate levels of different sizes. We compare this against the DirectGA method in \autoref{fig:mario_linegraphs}.

We find that \pcgnn's generation speed is still substantially higher than the DirectGA's (for both \mario and \maze), even as we increase the level size. \pcgnn generates \maze levels that have $2000^2$ tiles faster than DirectGA generates levels with $100^2$ tiles. In the \maze domain, our method achieves perfect solvability for all tested sizes, whereas the direct genetic algorithm's solvability decreases drastically as the level size increases. One potential explanation is that we keep the population size and number of generations constant as the level size increases; prior work has shown that standard genetic algorithms do not always perform very well with high dimensional problems~\citep{yao1998scaling, liu2001scaling}.
For \mario, we see a slight downward trend in solvability as the level size increases, whereas DirectGA remains constant. This DirectGA consistency is because the initial level, instead of being random as in the \maze game, is already solvable without any extra input from the genetic algorithm. Again, even with some loss of solvability as the level width increases, our fast generation time can ameliorate this issue.


\begin{table}[t!]
    \caption[Difficulty. The optimal value usually depends on the player's skill and designer's intentions.]{Difficulty. The optimal value usually depends on the player's skill and designer's intentions.\footnotemark}
    \label{tab:neat_vs_baselines_leniency}
    \centerfloat
    \begin{adjustbox}{width=1\linewidth}
        \input{998_tab_src_results_v400_methods_leniency.tex}

    \end{adjustbox}
\end{table}

\footnotetext{The difference between the leniency of the Random baseline and \pcgnn is not statistically significant. The reason for this is that there were ties in the leniency values (multiple seeds had 0 leniency), leading to the asymptotic Mann-Whitney U calculation being used over the exact version~\citep{mann_whitney_test}. The normal approximation, while decent, is not perfect for a small sample size of 5~\citep{bellera2010normal}. This, combined with the Bonferonni correction, leads to a $p$ value of $0.055$, slightly above our threshold of $0.05$.}

Since PCGRL relies on a specific input size (as it uses a neural network that can only process a fixed size input), it cannot generalise to different sized levels without an expensive retraining.

\newcommand{\ww}{1\linewidth}
\newcommand{\wwwwww}{0.5\linewidth}
\newcommand{\hh}{0.625\linewidth}
\begin{figure*}[ht!]
    \centerfloat
    \begin{subfigure}[t]{\wwwwww}
        \centering\captionsetup{width=.9\linewidth}
        \resizebox{\ww}{\hh}{\includegraphics[width=\ww]{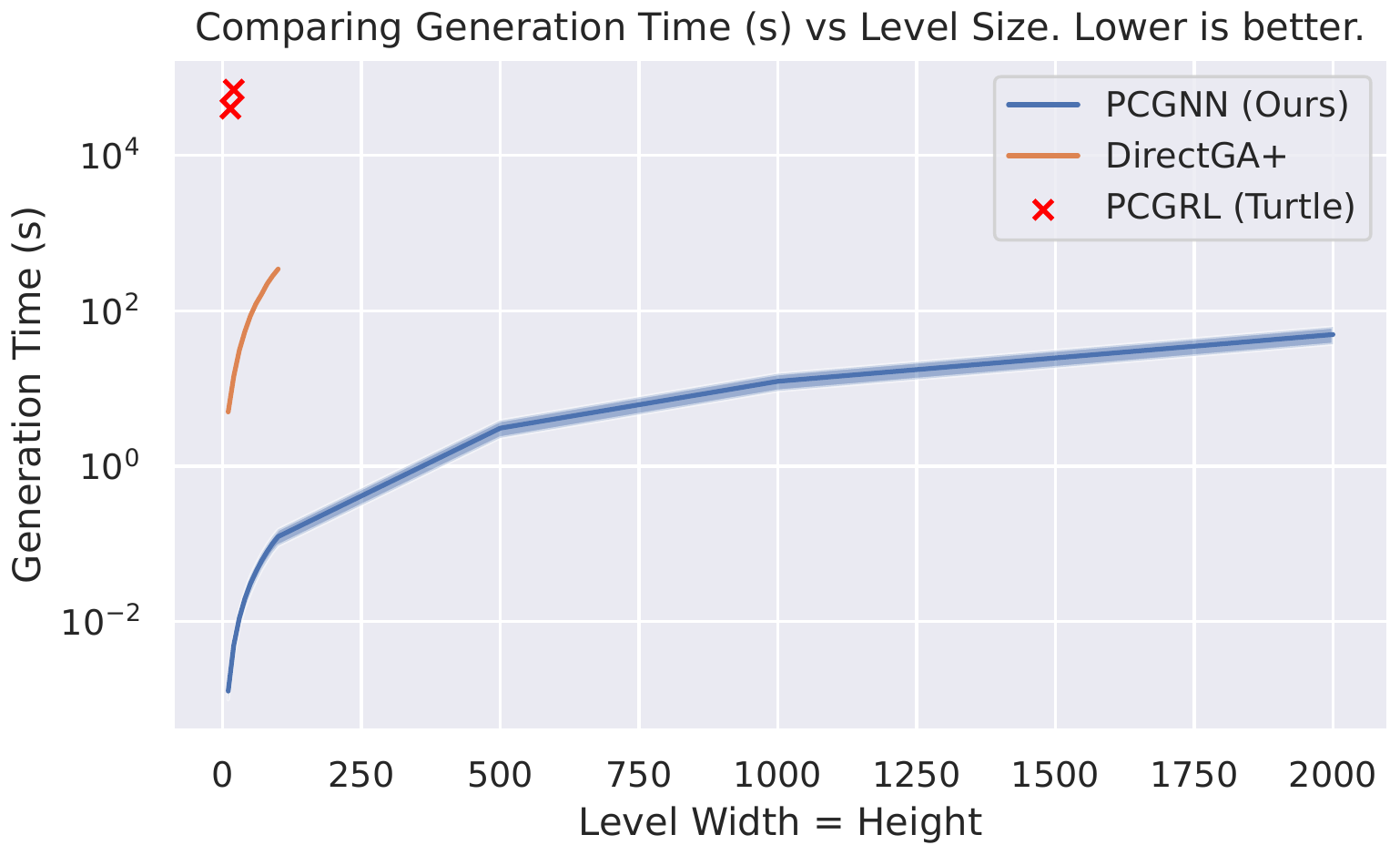}}
        \subcaption{A comparison of generation times on the \maze domain (lower is better) on a log scale.
        }
        \label{fig:sizes_time}

        \Description[Plot comparing generation time of all of the methods (on the \maze game) as the level size increases, showing \pcgnn generate levels faster.]{\pcgnn generates levels much faster than the other methods as the level size increases, while PCGRL generates levels the slowest.}
    \end{subfigure}
    \quad
    \begin{subfigure}[t]{\wwwwww}
        \centering\captionsetup{width=.9\linewidth}
        \resizebox{\ww}{\hh}{\includegraphics[width=\ww]{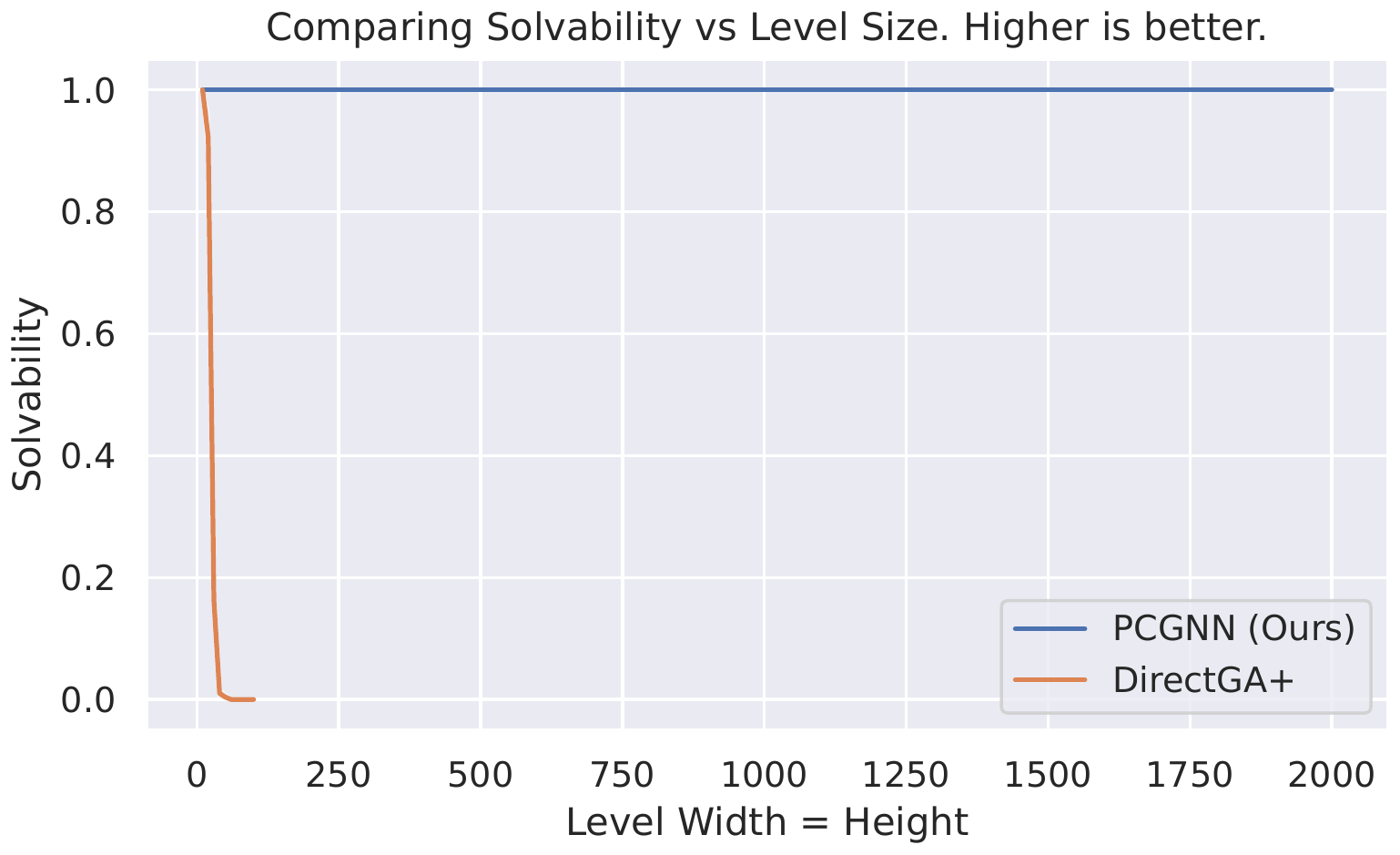}}
        \subcaption{A comparison of solvability with an increase in \maze sizes (higher is better). }
        \label{fig:sizes_solv}
        \Description[Plot comparing solvability of all of the methods (on the \maze game) as the level size increases, showing \pcgnn generate completely solvable levels for level widths from 20 to 100.]{\pcgnn generates solvable \maze levels as the size increases, while DirectGA+ rapidly falls off to 0 solvability as the level size increases past 40. PCGRL is not shown on this graph.}
    \end{subfigure}

    \centerfloat
    \begin{subfigure}[t]{\wwwwww}
        \centering\captionsetup{width=.9\linewidth}
        \resizebox{\ww}{\hh}{\includegraphics[width=\ww]{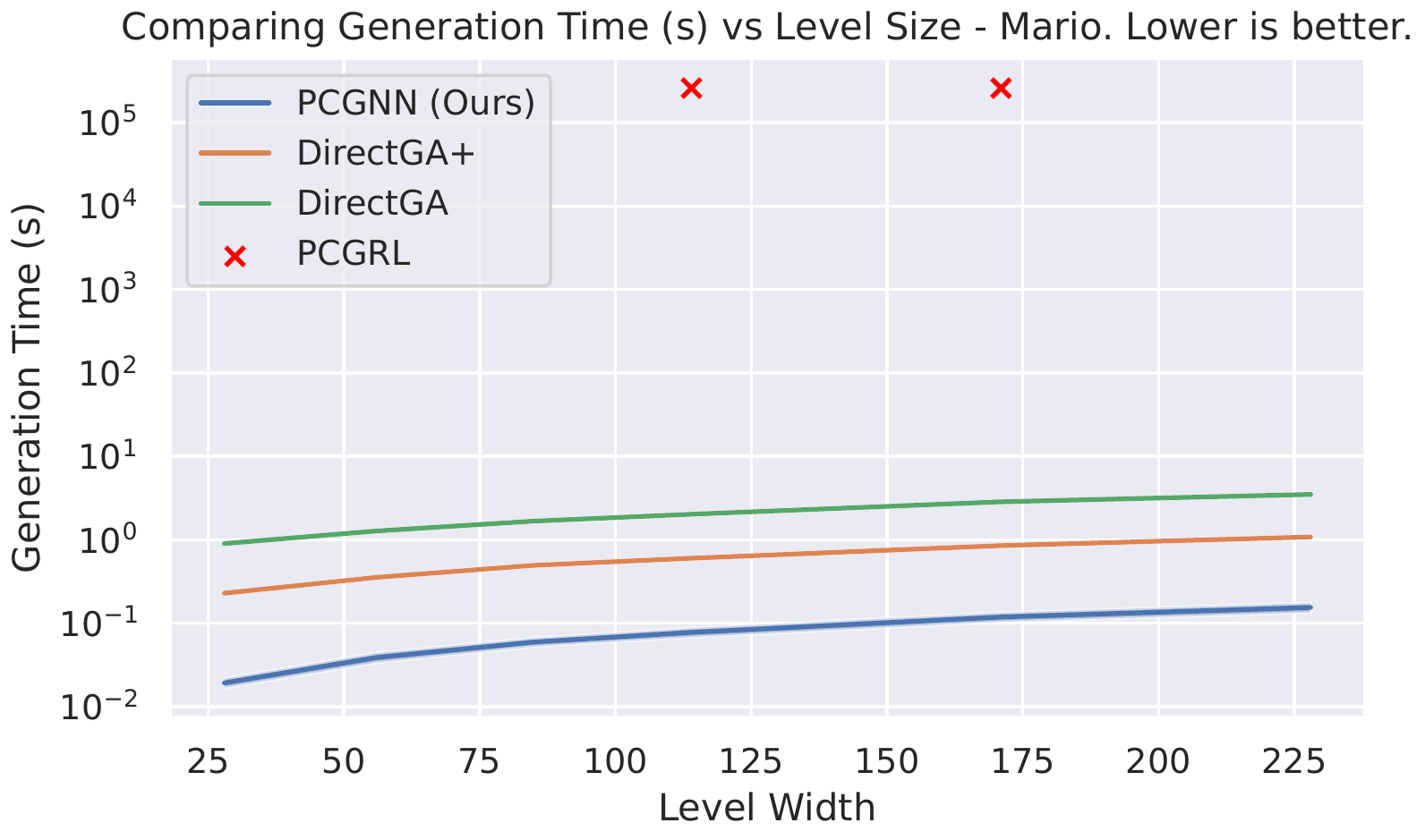}}
        \subcaption{A comparison of generation times on \mario (lower is better) on a log scale.
        }
        \label{fig:mario_sizes_time}

        \Description[Plot comparing generation time of all of the methods (on \mario) as the level size increases, showing \pcgnn generate levels faster.]{\pcgnn generates levels much faster than the other methods as the level size increases, while PCGRL generates levels the slowest.}
    \end{subfigure}
    \quad
    \begin{subfigure}[t]{\wwwwww}
        \centering\captionsetup{width=.9\linewidth}
        \resizebox{\ww}{\hh}{\includegraphics[width=\ww]{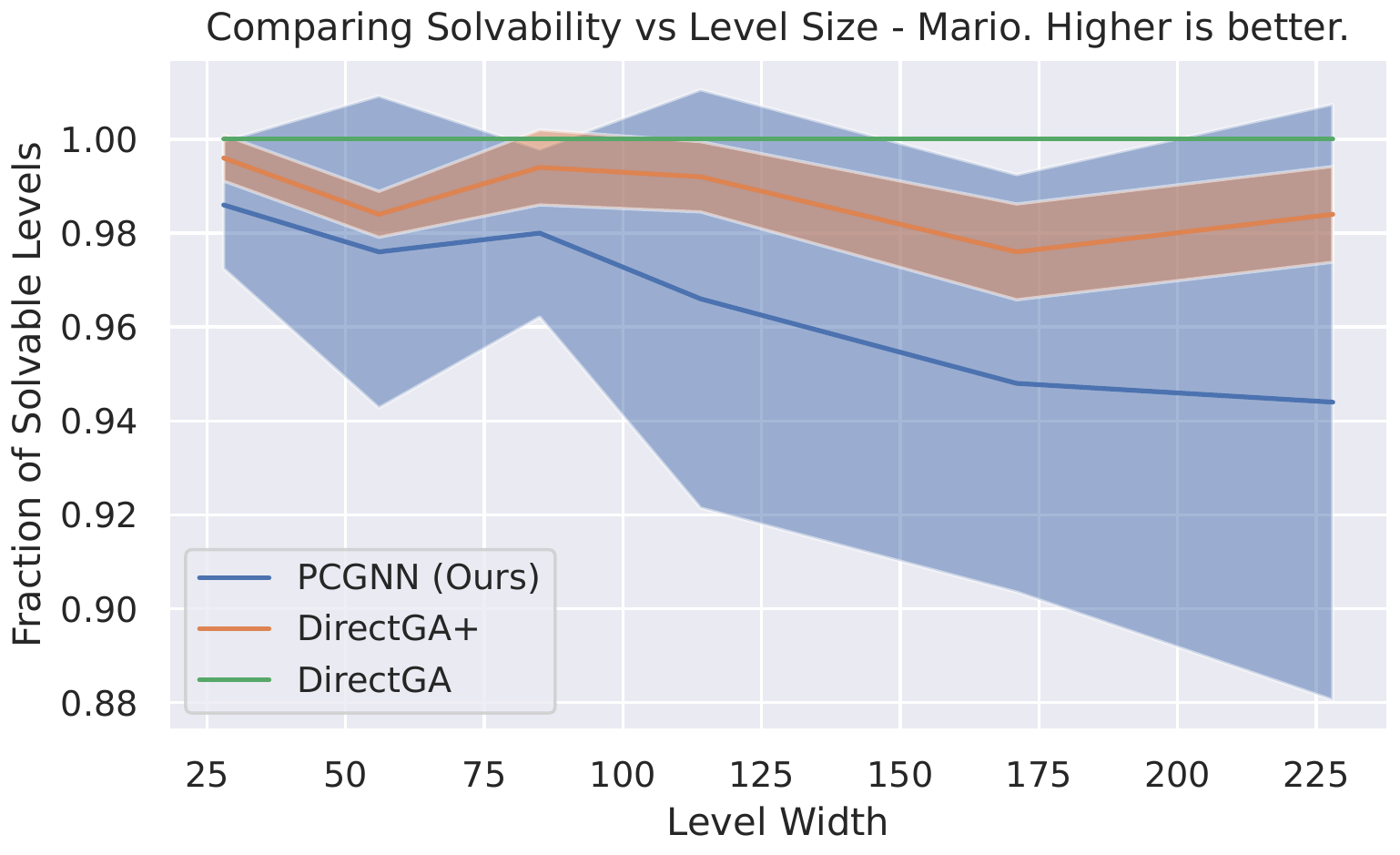}}
        \subcaption{A comparison of solvability with an increase in \mario sizes (higher is better). Note that the $y$-axis ranges from $0.88$ to $1$.}
        \label{fig:mario_sizes_solv}

        \Description[Plot comparing solvability of all of the methods (on \mario) as the level size increases, showing \pcgnn generate mostly solvable levels for level widths from 28 to 100, while DirectGA's solvability is near perfect]{\pcgnn generates \mario levels that are mostly solvable, tapering off linearly from 100\% for levels of width 50 to around 90\% solvability for levels of size 225. Both DirectGA methods stay above 95\% solvability. }
    \end{subfigure}
    \caption{Metrics for \maze (top row) and \mario (bottom row) levels of different sizes. Standard deviation is indicated by the shaded regions. For (a) and (c) we only plot two PCGRL points, since training for larger levels was prohibitively expensive.}
    \label{fig:mario_linegraphs}
\end{figure*}

\begin{figure}[h!]
    \centering
    \includegraphics[width=1\linewidth]{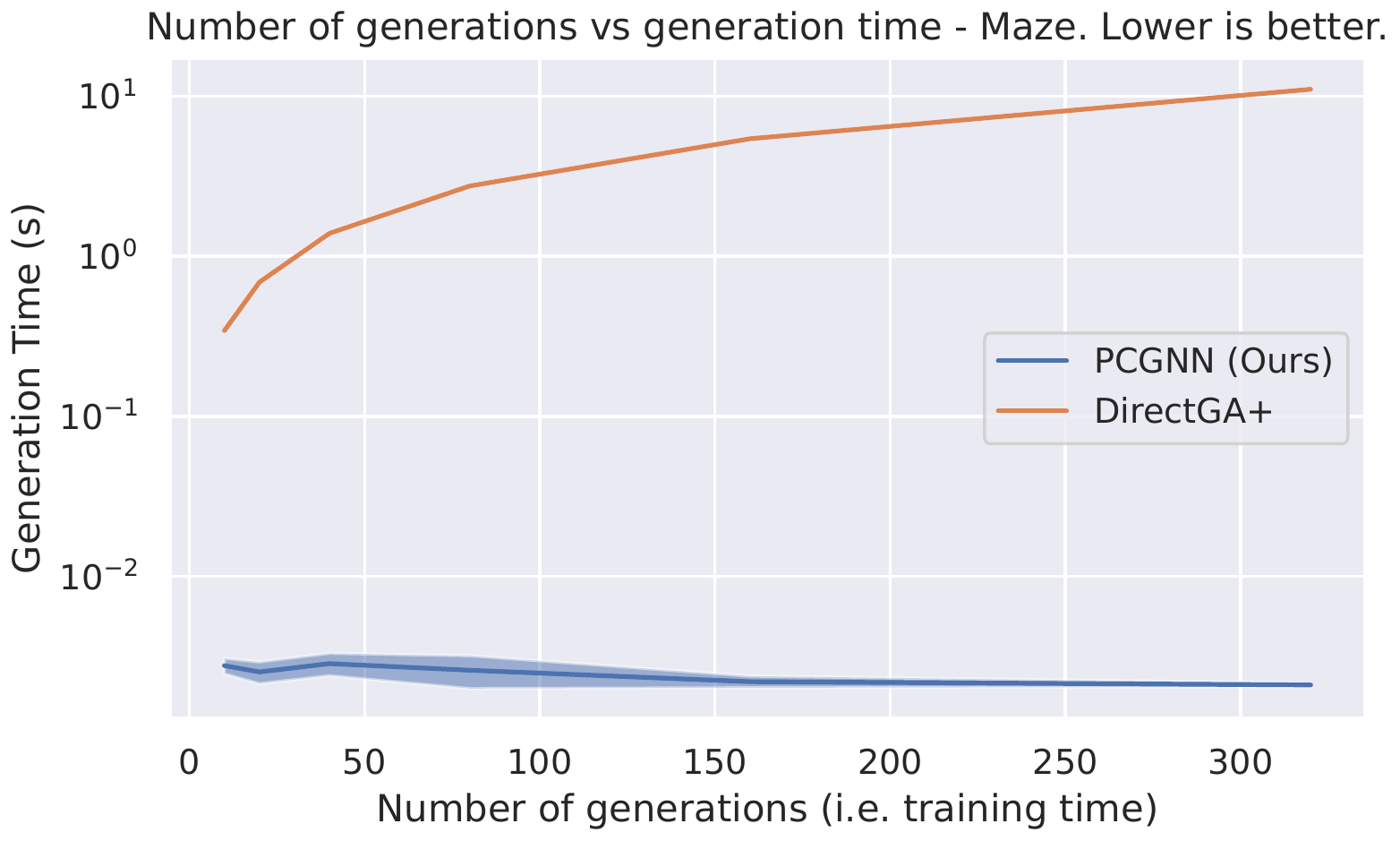}
    \caption{Comparing the effect of increasing the number of generations on generation time in the \maze domain with standard deviation shaded.}
    \label{fig:maze_number_of_generations}
    \Description[\pcgnn's generation time is roughly constant as the number of genetic algorithm iterations increase, while DirectGA's time increases drastically.]{DirectGA's generation time starts at around 0.7 seconds and increases to more than 10 as the number of generations increase from 10 to 300. \pcgnn's time remains at a roughly constant 0.004s.}
\end{figure}

The above results indicate that \pcgnn is well-suited to generalising to larger levels without the need for retraining. Thus, even when the fitness function used is expensive, or we require many generations or a large population size, all of this computation can be performed offline without negatively impacting the generation time. On the other hand, DirectGA's generation time is directly affected by these factors, making it increasingly unsuited for real-time generation, as shown in \autoref{fig:maze_number_of_generations}.

%% file: 998_tab_src_results_v400_methods_generation_time.tex
\begin{tabular}{lllll}
\toprule
{} & \multicolumn{2}{c}{Maze Time (s)} & \multicolumn{2}{c}{Mario Time (s)} \\
{} & \multicolumn{1}{c}{Generation} & \multicolumn{1}{c}{Train} & \multicolumn{1}{c}{Generation} & \multicolumn{1}{c}{Train  }\\
\midrule
PCGNN (Ours)       &          $2.4 \times 10^{-3}$ (0) &                     1100.46 (46.41) &                        0.08 (0.01) &                      11593 (347) \\
DirectGA           &                                 - &                                   - &   $\textbf{1.78 (0.02)}^{\dagger}$ &     $\textbf{0.0 (0)}^{\dagger}$ \\
DirectGA+          &  $\textbf{7.45 (0.18)}^{\dagger}$ &        $\textbf{0.0 (0)}^{\dagger}$ &   $\textbf{0.56 (0.01)}^{\dagger}$ &     $\textbf{0.0 (0)}^{\dagger}$ \\
DirectGA (Novelty) &  $\textbf{4.85 (0.05)}^{\dagger}$ &        $\textbf{0.0 (0)}^{\dagger}$ &  $\textbf{27.04 (0.49)}^{\dagger}$ &     $\textbf{0.0 (0)}^{\dagger}$ \\
PCGRL (Wide)       &  $\textbf{1.62 (2.10)}^{\dagger}$ &  $\textbf{245775 (6246)}^{\dagger}$ &   $\textbf{0.98 (1.10)}^{\dagger}$ &  $\textbf{259200 (0)}^{\dagger}$ \\
PCGRL (Turtle)     &  $\textbf{6.00 (5.49)}^{\dagger}$ &    $\textbf{40838 (496)}^{\dagger}$ &   $\textbf{7.35 (4.11)}^{\dagger}$ &  $\textbf{259200 (0)}^{\dagger}$ \\
Random             &                           0.0 (0) &        $\textbf{0.0 (0)}^{\dagger}$ &                            0.0 (0) &     $\textbf{0.0 (0)}^{\dagger}$ \\
\bottomrule
\end{tabular}

%% file: 998_tab_src_results_v400_methods_solvability.tex
\begin{tabular}{lll}
\toprule
{} & \multicolumn{1}{c}{Maze Solvability} & \multicolumn{1}{c}{Mario  Solvability}\\
\midrule
PCGNN (Ours)       &                                                 1.0 (0) &  0.98 (0.02) \\
DirectGA           &                                                       - &      1.0 (0) \\
DirectGA+          &                                                 1.0 (0) &  0.99 (0.01) \\
DirectGA (Novelty) &                        $\textbf{0.92 (0.04)}^{\dagger}$ &      1.0 (0) \\
PCGRL (Wide)       &                                                 1.0 (0) &  0.74 (0.03) \\
PCGRL (Turtle)     &                                             0.95 (0.08) &  0.70 (0.02) \\
Random             &  $\textbf{$\mathbf{2.0 \times 10^{-3}}$ (0)}^{\dagger}$ &  0.08 (0.03) \\
\bottomrule
\end{tabular}

%% file: 998_tab_src_results_v400_methods_cd.tex
\begin{tabular}{lllll}
\toprule
{} & \multicolumn{2}{c}{Maze} & \multicolumn{2}{c}{Mario} \\
\multicolumn{1}{c}{} & \multicolumn{1}{c}{Compression} & \multicolumn{1}{c}{A* Diversity} & \multicolumn{1}{c}{Compression} & \multicolumn{1}{c}{A* Diversity }\\
{} & \multicolumn{1}{c}{Distance} & \multicolumn{1}{c}{} & \multicolumn{1}{c}{Distance} & \multicolumn{1}{c}{}\\
\midrule
PCGNN (Ours)       &                       0.488 (0.002) &  0.13 (0.17) &          0.45 (0.10) &                       0.38 (0.11) \\
DirectGA           &                                   - &            - &             0.46 (0) &  $\textbf{0.10 (0.01)}^{\dagger}$ \\
DirectGA+          &                       0.493 (0.002) &  0.41 (0.01) &             0.55 (0) &                          0.33 (0) \\
DirectGA (Novelty) &                       0.494 (0.002) &  0.40 (0.01) &          0.44 (0.01) &                       0.20 (0.01) \\
PCGRL (Wide)       &  $\textbf{0.525 (0.021)}^{\dagger}$ &  0.43 (0.01) &             0.56 (0) &     $\textbf{0.55 (0)}^{\dagger}$ \\
PCGRL (Turtle)     &  $\textbf{0.513 (0.006)}^{\dagger}$ &  0.43 (0.01) &             0.56 (0) &     $\textbf{0.55 (0)}^{\dagger}$ \\
Random             &  $\textbf{0.494 (0.001)}^{\dagger}$ &  0.00 (0.01) &             0.47 (0) &  $\textbf{0.72 (0.02)}^{\dagger}$ \\
\bottomrule
\end{tabular}

%% file: 998_tab_src_results_v400_methods_leniency.tex
\begin{tabular}{lllll}
\toprule
{} & \multicolumn{2}{c}{Maze} & \multicolumn{2}{c}{Mario} \\
{} & \multicolumn{1}{c}{Leniency} & \multicolumn{1}{c}{A* Difficulty} & \multicolumn{1}{c}{Leniency} & \multicolumn{1}{c}{A* Difficulty }\\
\midrule
PCGNN (Ours)       &                       0.70 (0.08) &   0.06 (0.08) &                   0.17 (0.23) &                       0.24 (0.06) \\
DirectGA           &                                 - &             - &  $\textbf{1.0 (0)}^{\dagger}$ &                          0.27 (0) \\
DirectGA+          &  $\textbf{0.60 (0.01)}^{\dagger}$ &   0.16 (0.02) &                   0.31 (0.01) &  $\textbf{0.14 (0.01)}^{\dagger}$ \\
DirectGA (Novelty) &  $\textbf{0.59 (0.01)}^{\dagger}$ &   0.16 (0.01) &                   0.78 (0.04) &                       0.24 (0.01) \\
PCGRL (Wide)       &                       0.64 (0.04) &   0.21 (0.01) &                   0.47 (0.01) &                       0.29 (0.01) \\
PCGRL (Turtle)     &                       0.71 (0.02) &   0.18 (0.03) &                   0.47 (0.01) &                       0.28 (0.01) \\
Random             &          $2.3 \times 10^{-3}$ (0) &   0.52 (0.02) &                      0.14 (0) &  $\textbf{0.88 (0.02)}^{\dagger}$ \\
\bottomrule
\end{tabular}

%% file: 0_Chapter_Discussion.tex
\section{Discussion and Future Work}
\label{sec:disussion}
The method introduced here follows a recent trend focusing on level generators, which can be applied in real time after an offline training period~\citep{pcgrl, edrl} instead of searching for the levels directly. We demonstrate several advantages over these works, namely an even faster generation time, no need for training data or game-specific reward functions, and the ability to generalise to different sized levels without retraining. 

Limitations of the proposed method include generation time still scaling linearly with the level size, possibly leading to infeasible generation times for massive levels, although this performed well for the tested sizes.
Although we demonstrated comparable metric scores to the baselines, the sequential level generation process could still limit the characteristics of the levels by not taking into account global information. Further, while we do not require hand-specified fitness functions, the generated levels without these elements might not be desired in some cases. The \mario levels are also not very similar to the original game, but this was not our goal. More specific domain-knowledge can be incorporated, though, by changing the fitness function.

Future work could include analysing the effects of the different hyperparameters of our method (e.g. context or prediction sizes), or attempting to make the method more configurable during runtime. This would allow users to specify desired characteristics, similar to specifying the level size dynamically as we do here. 
Additional work could also investigate different novelty search distance functions (with a potential focus on simulation-based ones such as the A* diversity~\citep{beukman_metrics}), or attempt to apply multi-objective optimisation~\citep{tamaki1996multi} or illumination algorithms~\citep{map_elites} instead of na\"{i}vely optimising the sum of the different fitnesses. Another option is exploring the generation of endless levels, where the next parts of the level are generated as the player proceeds  (possibly using the player's behaviour as input to the network to generate adaptive levels)~\citep{edrl}.

%% file: 0_Chapter_Conclusion.tex
\section{Conclusion}
\label{chap:conclusion}
We introduced PCGNN, a NEAT and novelty search-based level generation approach that (1) does not require any training data; (2) requires minimal game-specific knowledge; (3) can be generally applied; and (4) generate levels of arbitrary size rapidly after an offline training period. We compare this method against a direct genetic algorithm and a reinforcement learning-based approach that learns a policy instead of directly searching for a level. Our method performs comparably to the baselines in terms of solvability, difficulty and diversity, while generating levels significantly faster.
Most importantly, our method requires no hand-engineered rewards or knowledge of game mechanics such as enemies. 
We believe this generality is an important step towards the widespread adoption of procedural content generation in a variety of contexts.

%% file: 0_Chapter_Acknowledgements.tex
This work is based on the research supported wholly by the \grantsponsor{nrf}{National Research Foundation of South Africa}{https://www.nrf.ac.za/} (Grant UID \grantnum{nrf}{133358}).

Computations were performed using High Performance Computing infrastructure provided by the Mathematical Sciences Support unit at the University of the Witwatersrand.

We thank the reviewers for their helpful and insightful comments, which helped to strengthen the final version of this paper.

%% file: 0_Chapter_Appendix.tex
\appendix
\section*{Appendix}
\label{sec:appendix}
Here we list the detailed hyperparameters that were used in the above experiments. Implementation details can be found in our source code located at \url{https://github.com/Michael-Beukman/PCGNN}.

\subsection*{Hardware Details}
We ran the experiments on Intel i9-10940X CPUs, and PCGRL's training and inference was performed on NVIDIA RTX3090 GPUs.
\subsection*{Hyperparameter Search}
We performed a small grid search over all variables for each method with between 2 and 4 values per variable. The hyperparameter set was chosen to maximise solvability, with compression distance used as a tie-breaker. \Autoref{tab:hps_lists_pcgnn,tab:hps_lists_directga} contain the values we tested.

\begin{table*}[b]
    \centerfloat
    \caption{The hyperparameters we searched over for PCGNN. $[0, 1]$ means the search was performed over the interval.}
    \label{tab:hps_lists_pcgnn}
    \begin{adjustbox}{width=1\linewidth}
        \input{994_HP_PCGNN.tex}
    \end{adjustbox}
\end{table*}

\begin{table*}[b]
    \centerfloat
    \caption{The hyperparameters we searched over for DirectGA}
    \label{tab:hps_lists_directga}
    \begin{adjustbox}{width=1\linewidth}
        \input{994_HP_DirectGA.tex}
    \end{adjustbox}
\end{table*}

For PCGRL, we only searched over one variable for \mario; namely, how much to weigh how far the agent progressed in the level. We specifically searched over the values $[0, 2, 5, 8, 10]$ and found that the default value of $5$ performed the best.

The following sections contain some more descriptions of the baselines, as well as the actual hyperparameters.
\subsection*{DirectGA}
For the DirectGA, we used 2 point crossover, flattening the 2D array in the \maze case, and 1 point crossover for \mario with roulette selection and a mutation probability of $20\%$. We used these, as they were unspecified by \citet{ferreira_mario}, simple to implement, relatively standard~\citep{goldberg1989genetic} and the crossover operations were shown to perform relatively well~\citep{syswerda1993simulated, spears1995adapting}.

We also make the fitness function simply the inverse of the absolute difference between the actual and desired levels of entropy, ensuring that values do not become larger than 10. This is the fitness we then maximise.

The specific hyperparameters used are shown in \Autoref{tab:directga_hps_maze, tab:directga_hps_mario}. For these tables, \textit{DE} is Desired Entropy and \textit{PSolvability} is Partial Solvability.
\begin{table}[H]
    \centerfloat
    \caption{\maze hyperparameters for DirectGA}
    \label{tab:directga_hps_maze}
    \begin{adjustbox}{width=1\linewidth}
        \input{998_tab_src_results_v400_hyperparams_maze_DirectGA_Combined_hps_acm.tex}
    \end{adjustbox}
\end{table}
For the DirectGA with novelty on \maze, we additionally use Visual Diversity, $\lambda=1$ and 15 neighbours, with a weight of $0.33$.
\begin{table}[h]
    \centerfloat
    \caption{\mario hyperparameters for DirectGA. Explanations of these parameters are provided by \citet{ferreira_mario}.}
    \label{tab:directga_hps_mario}
    \begin{adjustbox}{width=1\linewidth}
        \input{998_tab_src_results_v400_hyperparams_mario_DirectGA_Combined_hps_acm.tex}
    \end{adjustbox}

\end{table}

For the DirectGA with novelty on \mario, we additionally use Visual Diversity, 6 neighbours in the novelty calculation and $\lambda = 1$. This was done for each individual element (ground, enemies, coins, etc.) and novelty had the same weight as the normal fitness function.
\subsection*{PCGRL}

For PCGRL, as previously mentioned, we trained for 100 million timesteps for \maze, and 8 million and 12 million for \mario Turtle and Wide respectively. 

The reward function used for the \maze was similar to the one used by \citet{pcgrl} in that we reward having only one region, but we differ in that we reward paths only between the start and goal tiles, with lengths between 20 and 80. We weight these different effects with a ratio of 5:2.

For \mario, the exact reward function we use can be found in the PCGRL Github repository,\footnote{\url{https://github.com/amidos2006/gym-pcgrl/blob/master/gym_pcgrl/envs/probs/smb_prob.py}} 
and this rewards levels that have the following characteristics:
\begin{itemize}
    \item Enemies are directly above solid tiles and are not floating in the air.
    \item Tubes are not disjoint.
    \item There are between 10 and 30 enemies per level.
    \item At least 60\% of the level is empty.
    \item The level has minimal noise, i.e. tiles of the same type are usually grouped together.
    \item The level requires at least 20 jumps to solve.
    \item The distance required to jump is low.
    \item The level is solvable.
\end{itemize}

\subsection*{PCGNN}

\begin{table*}[!htbp]
    \centerfloat
    \caption{Hyperparameters for PCGNN}
    \label{tab:novneat_hyperparams}
    \begin{adjustbox}{width=1\linewidth}
        \input{998_tab_src_results_v400_hyperparams_all_NoveltyNEAT_hps_acm.tex}
    \end{adjustbox}

\end{table*}
The hyperparameters for PCGNN are shown in \autoref{tab:novneat_hyperparams}.

\subsection*{Examples}
\begin{figure*}
    \centerfloat
    \begin{subfigure}[t]{0.2\linewidth}
        \fbox{\includegraphics[width=1\linewidth]{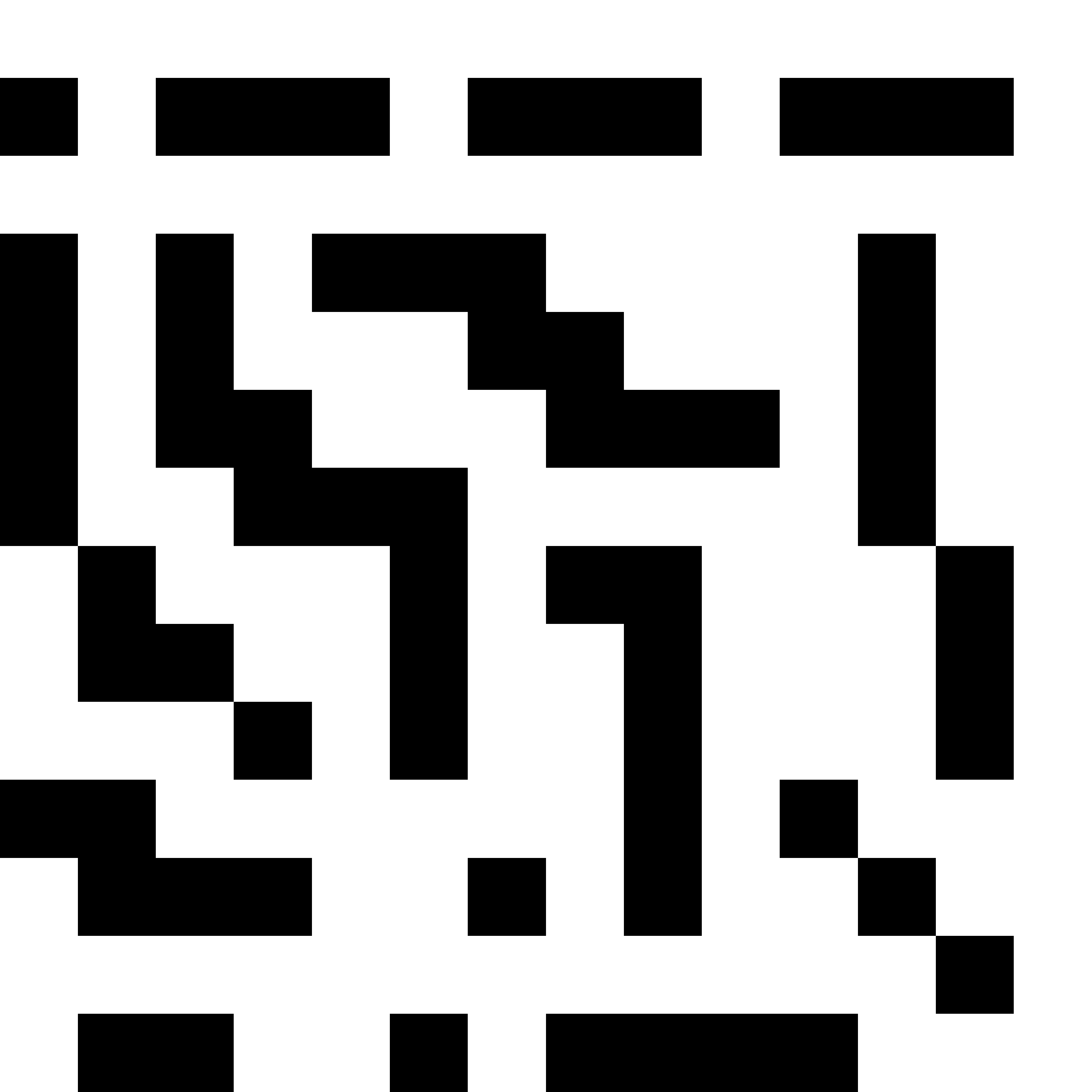}}
        \subcaption{PCGNN (Ours)}
    \end{subfigure}
    \begin{subfigure}[t]{0.2\linewidth}
        \fbox{\includegraphics[width=1\linewidth]{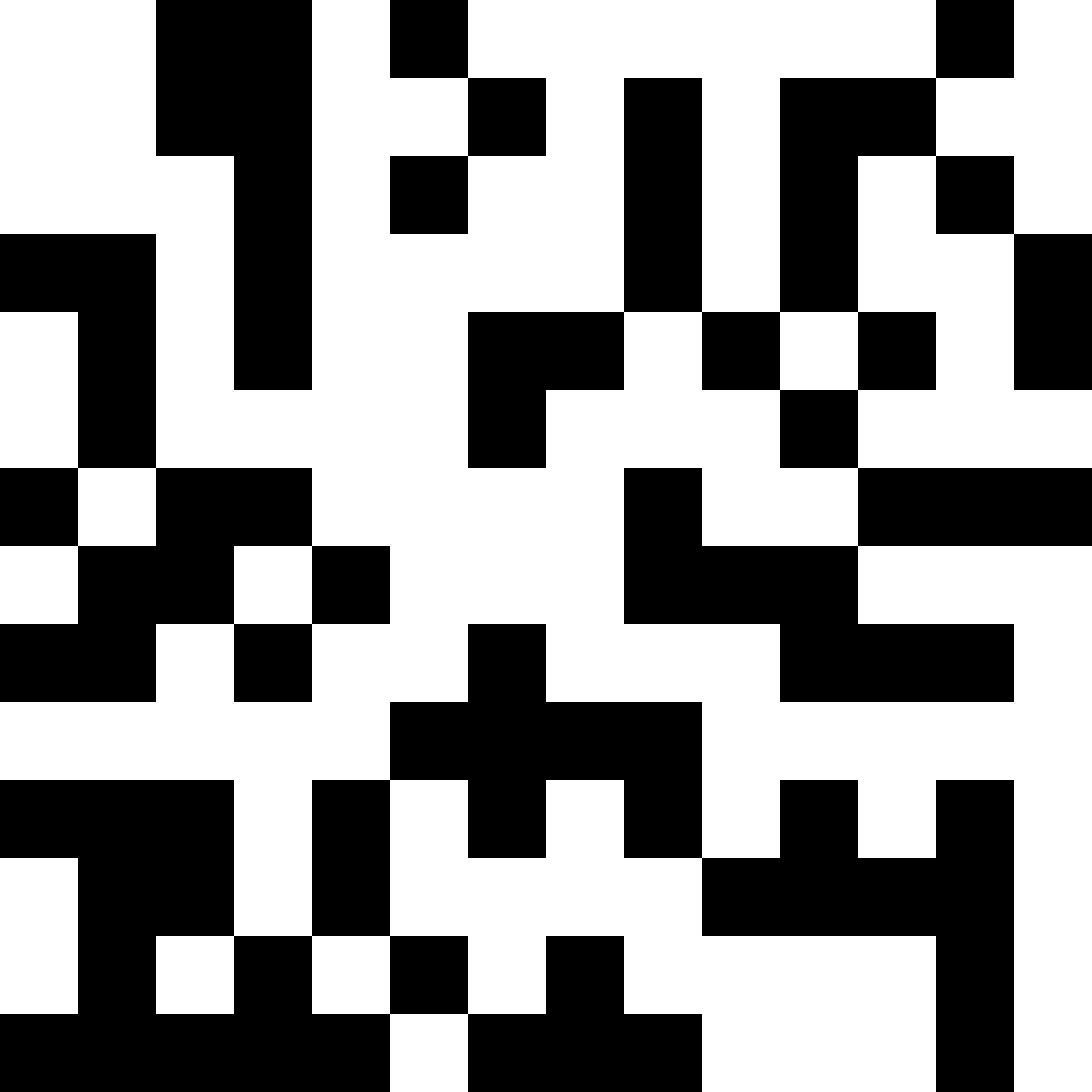}}
        \subcaption{DirectGA+}
    \end{subfigure}
    \begin{subfigure}[t]{0.2\linewidth}
        \fbox{\includegraphics[width=1\linewidth]{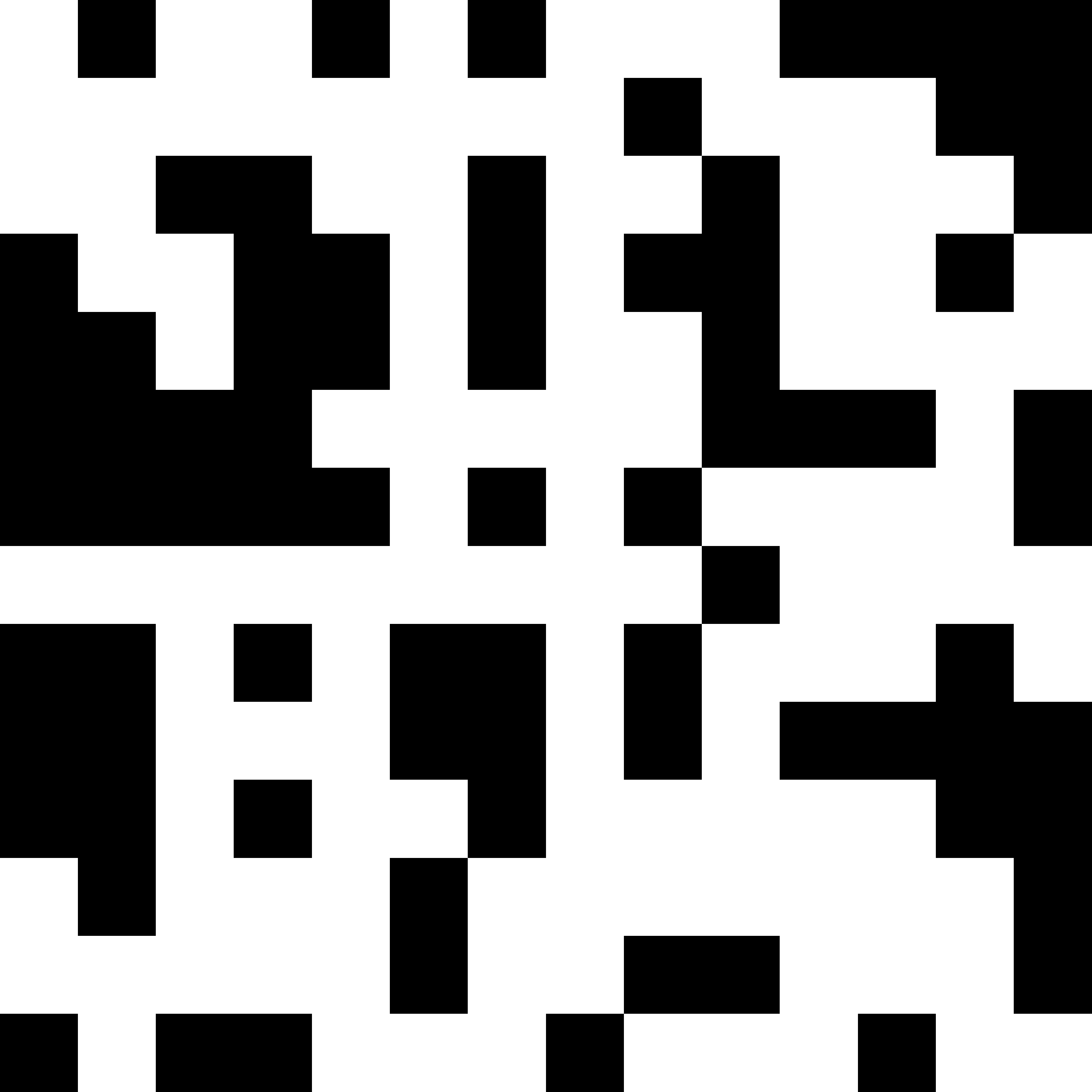}}
        \subcaption{PCGRL (Turtle)}
    \end{subfigure}
        \begin{subfigure}[t]{0.2\linewidth}
        \fbox{\includegraphics[width=1\linewidth]{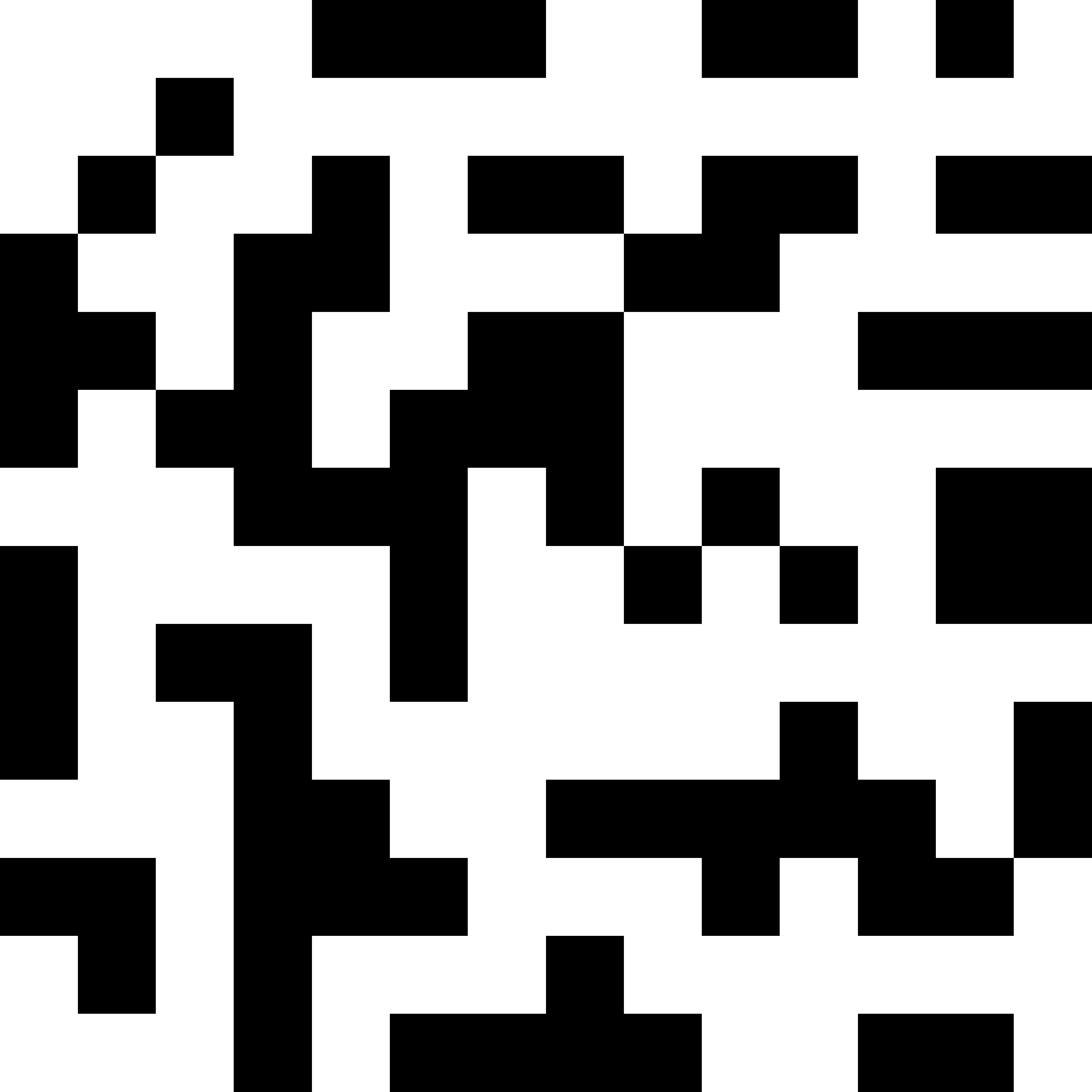}}
        \subcaption{PCGRL (Wide)}
    \end{subfigure}
    \begin{subfigure}[t]{0.2\linewidth}
        \fbox{\includegraphics[width=1\linewidth]{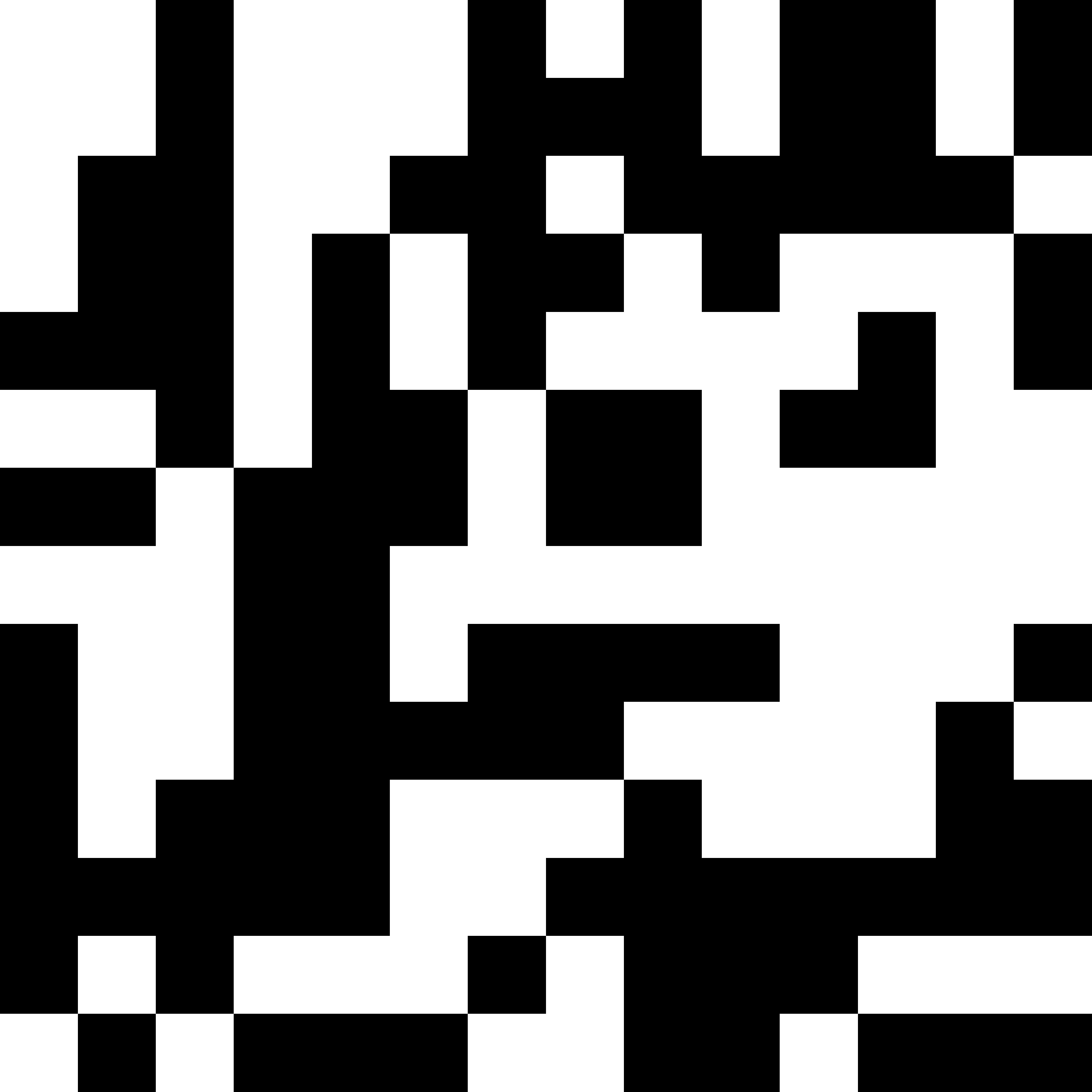}}
        \subcaption{Random}
    \end{subfigure}
    \caption{\maze: Example levels.}
    \label{fig:maze_example_levels}
    \Description[Example \maze levels for all baselines.]{Example \maze levels for all baselines.}
\end{figure*}

\begin{figure*}
    \centerfloat
    \begin{subfigure}[t]{1\linewidth}
        \includegraphics[width=1\linewidth]{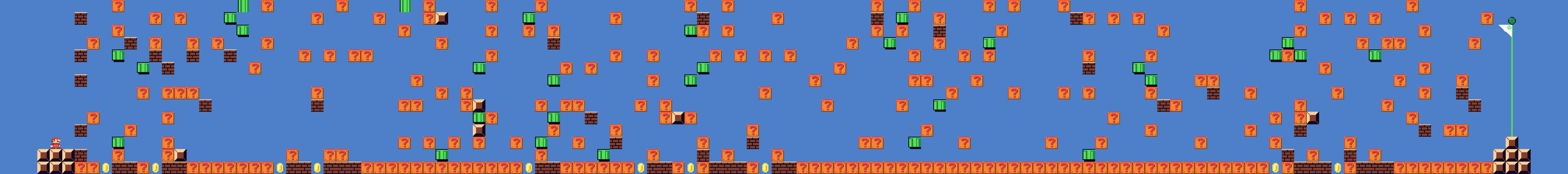}
        \subcaption{PCGNN (Ours)}
    \end{subfigure}
    \begin{subfigure}[t]{1\linewidth}
        \includegraphics[width=1\linewidth]{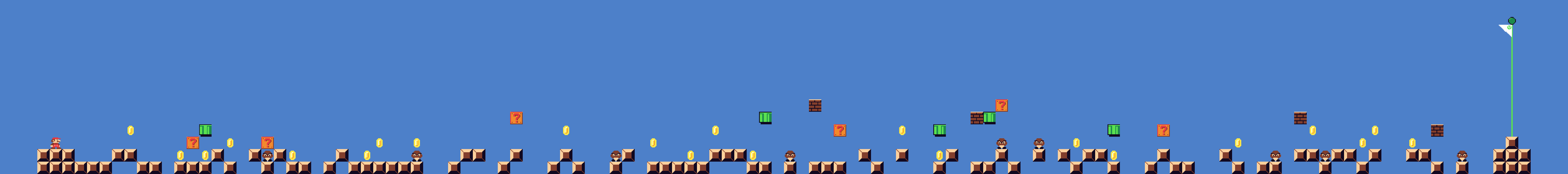}
        \subcaption{DirectGA+}
    \end{subfigure}
    \begin{subfigure}[t]{1\linewidth}
        \includegraphics[width=1\linewidth]{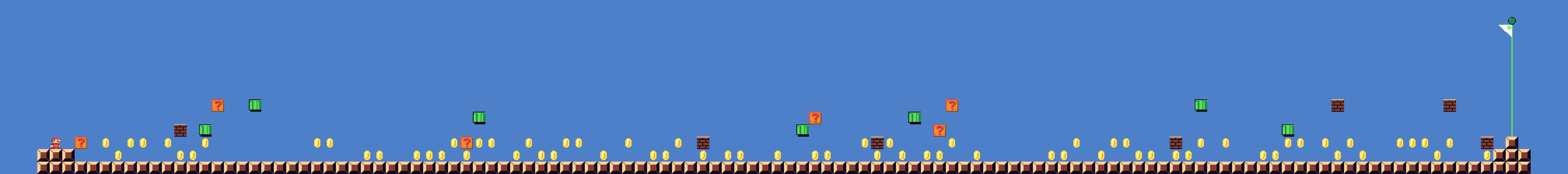}
        \subcaption{DirectGA}
        \label{fig:mario_eg_directga}
    \end{subfigure}
    \begin{subfigure}[t]{1\linewidth}
        \includegraphics[width=1\linewidth]{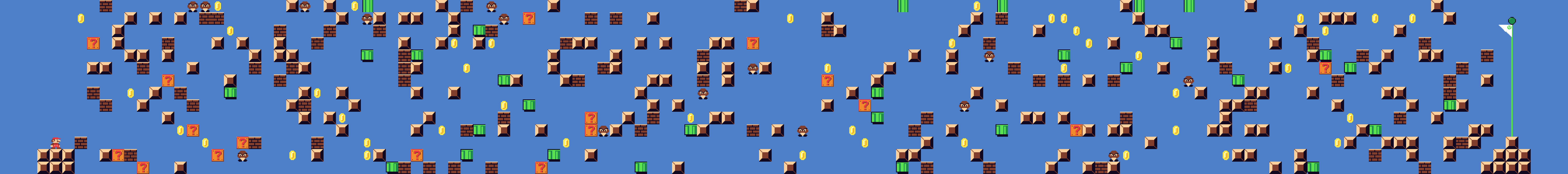}
        \subcaption{PCGRL (Turtle)}
    \end{subfigure}
        \begin{subfigure}[t]{1\linewidth}
        \includegraphics[width=1\linewidth]{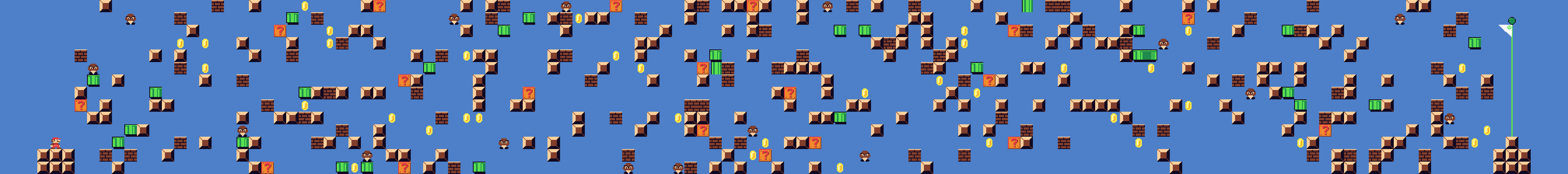}
        \subcaption{PCGRL (Wide)}
    \end{subfigure}
    \begin{subfigure}[t]{1\linewidth}
        \includegraphics[width=1\linewidth]{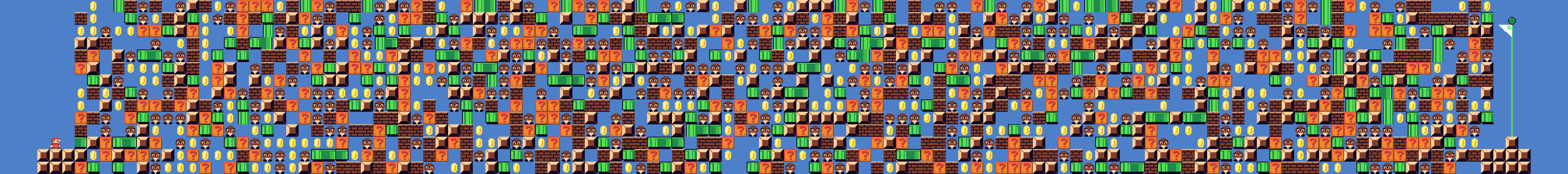}
        \subcaption{Random}
    \end{subfigure}
    \caption{\mario: Example levels.}
    \label{fig:mario_example_levels}
    \Description[Example \mario levels for all baselines.]{Example \mario levels for all baselines.}
\end{figure*}

Example levels from the tested methods are shown in \Autoref{fig:maze_example_levels,fig:mario_example_levels}.

\subsection*{Additional Experiments}
Here we show some additional results.
\subsubsection*{Novelty Distance Functions}
Firstly, there are many different distance functions we can use for the novelty fitness calculation. Some of the ones we tried are listed below. All image hashing methods used the ImageHash library.\footnote{\url{https://github.com/JohannesBuchner/imagehash}}
\begin{description}
\item[Euclidean                   ] The Euclidean distance between levels with $n$ unique tiles, where each tile is an integer from $0$ to $n-1$.
\item[Hashing (Perceptual Simple) ] Simple Perceptual Image Hashing.
\item[Hashing (Average)           ] Average Image Hashing.

\item[Visual Diversity            ] Hamming distance, i.e. fraction of tiles that are different.

\item[Hashing (Perceptual)        ] Perceptual Image Hashing.

\item[Hashing (Wavelet)           ] Wavelet Image Hashing.
\end{description}
Some distance functions are only applicable for the \maze, as some aspects (like pathfinding) were much faster on this domain, making these functions feasible.
\begin{description}
\item[Visual Diversity Reachable  ] Visual Diversity, but all tiles that are not reachable from the starting tile are set to a wall.
\item[JS                          ] This takes the coordinates of each reachable tile, and creates a probability distribution for each level. The final distance is then the Jensen-Shannon divergence between these two distributions.
\item[Path                        ] Comparing the shortest paths from start to end in each level. The paths are compared using the average Manhattan distance between corresponding steps in the trajectories.
\item[Window                      ] This considers all reachable tiles in the level, and sets the distance to the average visual diversity of $3\times 3$ blocks centred at each location.
\item[Window (V2)                 ] This is similar to the above, but instead of taking the reachable tiles, uses only the shortest path as a trajectory.
\end{description}

Results when using different distance functions for \maze and \mario are shown in \Autoref{tab:maze_distance_funcs,tab:mario_distance_funcs} respectively. Example levels for each of these functions are also shown in \Autoref{fig:fig_all_dist_maze,fig:fig_all_dist_mario}.
\begin{table*}[hb!]
    \centerfloat
    \caption{\maze: Comparing a collection of different distance functions for the novelty and intra-novelty fitness functions. \textcolor{blue}{Blue} denotes the maximum and \textcolor{darkgreen}{green} denotes the minimum per column. This table shows $\mu (\sigma)$. The generation time is also not dependent on the distance function -- so even if the fitness takes a long time to compute, generation is still fast.}
    \label{tab:maze_distance_funcs}
    \begin{adjustbox}{width=1\linewidth}
        \input{997_Maze_Distance_Functions.tex}

    \end{adjustbox}
\end{table*}

\begin{table*}[hb!]
    \centerfloat
    \caption{\mario: Comparing a collection of different distance functions for the novelty and intra-novelty fitness functions. \textcolor{blue}{Blue} denotes the maximum and \textcolor{darkgreen}{green} denotes the minimum per column. This table shows $\mu (\sigma)$. The generation time is also not dependent on the distance function -- so even if the fitness takes a long time to compute, generation is still fast. The A* diversity metric value for Visual Diversity is slightly different from that shown in the main text due to some stochasticity in the behaviour of the A* agent.}
    \label{tab:mario_distance_funcs}
    \begin{adjustbox}{width=1\linewidth}
        \input{997_Mario_Distance_Functions.tex}

    \end{adjustbox}
\end{table*}

\newcommand{\wwww}{0.24\linewidth}
\begin{figure*}[h]
    \centering

\begin{subfigure}[t]{\wwww}
        \includegraphics[width=1\linewidth]{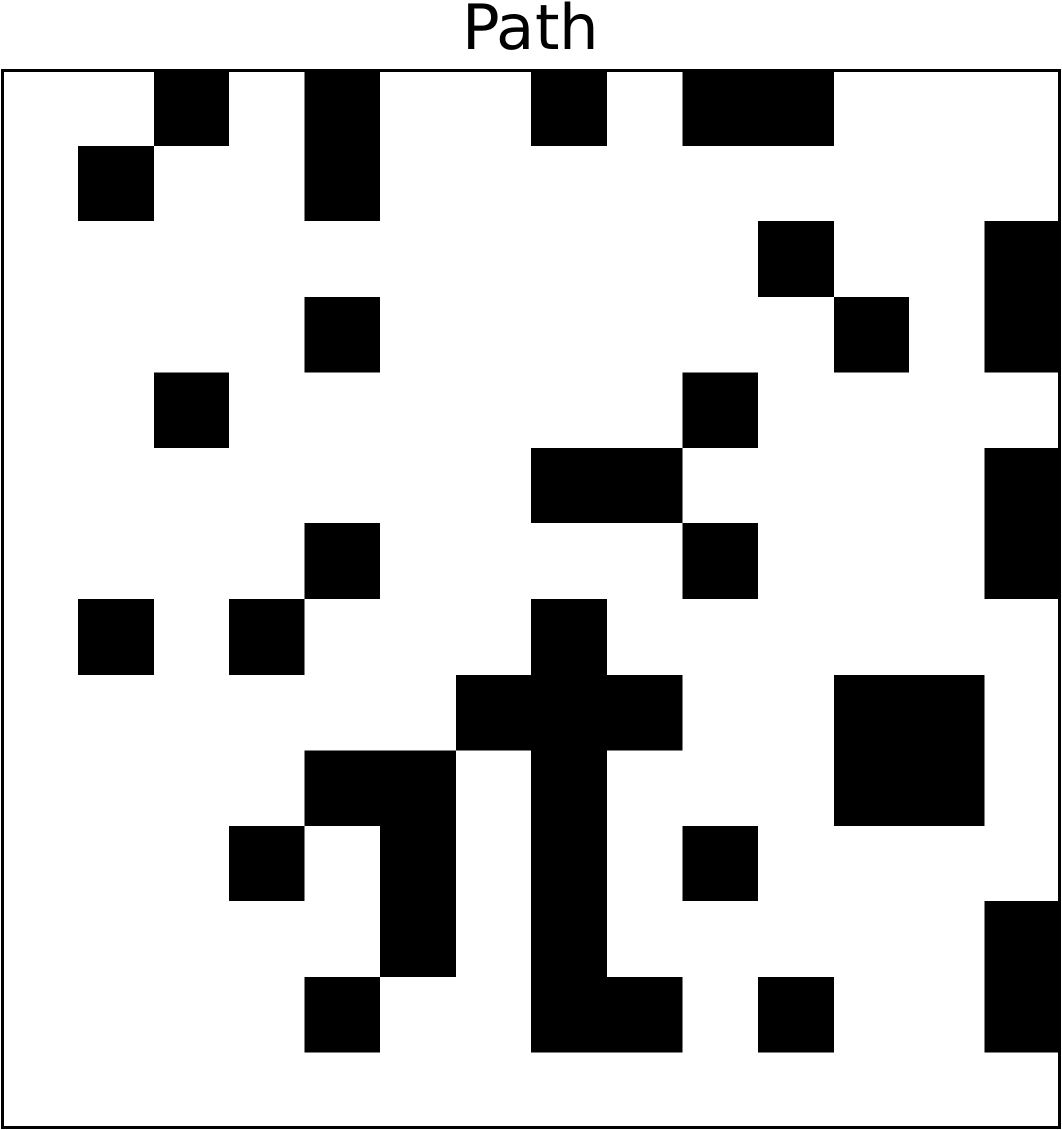}
    \end{subfigure}
\begin{subfigure}[t]{\wwww}
        \includegraphics[width=1\linewidth]{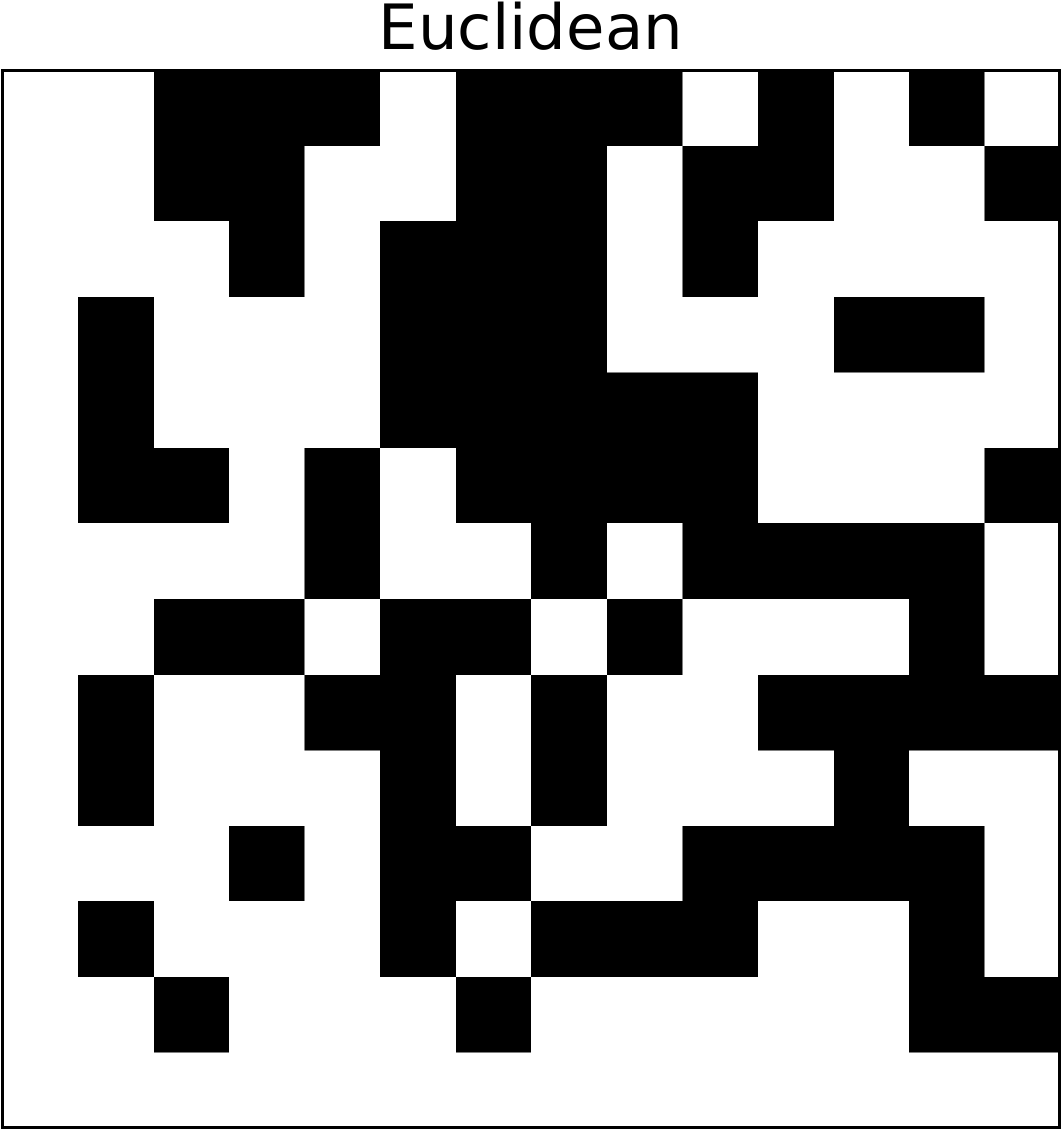}
    \end{subfigure}
\begin{subfigure}[t]{\wwww}
        \includegraphics[width=1\linewidth]{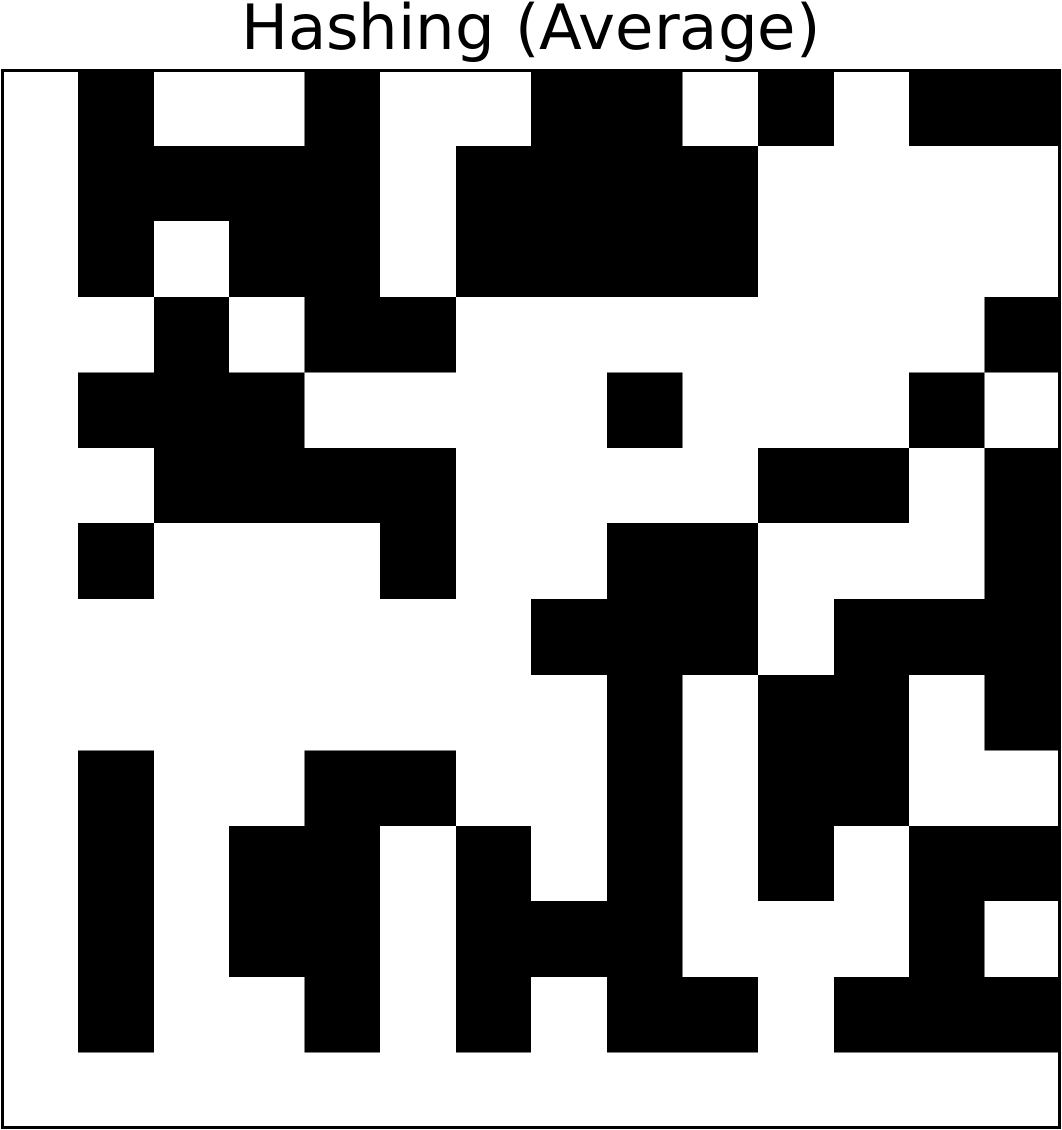}
    \end{subfigure}
\begin{subfigure}[t]{\wwww}
        \includegraphics[width=1\linewidth]{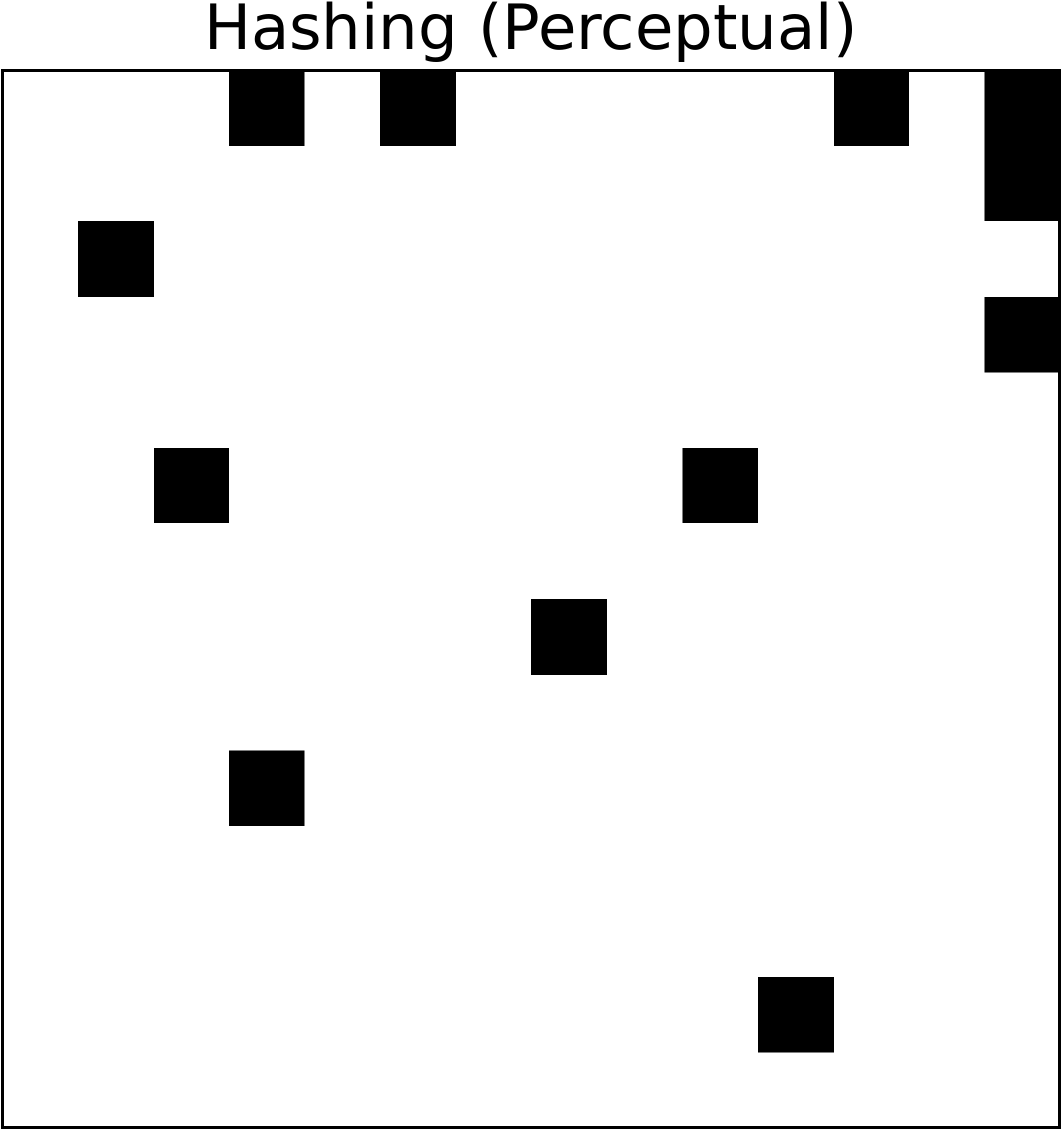}
    \end{subfigure}

\begin{subfigure}[t]{\wwww}
        \includegraphics[width=1\linewidth]{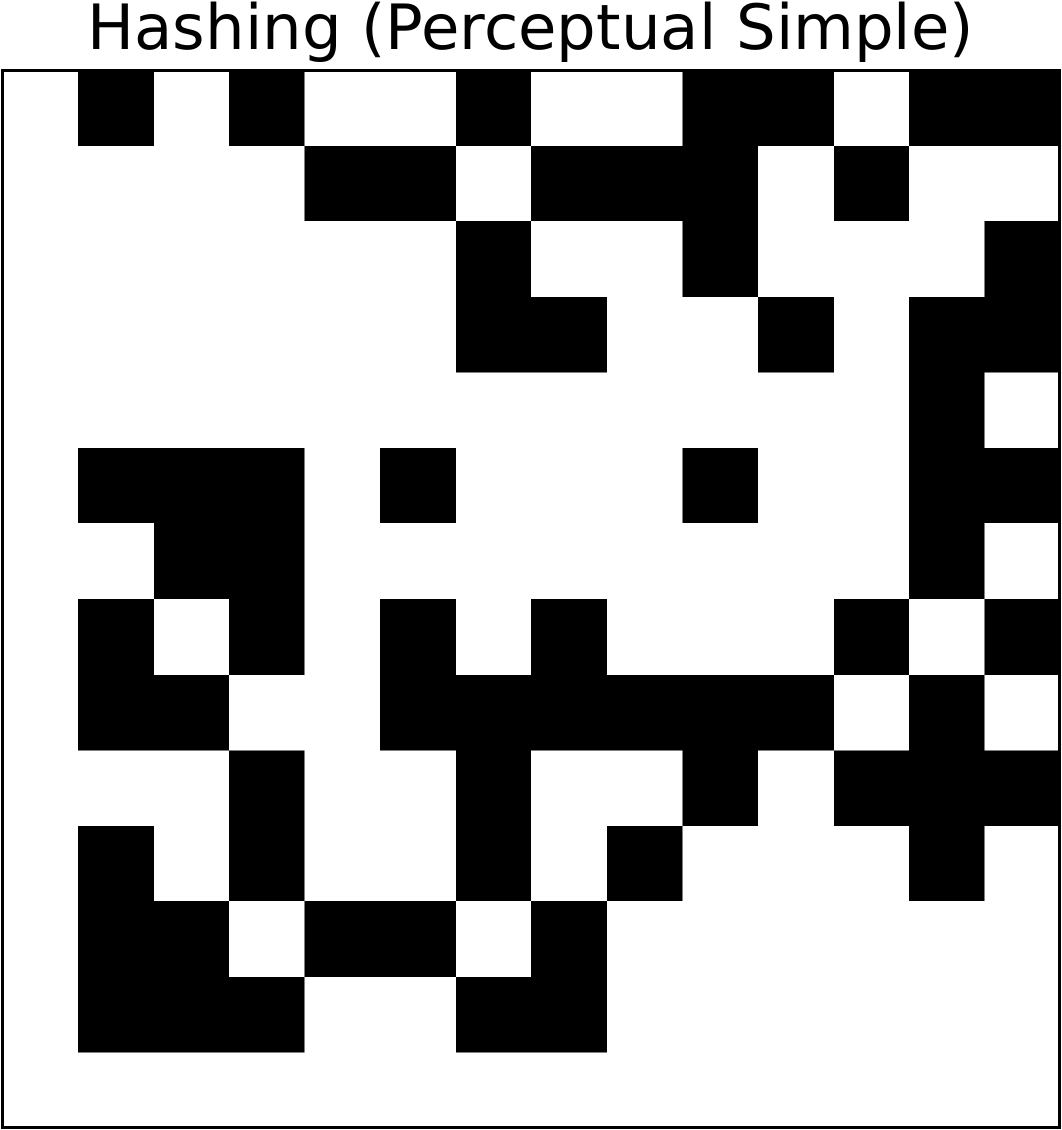}
    \end{subfigure}
\begin{subfigure}[t]{\wwww}
        \includegraphics[width=1\linewidth]{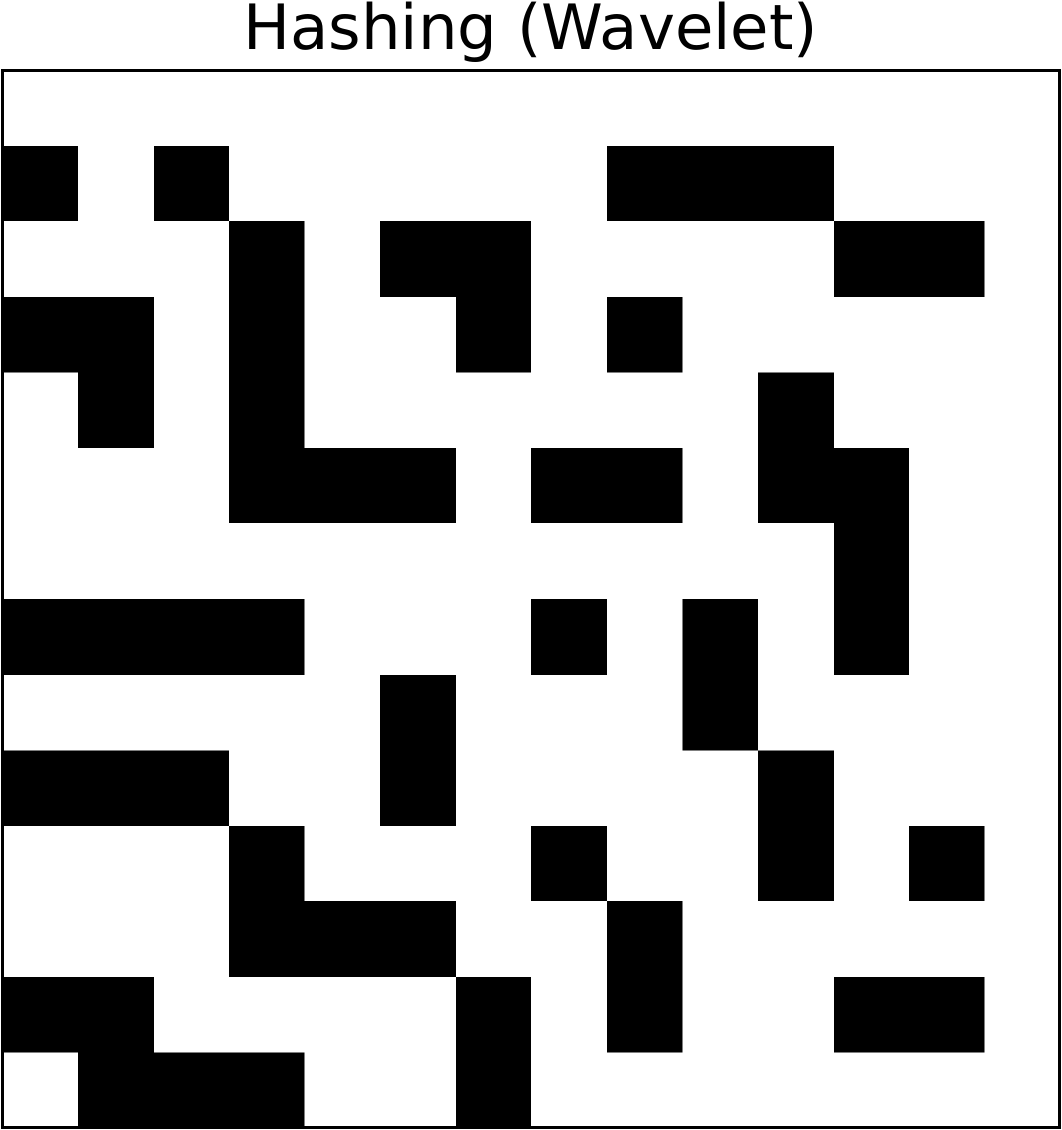}
    \end{subfigure}
\begin{subfigure}[t]{\wwww}
        \includegraphics[width=1\linewidth]{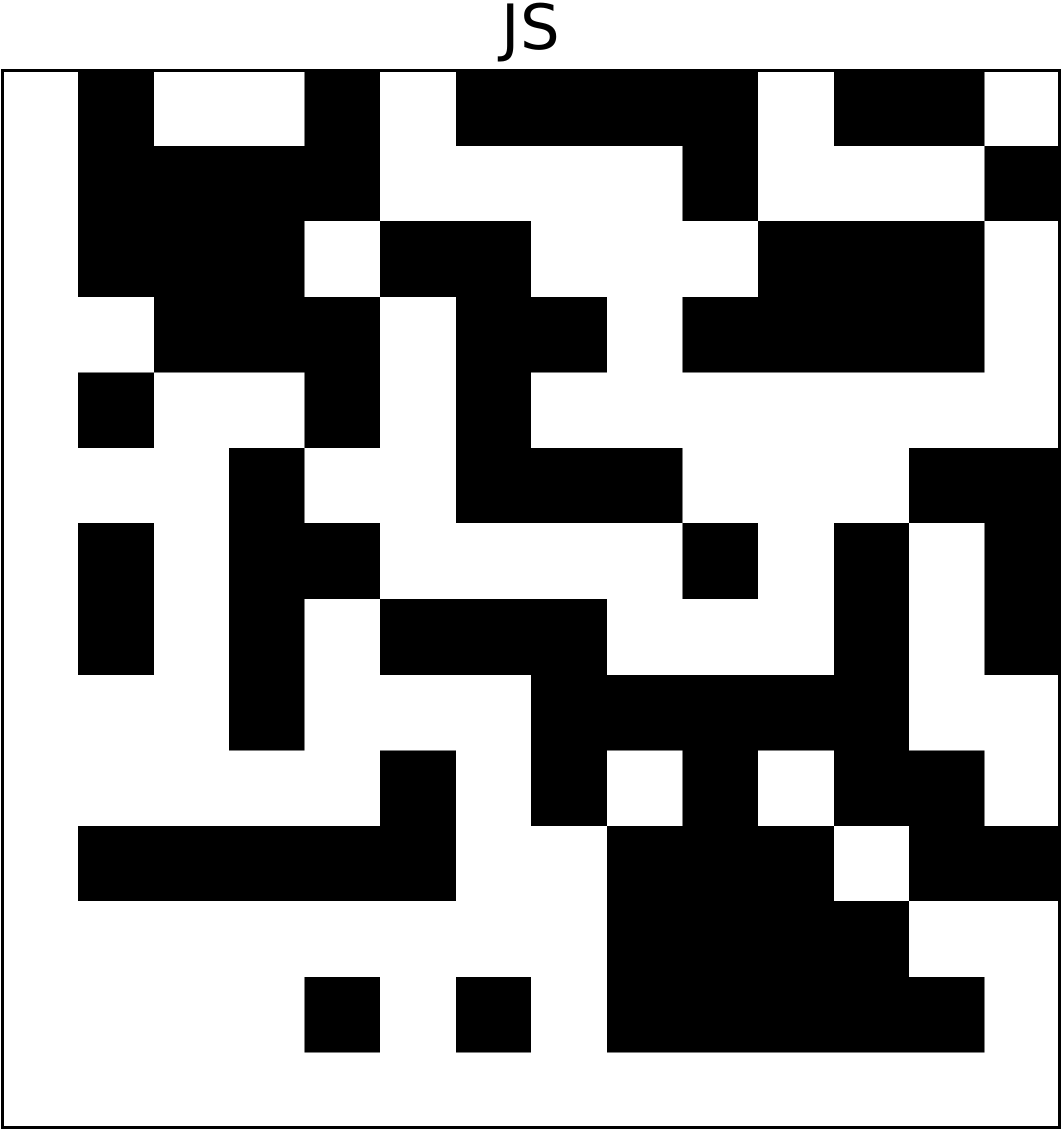}
    \end{subfigure}
\begin{subfigure}[t]{\wwww}
        \includegraphics[width=1\linewidth]{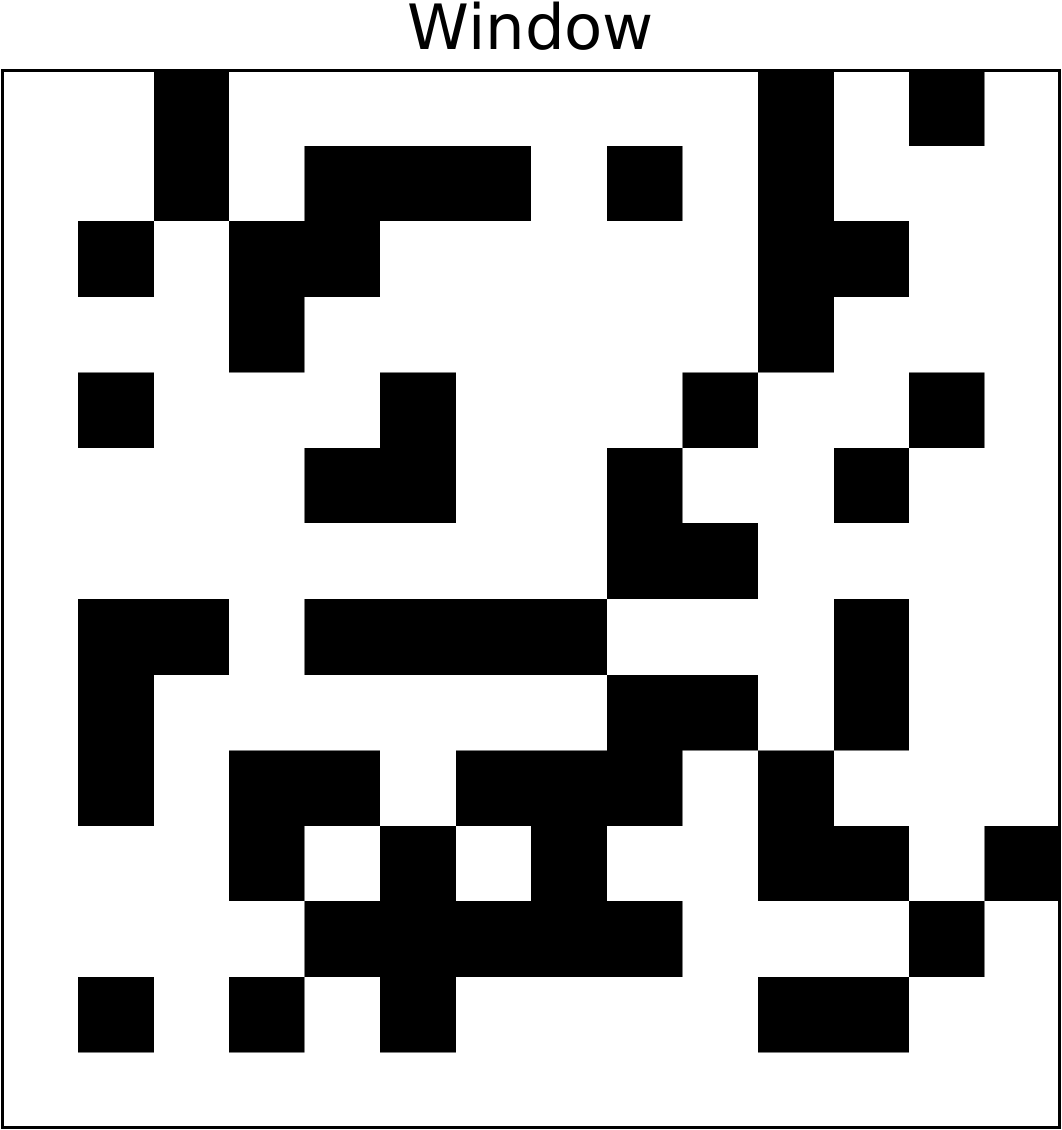}
    \end{subfigure}

\begin{subfigure}[t]{\wwww}
        \includegraphics[width=1\linewidth]{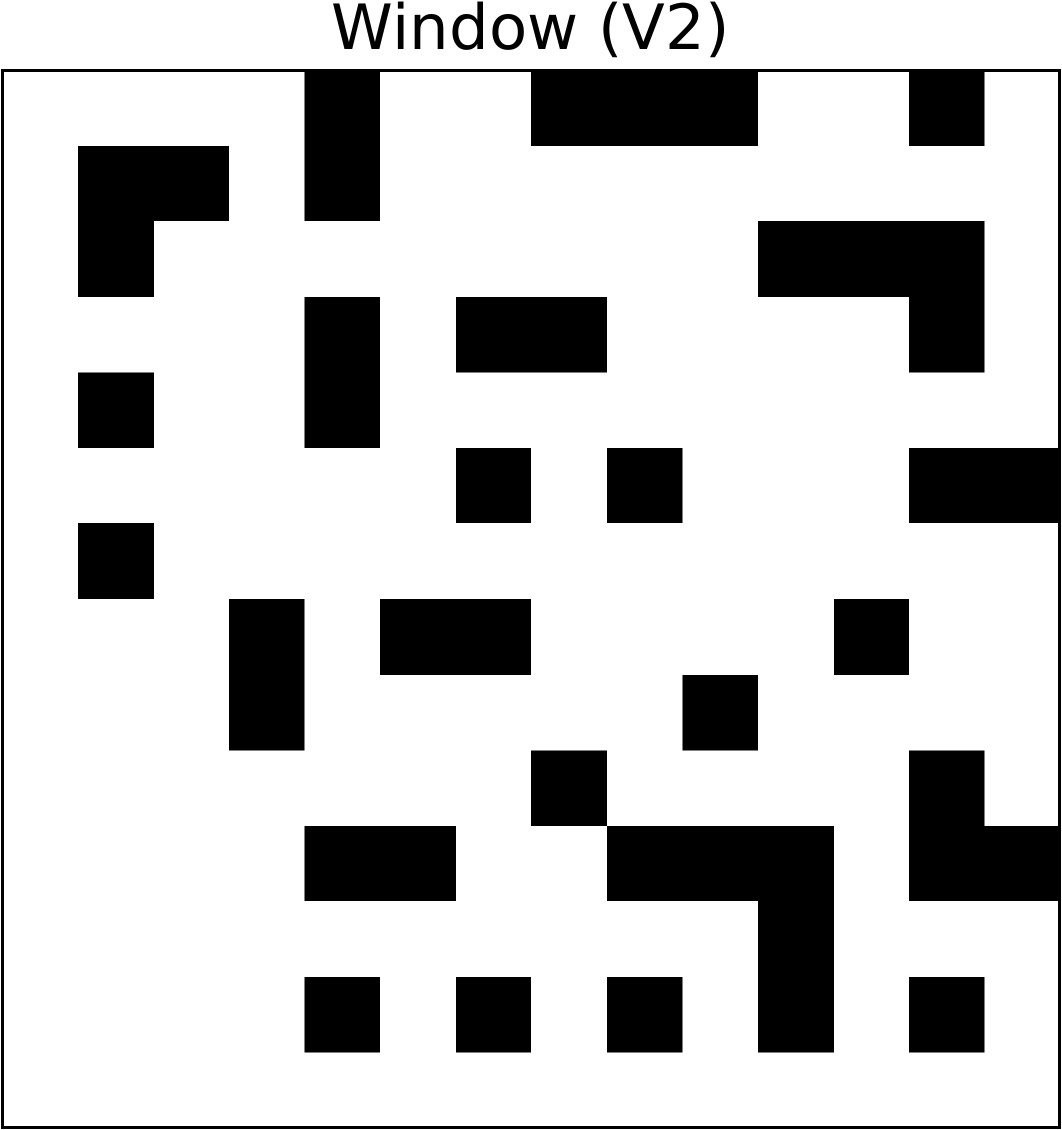}
    \end{subfigure}
\begin{subfigure}[t]{\wwww}
        \includegraphics[width=1\linewidth]{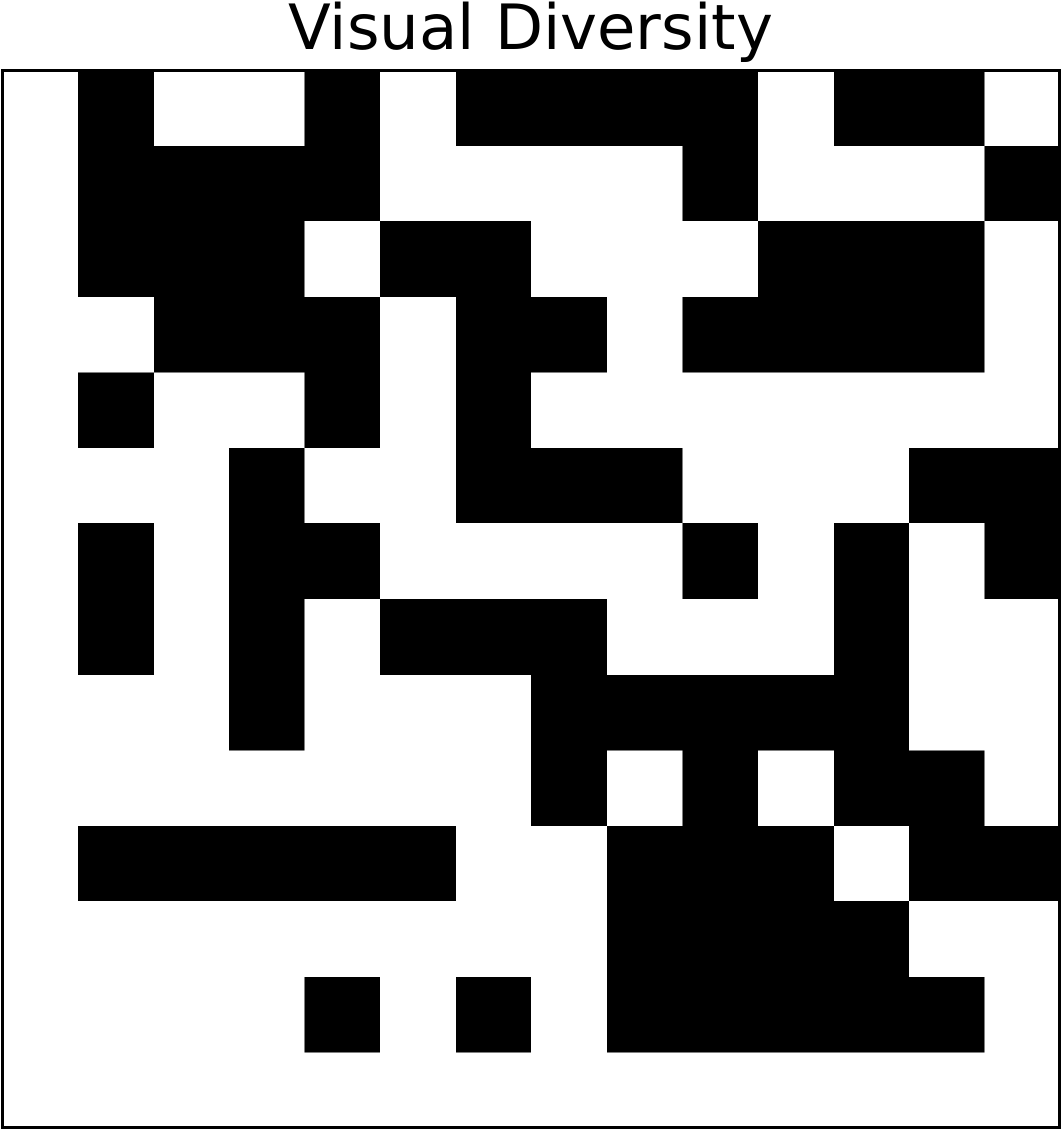} 
    \end{subfigure}
\begin{subfigure}[t]{\wwww}
        \includegraphics[width=1\linewidth]{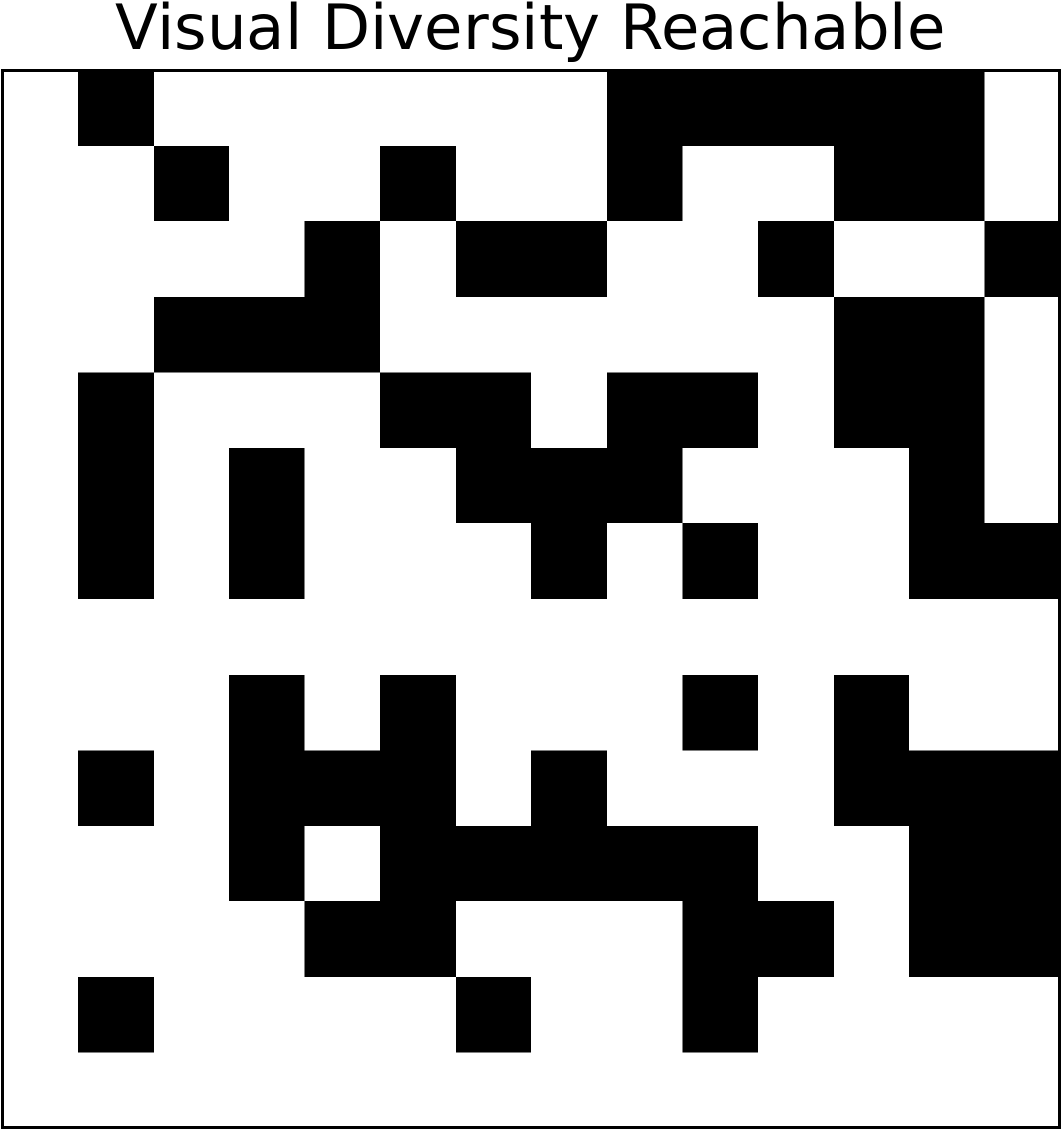}
    \end{subfigure}
    \caption{Showcasing sample \maze levels for different novelty distance functions.}
    \label{fig:fig_all_dist_maze}

    \Description[\maze levels for different novelty distance functions.]{\maze levels for different novelty distance functions.}
\end{figure*}

\newcommand{\wwwwm}{1\linewidth}
\begin{figure*}[h]
    \centering

\begin{subfigure}[t]{\wwwwm}
        \includegraphics[width=1\linewidth]{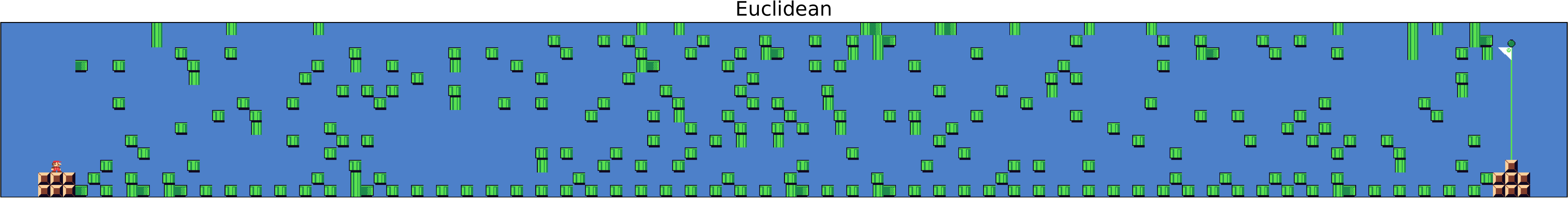}
    \end{subfigure}
\begin{subfigure}[t]{\wwwwm}
        \includegraphics[width=1\linewidth]{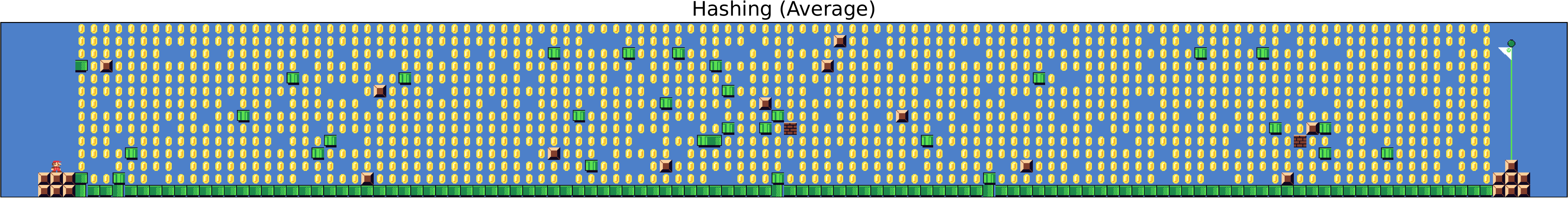}
    \end{subfigure}
\begin{subfigure}[t]{\wwwwm}
        \includegraphics[width=1\linewidth]{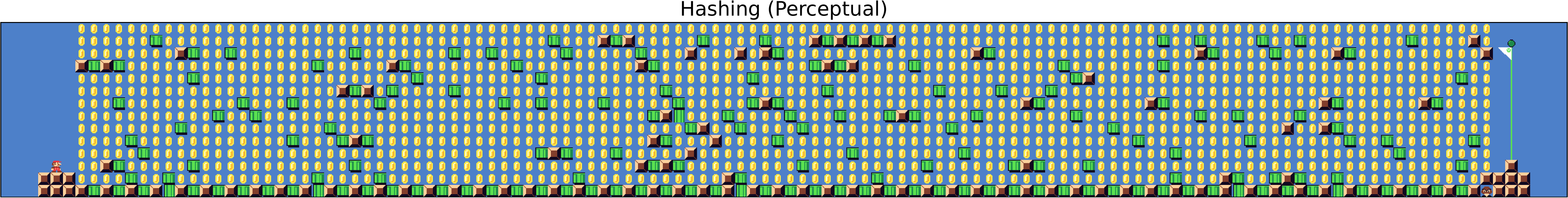}
    \end{subfigure}

\begin{subfigure}[t]{\wwwwm}
        \includegraphics[width=1\linewidth]{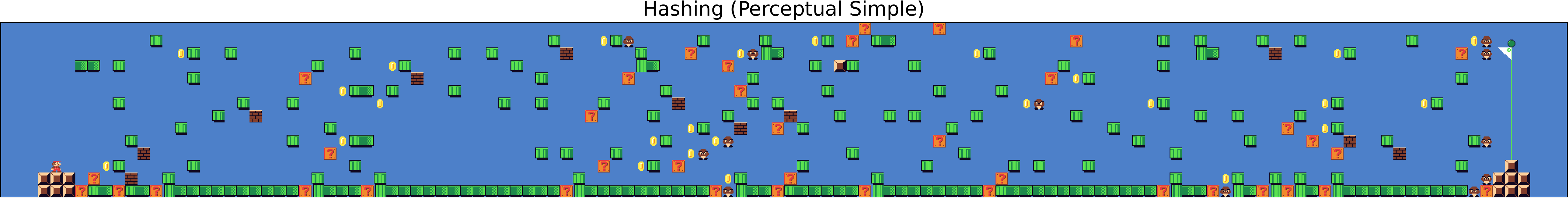}
    \end{subfigure}
\begin{subfigure}[t]{\wwwwm}
        \includegraphics[width=1\linewidth]{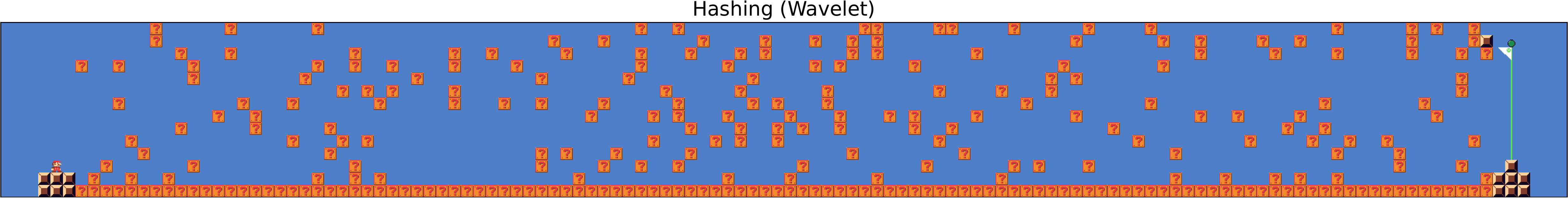}
    \end{subfigure}

\begin{subfigure}[t]{\wwwwm}
        \includegraphics[width=1\linewidth]{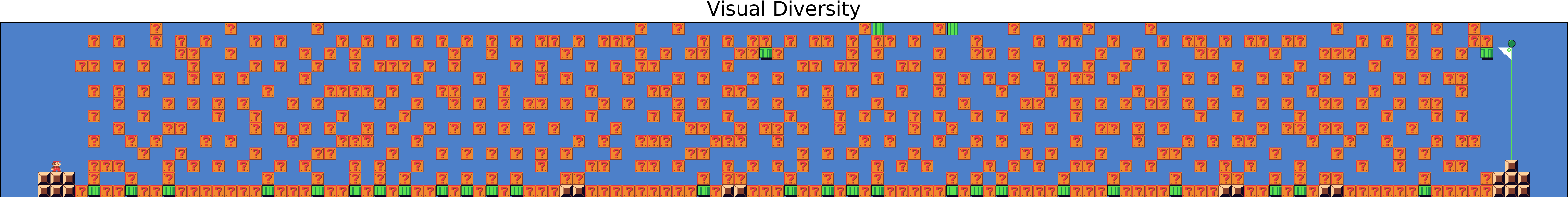}
    \end{subfigure}
    \caption{Showcasing sample levels for different novelty distance functions for \mario.}
    \label{fig:fig_all_dist_mario}
    \Description[\mario levels for different novelty distance functions.]{\mario levels for different novelty distance functions.}
\end{figure*}

\subsubsection*{PCGNN Fitness Plots}
\begin{figure*}
    \begin{subfigure}[t]{1\linewidth}
        \includegraphics[width=1\linewidth]{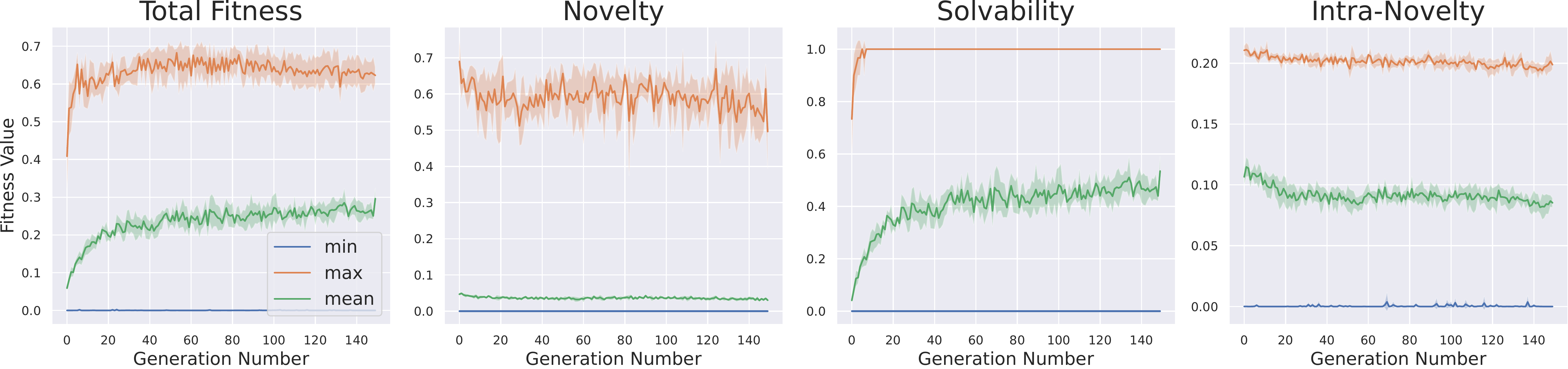}
        \caption{\mario}
    \end{subfigure}
    
    \begin{subfigure}[t]{1\linewidth}
        \includegraphics[width=1\linewidth]{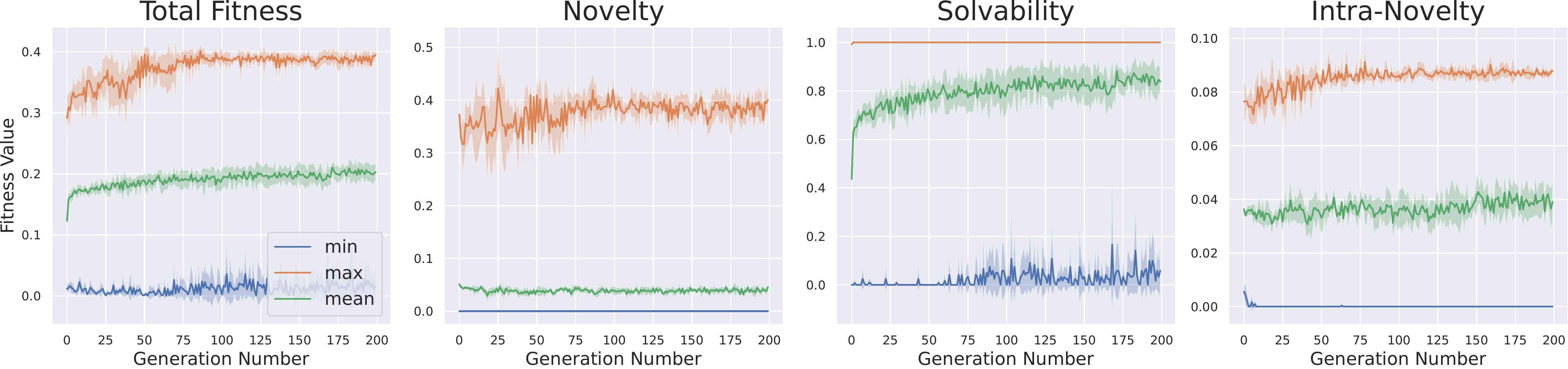}
        \caption{\maze}
    \end{subfigure}
    
    \caption{Fitness as the training (evolution) process progresses. Standard deviation over five seeds is shaded. For each plot we show the fitness of the best individual (orange), the worst individual (blue), and the average over the entire population (green). Note that the novelty fitness function is relative to the population, and that the intra-novelty fitness is similarly relative to each individual. So, the same novelty fitness at two different generations may not be equivalent, as the populations differ between these generations.}
    \label{fig:fig_all_fitness_plots}
    \Description[\pcgnn fitness plots as evolution progresses.]{For both games, the total fitness increases as more generations pass. The novelty fitnesses are more flat, and do not increase as much as the solvability fitness.}
\end{figure*}

\autoref{fig:fig_all_fitness_plots} shows the different fitness functions and their value over the evolution process for \pcgnn.

%% file: 994_HP_PCGNN.tex
\begin{tabular}{p{0.3\linewidth}p{0.35\linewidth}p{0.35\linewidth}}
\toprule
{} &                                                                                                                                   Maze &                 Mario \\
\midrule
Context Size              &                                                                                                                                      1 &                     1 \\
Predict Size              &                                                                                                                                      1 &               1, 2, 3 \\
Number of Random Variables              &                                                                                                                                      4 &               4 \\
Random Perturb Size       &                                                                                                                                 [0, 1] &                     0 \\
Number of Generations     &                                                                                                                            10, 50, 100, 200 &  10, 20, 50, 150, 300 \\
Population Size           &                                                                                                                            20, 50, 100 &           20, 50, 100 \\
Number of Levels          &                                                                                                                                 15, 24 &           3, 5, 6, 15 \\
$\lambda$                 &                                                                                                                    0, 1, 2, 3, 4, 5, 6 &                  0, 1 \\
One Hot Inputs            &                                                                                                                                  False &           False, True \\
Novelty Distance Function &  Hashing (Average), Hashing (Perceptual Simple), Hashing (Perceptual), Hashing (Wavelet), Visual Diversity, Visual Diversity Reachable &      Visual Diversity \\
Intra-Novelty Weight      &                                                                                                                                 [0, 1] &                     1 \\
Novelty Weight            &                                                                                                                                 [0, 1] &                     1 \\
Solvability Weight        &                                                                                                                                 [0, 1] &     1, 2, 3, 4, 6, 10 \\
Use Entropy Fitness       &                                                                                                                            False, True &                 False \\
Use Intra-Novelty Fitness &                                                                                                                            False, True &                  True \\
Use Solvability Fitness   &                                                                                                                            False, True &                  True \\
\bottomrule
\end{tabular}

%% file: 994_HP_DirectGA.tex
\begin{tabular}{p{0.3\linewidth}p{0.35\linewidth}p{0.35\linewidth}}
\toprule
{} &                                    Mario &         Maze \\
\midrule
Population Size            &                              10, 50, 100 &  10, 50, 100 \\
Number of Generations      &                              10, 50, 100 &  10, 50, 100 \\
Desired Entropy            &                              0, 0.5, 1.0 &    0, 0.5, 1 \\
Desired Sparseness Enemies &        $\langle$Desired Entropy$\rangle$ &            - \\
Desired Sparseness Coins   &        $\langle$Desired Entropy$\rangle$ &            - \\
Desired Sparseness Blocks  &        $\langle$Desired Entropy$\rangle$ &            - \\
Entropy Block Size         &                                  20, 114 &            - \\
Enemies Block Size         &                                       20 &            - \\
Coin Block Size           &                                       10 &            - \\
Blocks Block Size          &                                   10, 40 &            - \\
Ground Maximum Height      &                                     2, 5 &            - \\
Coin Maximum Height        &  $\langle$Ground Maximum Height$\rangle$ &            - \\
Use Novelty Fitness        &                                     True &  False, True \\
Use Solvability Fitness    &                                    False &  False, True \\
\bottomrule
\end{tabular}

%% file: 998_tab_src_results_v400_hyperparams_maze_DirectGA_Combined_hps_acm.tex
\begin{tabular}{p{0.3\linewidth}p{0.35\linewidth}p{0.35\linewidth}}
\toprule
{} &                                                                                                      DirectGA+ &                                                                                                  DirectGA Novelty \\
\midrule
Population Size       &                                                                                                            100 &                                                                                                                50 \\
Number of Generations &                                                                                                            100 &                                                                                                               100 \\
Fitness               &  \parbox{1\linewidth}{Entropy(DE=1) $\times$ 0.5\\PSolvability() $\times$ 0.5} &  \parbox{1\linewidth}{Entropy(DE=0) $\times$ 0.33 \\PSolvability() $\times$ 0.33} \\
\bottomrule
\end{tabular}

%% file: 998_tab_src_results_v400_hyperparams_mario_DirectGA_Combined_hps_acm.tex
\begin{tabular}{p{0.4\linewidth}p{0.2\linewidth}p{0.2\linewidth}p{0.2\linewidth}}
\toprule
{} & DirectGA+ & DirectGA Novelty & DirectGA \\
\midrule
Population Size            &        10 &              100 &       20 \\
Number of Generations      &        50 &              100 &      100 \\
Desired Entropy            &       0.5 &              0.0 &      0.0 \\
Desired Sparseness Enemies &       0.5 &              0.0 &      0.0 \\
Desired Sparseness Coins   &       0.5 &              1.0 &      1.0 \\
Desired Sparseness Blocks  &       0.5 &              0.5 &      0.5 \\
Entropy Block Size         &        20 &              114 &      114 \\
Enemies Block Size         &        20 &               20 &       20 \\
Coin Block Size           &        10 &               10 &       10 \\
Blocks Block Size          &        40 &               10 &       10 \\
Ground Maximum Height      &         2 &                2 &        2 \\
Coin Maximum Height      &         2 &                2 &        2 \\
\bottomrule
\end{tabular}

%% file: 998_tab_src_results_v400_hyperparams_all_NoveltyNEAT_hps_acm.tex
\begin{tabular}{p{0.3\linewidth}p{0.35\linewidth}p{0.35\linewidth}}
\toprule
{} &                                                                                                                                                                Maze &                                                                                                                                                             Mario \\
\midrule
Context Size                                 &                                                                                                                                                                   1 &                                                                                                                                                                 1 \\
Predict Size                                 &                                                                                                                                                                   1 &                                                                                                                                                                 1 \\
Number of Random Variables                  &                                                                                                                                                                   4 &                                                                                                                                                                 4 \\
Padding                                      &                                                                                                                                                                  -1 &                                                                                                                                                                -1 \\
Random Perturb Size                          &                                                                                                                                                              0.1565 &                                                                                                                                                               0.0 \\
Number of Generations                        &                                                                                                                                                                 200 &                                                                                                                                                               150 \\
Population Size                              &                                                                                                                                                                  50 &                                                                                                                                                               100 \\
Number of Levels                             &                                                                                                                                                                  24 &                                                                                                                                                                 6 \\
Number of Neighbours for Novelty Calculation &                                                                                                                                                                  15 &                                                                                                                                                                15 \\
$\lambda$                                    &                                                                                                                                                                   0 &                                                                                                                                                                 0 \\
Novelty Distance Function                    &                                                                                                                           Visual Diversity, only on reachable tiles &                                                                                                                                                  Visual Diversity \\
Fitness                                      &  \parbox{1\linewidth}{Novelty() $\times$ 0.399\\Solvability() $\times$ 0.202\\\textit{Intra-Novelty}(neighbours=10) $\times$ 0.399} &  \parbox{1\linewidth}{Novelty() $\times$ 0.25 \\Solvability()$\times$ 0.50 \\\textit{Intra-Novelty}(neighbours=2) $\times$ 0.25} \\
\bottomrule
\end{tabular}

%% file: 997_Maze_Distance_Functions.tex
\begin{tabular}{llllllll}
\toprule
{} &                 Generation Time (s) &                   Train Time (s) &                         Solvability &                  Compression Distance &                    A* Diversity &                            Leniency &                     A* Difficulty \\
Distance Function           &                                     &                                  &                                     &                                       &                                 &                                     &                                   \\
\midrule
Path                        &  \textcolor{darkgreen}{0.002 (0.0)} &    \textcolor{blue}{14528 (384)} &           \textcolor{blue}{1.0 (0)} &                         0.460 (0.004) &                     0.001 (0.0) &                         0.86 (0.03) &  \textcolor{darkgreen}{0.0 (0.0)} \\
Window (V2)                 &  \textcolor{darkgreen}{0.002 (0.0)} &                      14331 (357) &  \textcolor{darkgreen}{0.95 (0.09)} &                         0.466 (0.014) &                     0.10 (0.20) &                         0.79 (0.11) &                       0.04 (0.08) \\
JS                          &       \textcolor{blue}{0.003 (0.0)} &                        3441 (50) &           \textcolor{blue}{1.0 (0)} &                         0.490 (0.007) &  \textcolor{darkgreen}{0.0 (0)} &  \textcolor{darkgreen}{0.62 (0.04)} &    \textcolor{darkgreen}{0.0 (0)} \\
Visual Diversity Reachable  &       \textcolor{blue}{0.003 (0.0)} &                        1107 (46) &           \textcolor{blue}{1.0 (0)} &                         0.488 (0.002) &                     0.13 (0.17) &                         0.70 (0.08) &                       0.06 (0.08) \\
Hashing (Wavelet)           &  \textcolor{darkgreen}{0.002 (0.0)} &                        8156 (54) &           \textcolor{blue}{1.0 (0)} &       \textcolor{blue}{0.498 (0.017)} &   \textcolor{blue}{0.24 (0.19)} &                         0.75 (0.11) &                       0.06 (0.05) \\
Euclidean                   &  \textcolor{darkgreen}{0.002 (0.0)} &                         794 (40) &           \textcolor{blue}{1.0 (0)} &                         0.495 (0.001) &  \textcolor{darkgreen}{0.0 (0)} &  \textcolor{darkgreen}{0.62 (0.01)} &    \textcolor{darkgreen}{0.0 (0)} \\
Hashing (Perceptual)        &       \textcolor{blue}{0.003 (0.0)} &                        3373 (62) &                         0.99 (0.02) &  \textcolor{darkgreen}{0.390 (0.087)} &                     0.06 (0.08) &       \textcolor{blue}{0.98 (0.02)} &                       0.02 (0.02) \\
Window                      &  \textcolor{darkgreen}{0.002 (0.0)} &                      11546 (172) &                         0.99 (0.01) &                         0.481 (0.008) &                     0.17 (0.18) &                         0.71 (0.03) &     \textcolor{blue}{0.07 (0.07)} \\
Visual Diversity            &  \textcolor{darkgreen}{0.002 (0.0)} &  \textcolor{darkgreen}{739 (24)} &           \textcolor{blue}{1.0 (0)} &                         0.493 (0.002) &                     0.07 (0.14) &                         0.66 (0.07) &                       0.02 (0.04) \\
Hashing (Average)           &  \textcolor{darkgreen}{0.002 (0.0)} &                        1860 (24) &           \textcolor{blue}{1.0 (0)} &                         0.495 (0.001) &  \textcolor{darkgreen}{0.0 (0)} &  \textcolor{darkgreen}{0.62 (0.01)} &    \textcolor{darkgreen}{0.0 (0)} \\
Hashing (Perceptual Simple) &       \textcolor{blue}{0.003 (0.0)} &                        2392 (55) &           \textcolor{blue}{1.0 (0)} &                         0.489 (0.014) &  \textcolor{darkgreen}{0.0 (0)} &                         0.68 (0.10) &    \textcolor{darkgreen}{0.0 (0)} \\
\bottomrule
\end{tabular}

%% file: 997_Mario_Distance_Functions.tex
\begin{tabular}{llllllll}
\toprule
{} &                 Generation Time (s) &                      Train Time (s) &                         Solvability &                Compression Distance &                        A* Diversity &                            Leniency &                       A* Difficulty \\
Distance Function           &                                     &                                     &                                     &                                     &                                     &                                     &                                     \\
\midrule
Hashing (Perceptual)        &  \textcolor{darkgreen}{0.07 (0.01)} &                          12929 (80) &                         0.92 (0.07) &                         0.35 (0.23) &                         0.46 (0.07) &                         0.14 (0.22) &  \textcolor{darkgreen}{0.20 (0.02)} \\
Visual Diversity            &       \textcolor{blue}{0.08 (0.01)} &  \textcolor{darkgreen}{11508 (279)} &       \textcolor{blue}{0.98 (0.02)} &                         0.45 (0.10) &  \textcolor{darkgreen}{0.39 (0.11)} &                         0.17 (0.23) &                         0.24 (0.06) \\
Hashing (Average)           &  \textcolor{darkgreen}{0.07 (0.01)} &                         12273 (325) &                         0.75 (0.18) &       \textcolor{blue}{0.50 (0.06)} &       \textcolor{blue}{0.58 (0.10)} &                         0.23 (0.22) &                         0.33 (0.08) \\
Hashing (Wavelet)           &  \textcolor{darkgreen}{0.07 (0.01)} &       \textcolor{blue}{15898 (305)} &  \textcolor{darkgreen}{0.74 (0.21)} &                         0.44 (0.14) &                         0.56 (0.11) &       \textcolor{blue}{0.41 (0.37)} &       \textcolor{blue}{0.39 (0.24)} \\
Hashing (Perceptual Simple) &       \textcolor{blue}{0.08 (0.01)} &                         12753 (190) &                         0.81 (0.23) &                         0.28 (0.18) &                         0.48 (0.13) &                         0.08 (0.08) &                         0.36 (0.27) \\
Euclidean                   &  \textcolor{darkgreen}{0.07 (0.01)} &                         11689 (648) &                         0.84 (0.28) &  \textcolor{darkgreen}{0.21 (0.18)} &                         0.46 (0.08) &  \textcolor{darkgreen}{0.001 (0.0)} &       \textcolor{blue}{0.39 (0.36)} \\
\bottomrule
\end{tabular}

%% file: gecco.bbl

\begin{thebibliography}{62}


\ifx \showCODEN    \undefined \def \showCODEN     #1{\unskip}     \fi
\ifx \showDOI      \undefined \def \showDOI       #1{#1}\fi
\ifx \showISBNx    \undefined \def \showISBNx     #1{\unskip}     \fi
\ifx \showISBNxiii \undefined \def \showISBNxiii  #1{\unskip}     \fi
\ifx \showISSN     \undefined \def \showISSN      #1{\unskip}     \fi
\ifx \showLCCN     \undefined \def \showLCCN      #1{\unskip}     \fi
\ifx \shownote     \undefined \def \shownote      #1{#1}          \fi
\ifx \showarticletitle \undefined \def \showarticletitle #1{#1}   \fi
\ifx \showURL      \undefined \def \showURL       {\relax}        \fi
\providecommand\bibfield[2]{#2}
\providecommand\bibinfo[2]{#2}
\providecommand\natexlab[1]{#1}
\providecommand\showeprint[2][]{arXiv:#2}

\bibitem[\protect\citeauthoryear{Bellera, Julien, and Hanley}{Bellera
  et~al\mbox{.}}{2010}]%
        {bellera2010normal}
\bibfield{author}{\bibinfo{person}{Carine~A Bellera}, \bibinfo{person}{Marilyse
  Julien}, {and} \bibinfo{person}{James~A Hanley}.}
  \bibinfo{year}{2010}\natexlab{}.
\newblock \showarticletitle{Normal approximations to the distributions of the
  Wilcoxon statistics: accurate to what N? Graphical insights}.
\newblock \bibinfo{journal}{\emph{Journal of Statistics Education}}
  \bibinfo{volume}{18}, \bibinfo{number}{2} (\bibinfo{year}{2010}).
\newblock


\bibitem[\protect\citeauthoryear{Beukman, James, and Cleghorn}{Beukman
  et~al\mbox{.}}{2022}]%
        {beukman_metrics}
\bibfield{author}{\bibinfo{person}{Michael Beukman}, \bibinfo{person}{Steven
  James}, {and} \bibinfo{person}{Christopher~W. Cleghorn}.}
  \bibinfo{year}{2022}\natexlab{}.
\newblock \showarticletitle{Towards Objective Metrics for Procedurally
  Generated Video Game Levels}.
\newblock \bibinfo{journal}{\emph{CoRR}}  \bibinfo{volume}{abs/2201.10334}
  (\bibinfo{year}{2022}).
\newblock
\showeprint[arXiv]{2201.10334}
\urldef\tempurl%
\url{https://arxiv.org/abs/2201.10334}
\showURL{%
\tempurl}


\bibitem[\protect\citeauthoryear{Bonferroni}{Bonferroni}{1936}]%
        {bonferroni1936teoria}
\bibfield{author}{\bibinfo{person}{Carlo Bonferroni}.}
  \bibinfo{year}{1936}\natexlab{}.
\newblock \showarticletitle{Teoria statistica delle classi e calcolo delle
  probabilita}.
\newblock \bibinfo{journal}{\emph{Pubblicazioni del R Istituto Superiore di
  Scienze Economiche e Commericiali di Firenze}}  \bibinfo{volume}{8}
  (\bibinfo{year}{1936}), \bibinfo{pages}{3--62}.
\newblock


\bibitem[\protect\citeauthoryear{Cardamone, Loiacono, and Lanzi}{Cardamone
  et~al\mbox{.}}{2011}]%
        {Cardamone_racing}
\bibfield{author}{\bibinfo{person}{Luigi Cardamone}, \bibinfo{person}{Daniele
  Loiacono}, {and} \bibinfo{person}{Pier~Luca Lanzi}.}
  \bibinfo{year}{2011}\natexlab{}.
\newblock \showarticletitle{Interactive Evolution for the Procedural Generation
  of Tracks in a High-End Racing Game}. In
  \bibinfo{booktitle}{\emph{Proceedings of the 13th Annual Conference on
  Genetic and Evolutionary Computation}} (Dublin, Ireland)
  \emph{(\bibinfo{series}{GECCO '11})}. \bibinfo{publisher}{Association for
  Computing Machinery}, \bibinfo{address}{New York, NY, USA},
  \bibinfo{pages}{395–402}.
\newblock
\showISBNx{9781450305570}
\urldef\tempurl%
\url{https://doi.org/10.1145/2001576.2001631}
\showDOI{\tempurl}


\bibitem[\protect\citeauthoryear{Cobbe, Klimov, Hesse, Kim, and Schulman}{Cobbe
  et~al\mbox{.}}{2019}]%
        {cobbe2019quantifying}
\bibfield{author}{\bibinfo{person}{Karl Cobbe}, \bibinfo{person}{Oleg Klimov},
  \bibinfo{person}{Christopher Hesse}, \bibinfo{person}{Taehoon Kim}, {and}
  \bibinfo{person}{John Schulman}.} \bibinfo{year}{2019}\natexlab{}.
\newblock \showarticletitle{Quantifying Generalization in Reinforcement
  Learning}. In \bibinfo{booktitle}{\emph{Proceedings of the 36th International
  Conference on Machine Learning, {ICML} 2019, 9-15 June 2019, Long Beach,
  California, {USA}}} \emph{(\bibinfo{series}{Proceedings of Machine Learning
  Research}, Vol.~\bibinfo{volume}{97})},
  \bibfield{editor}{\bibinfo{person}{Kamalika Chaudhuri} {and}
  \bibinfo{person}{Ruslan Salakhutdinov}} (Eds.). \bibinfo{publisher}{{PMLR}},
  \bibinfo{pages}{1282--1289}.
\newblock
\urldef\tempurl%
\url{http://proceedings.mlr.press/v97/cobbe19a.html}
\showURL{%
\tempurl}


\bibitem[\protect\citeauthoryear{Cohen}{Cohen}{2013}]%
        {cohen2013statistical}
\bibfield{author}{\bibinfo{person}{Jacob Cohen}.}
  \bibinfo{year}{2013}\natexlab{}.
\newblock \bibinfo{booktitle}{\emph{Statistical power analysis for the
  behavioral sciences}}.
\newblock \bibinfo{publisher}{Academic press}.
\newblock


\bibitem[\protect\citeauthoryear{Cook and Colton}{Cook and Colton}{2011}]%
        {cook_evolve}
\bibfield{author}{\bibinfo{person}{Michael Cook} {and} \bibinfo{person}{Simon
  Colton}.} \bibinfo{year}{2011}\natexlab{}.
\newblock \showarticletitle{Multi-faceted evolution of simple arcade games}. In
  \bibinfo{booktitle}{\emph{2011 {IEEE} Conference on Computational
  Intelligence and Games, {CIG} 2011, Seoul, South Korea, August 31 - September
  3, 2011}}, \bibfield{editor}{\bibinfo{person}{Sung{-}Bae Cho},
  \bibinfo{person}{Simon~M. Lucas}, {and} \bibinfo{person}{Philip Hingston}}
  (Eds.). \bibinfo{publisher}{{IEEE}}, \bibinfo{pages}{289--296}.
\newblock
\urldef\tempurl%
\url{https://doi.org/10.1109/CIG.2011.6032019}
\showDOI{\tempurl}


\bibitem[\protect\citeauthoryear{Creswell, White, Dumoulin, Arulkumaran,
  Sengupta, and Bharath}{Creswell et~al\mbox{.}}{2018}]%
        {creswell2018generative}
\bibfield{author}{\bibinfo{person}{Antonia Creswell}, \bibinfo{person}{Tom
  White}, \bibinfo{person}{Vincent Dumoulin}, \bibinfo{person}{Kai
  Arulkumaran}, \bibinfo{person}{Biswa Sengupta}, {and} \bibinfo{person}{Anil~A
  Bharath}.} \bibinfo{year}{2018}\natexlab{}.
\newblock \showarticletitle{Generative adversarial networks: An overview}.
\newblock \bibinfo{journal}{\emph{IEEE Signal Processing Magazine}}
  \bibinfo{volume}{35}, \bibinfo{number}{1} (\bibinfo{year}{2018}),
  \bibinfo{pages}{53--65}.
\newblock


\bibitem[\protect\citeauthoryear{ESA}{ESA}{2021}]%
        {esa_2022}
\bibfield{author}{\bibinfo{person}{ESA}.} \bibinfo{year}{2021}\natexlab{}.
\newblock \bibinfo{title}{2021 Essential Facts about the video game industry}.
\newblock
\newblock
\urldef\tempurl%
\url{https://www.theesa.com/resource/2021-essential-facts/}
\showURL{%
\tempurl}


\bibitem[\protect\citeauthoryear{Ferreira, Pereira, and Toledo}{Ferreira
  et~al\mbox{.}}{2014}]%
        {ferreira_mario}
\bibfield{author}{\bibinfo{person}{Lucas Ferreira}, \bibinfo{person}{Leonardo
  Pereira}, {and} \bibinfo{person}{Claudio Toledo}.}
  \bibinfo{year}{2014}\natexlab{}.
\newblock \showarticletitle{A Multi-Population Genetic Algorithm for Procedural
  Generation of Levels for Platform Games}. In
  \bibinfo{booktitle}{\emph{Proceedings of the Companion Publication of the
  2014 Annual Conference on Genetic and Evolutionary Computation}} (Vancouver,
  BC, Canada) \emph{(\bibinfo{series}{GECCO Comp '14})}.
  \bibinfo{publisher}{Association for Computing Machinery},
  \bibinfo{address}{New York, NY, USA}, \bibinfo{pages}{45–46}.
\newblock
\showISBNx{9781450328814}
\urldef\tempurl%
\url{https://doi.org/10.1145/2598394.2598489}
\showDOI{\tempurl}


\bibitem[\protect\citeauthoryear{Goldberg}{Goldberg}{1989}]%
        {goldberg1989genetic}
\bibfield{author}{\bibinfo{person}{David~E Goldberg}.}
  \bibinfo{year}{1989}\natexlab{}.
\newblock \bibinfo{booktitle}{\emph{Genetic algorithms in search}}.
\newblock \bibinfo{publisher}{Addison Wesley Publishing Co. Inc.}, Chapter~1.
\newblock


\bibitem[\protect\citeauthoryear{Gomes, Mariano, and Christensen}{Gomes
  et~al\mbox{.}}{2015}]%
        {gomes2015devising}
\bibfield{author}{\bibinfo{person}{Jorge Gomes}, \bibinfo{person}{Pedro
  Mariano}, {and} \bibinfo{person}{Anders~Lyhne Christensen}.}
  \bibinfo{year}{2015}\natexlab{}.
\newblock \showarticletitle{Devising effective novelty search algorithms: A
  comprehensive empirical study}. In \bibinfo{booktitle}{\emph{Proceedings of
  the 2015 Annual Conference on Genetic and Evolutionary Computation}}.
  \bibinfo{pages}{943--950}.
\newblock
\urldef\tempurl%
\url{https://citeseerx.ist.psu.edu/viewdoc/download?doi=10.1.1.717.6684&rep=rep1&type=pdf}
\showURL{%
\tempurl}


\bibitem[\protect\citeauthoryear{Gomes, Urbano, and Christensen}{Gomes
  et~al\mbox{.}}{2013}]%
        {gomes2013evolution}
\bibfield{author}{\bibinfo{person}{Jorge Gomes}, \bibinfo{person}{Paulo
  Urbano}, {and} \bibinfo{person}{Anders~Lyhne Christensen}.}
  \bibinfo{year}{2013}\natexlab{}.
\newblock \showarticletitle{Evolution of swarm robotics systems with novelty
  search}.
\newblock \bibinfo{journal}{\emph{Swarm Intelligence}} \bibinfo{volume}{7},
  \bibinfo{number}{2} (\bibinfo{year}{2013}), \bibinfo{pages}{115--144}.
\newblock


\bibitem[\protect\citeauthoryear{Gravina, Khalifa, Liapis, Togelius, and
  Yannakakis}{Gravina et~al\mbox{.}}{2019}]%
        {quality_diversity_pcg}
\bibfield{author}{\bibinfo{person}{Daniele Gravina}, \bibinfo{person}{Ahmed
  Khalifa}, \bibinfo{person}{Antonios Liapis}, \bibinfo{person}{Julian
  Togelius}, {and} \bibinfo{person}{Georgios~N. Yannakakis}.}
  \bibinfo{year}{2019}\natexlab{}.
\newblock \showarticletitle{Procedural Content Generation through Quality
  Diversity}. In \bibinfo{booktitle}{\emph{{IEEE} Conference on Games, CoG
  2019, London, United Kingdom, August 20-23, 2019}}.
  \bibinfo{publisher}{{IEEE}}, \bibinfo{pages}{1--8}.
\newblock
\urldef\tempurl%
\url{https://doi.org/10.1109/CIG.2019.8848053}
\showDOI{\tempurl}


\bibitem[\protect\citeauthoryear{Hadmi, Puech, Said, and Ouahman}{Hadmi
  et~al\mbox{.}}{2012}]%
        {hadmi2012perceptual}
\bibfield{author}{\bibinfo{person}{Azhar Hadmi}, \bibinfo{person}{William
  Puech}, \bibinfo{person}{Brahim Ait~Es Said}, {and}
  \bibinfo{person}{Abdellah~Ait Ouahman}.} \bibinfo{year}{2012}\natexlab{}.
\newblock \showarticletitle{Perceptual image hashing}.
\newblock In \bibinfo{booktitle}{\emph{Watermarking-Volume 2}}.
  \bibinfo{publisher}{IntechOpen}.
\newblock


\bibitem[\protect\citeauthoryear{Hart, Nilsson, and Raphael}{Hart
  et~al\mbox{.}}{1968}]%
        {astar_method}
\bibfield{author}{\bibinfo{person}{Peter~E. Hart}, \bibinfo{person}{Nils~J.
  Nilsson}, {and} \bibinfo{person}{Bertram Raphael}.}
  \bibinfo{year}{1968}\natexlab{}.
\newblock \showarticletitle{A Formal Basis for the Heuristic Determination of
  Minimum Cost Paths}.
\newblock \bibinfo{journal}{\emph{IEEE Transactions on Systems Science and
  Cybernetics}} \bibinfo{volume}{4}, \bibinfo{number}{2}
  (\bibinfo{year}{1968}), \bibinfo{pages}{100--107}.
\newblock
\urldef\tempurl%
\url{https://doi.org/10.1109/TSSC.1968.300136}
\showDOI{\tempurl}


\bibitem[\protect\citeauthoryear{Hendrikx, Meijer, Van Der~Velden, and
  Iosup}{Hendrikx et~al\mbox{.}}{2013}]%
        {hendrikx2013procedural}
\bibfield{author}{\bibinfo{person}{Mark Hendrikx}, \bibinfo{person}{Sebastiaan
  Meijer}, \bibinfo{person}{Joeri Van Der~Velden}, {and}
  \bibinfo{person}{Alexandru Iosup}.} \bibinfo{year}{2013}\natexlab{}.
\newblock \showarticletitle{Procedural content generation for games: A survey}.
\newblock \bibinfo{journal}{\emph{ACM Transactions on Multimedia Computing,
  Communications, and Applications (TOMM)}} \bibinfo{volume}{9},
  \bibinfo{number}{1} (\bibinfo{year}{2013}), \bibinfo{pages}{1--22}.
\newblock


\bibitem[\protect\citeauthoryear{Hoover, Togelius, and Yannakis}{Hoover
  et~al\mbox{.}}{2015}]%
        {hoover2015composing}
\bibfield{author}{\bibinfo{person}{Amy~K Hoover}, \bibinfo{person}{Julian
  Togelius}, {and} \bibinfo{person}{Georgios~N Yannakis}.}
  \bibinfo{year}{2015}\natexlab{}.
\newblock \showarticletitle{Composing video game levels with music metaphors
  through functional scaffolding}. In \bibinfo{booktitle}{\emph{First
  computational creativity and games workshop. ACC}}.
\newblock


\bibitem[\protect\citeauthoryear{Horn, Dahlskog, Shaker, Smith, and
  Togelius}{Horn et~al\mbox{.}}{2014}]%
        {horn2014comparative}
\bibfield{author}{\bibinfo{person}{Britton Horn}, \bibinfo{person}{Steve
  Dahlskog}, \bibinfo{person}{Noor Shaker}, \bibinfo{person}{Gillian Smith},
  {and} \bibinfo{person}{Julian Togelius}.} \bibinfo{year}{2014}\natexlab{}.
\newblock \showarticletitle{A comparative evaluation of procedural level
  generators in the {Mario AI} framework}. In
  \bibinfo{booktitle}{\emph{Foundations of Digital Games 2014, Ft. Lauderdale,
  Florida, USA (2014)}}. Society for the Advancement of the Science of Digital
  Games, \bibinfo{pages}{1--8}.
\newblock


\bibitem[\protect\citeauthoryear{Jafari and Ansari-Pour}{Jafari and
  Ansari-Pour}{2019}]%
        {jafari2019and}
\bibfield{author}{\bibinfo{person}{Mohieddin Jafari} {and}
  \bibinfo{person}{Naser Ansari-Pour}.} \bibinfo{year}{2019}\natexlab{}.
\newblock \showarticletitle{Why, when and how to adjust your P values?}
\newblock \bibinfo{journal}{\emph{Cell Journal (Yakhteh)}}
  \bibinfo{volume}{20}, \bibinfo{number}{4} (\bibinfo{year}{2019}),
  \bibinfo{pages}{604}.
\newblock


\bibitem[\protect\citeauthoryear{Justesen, Torrado, Bontrager, Khalifa,
  Togelius, and Risi}{Justesen et~al\mbox{.}}{2018}]%
        {using_pcg_to_train_rl}
\bibfield{author}{\bibinfo{person}{Niels Justesen},
  \bibinfo{person}{Ruben~Rodriguez Torrado}, \bibinfo{person}{Philip
  Bontrager}, \bibinfo{person}{Ahmed Khalifa}, \bibinfo{person}{Julian
  Togelius}, {and} \bibinfo{person}{Sebastian Risi}.}
  \bibinfo{year}{2018}\natexlab{}.
\newblock \showarticletitle{Illuminating generalization in deep reinforcement
  learning through procedural level generation}.
\newblock \bibinfo{journal}{\emph{arXiv preprint arXiv:1806.10729}}
  (\bibinfo{year}{2018}).
\newblock


\bibitem[\protect\citeauthoryear{Kerssemakers, Tuxen, Togelius, and
  Yannakakis}{Kerssemakers et~al\mbox{.}}{2012}]%
        {pplgg}
\bibfield{author}{\bibinfo{person}{Manuel Kerssemakers}, \bibinfo{person}{Jeppe
  Tuxen}, \bibinfo{person}{Julian Togelius}, {and} \bibinfo{person}{Georgios
  Yannakakis}.} \bibinfo{year}{2012}\natexlab{}.
\newblock \showarticletitle{A procedural procedural level generator generator}.
  In \bibinfo{booktitle}{\emph{2012 IEEE Conference on Computational
  Intelligence and Games, CIG 2012}}. \bibinfo{pages}{335--341}.
\newblock
\showISBNx{978-1-4673-1193-9}
\urldef\tempurl%
\url{https://doi.org/10.1109/CIG.2012.6374174}
\showDOI{\tempurl}


\bibitem[\protect\citeauthoryear{Khalifa, Bontrager, Earle, and
  Togelius}{Khalifa et~al\mbox{.}}{2020}]%
        {pcgrl}
\bibfield{author}{\bibinfo{person}{Ahmed Khalifa}, \bibinfo{person}{Philip
  Bontrager}, \bibinfo{person}{Sam Earle}, {and} \bibinfo{person}{Julian
  Togelius}.} \bibinfo{year}{2020}\natexlab{}.
\newblock \showarticletitle{{PCGRL}: Procedural content generation via
  reinforcement learning}. In \bibinfo{booktitle}{\emph{Proceedings of the AAAI
  Conference on Artificial Intelligence and Interactive Digital
  Entertainment}}, Vol.~\bibinfo{volume}{16}. \bibinfo{pages}{95--101}.
\newblock


\bibitem[\protect\citeauthoryear{Kruskal and Wallis}{Kruskal and
  Wallis}{1952}]%
        {kruskal1952use}
\bibfield{author}{\bibinfo{person}{William~H Kruskal} {and}
  \bibinfo{person}{W~Allen Wallis}.} \bibinfo{year}{1952}\natexlab{}.
\newblock \showarticletitle{Use of ranks in one-criterion variance analysis}.
\newblock \bibinfo{journal}{\emph{Journal of the American statistical
  Association}} \bibinfo{volume}{47}, \bibinfo{number}{260}
  (\bibinfo{year}{1952}), \bibinfo{pages}{583--621}.
\newblock


\bibitem[\protect\citeauthoryear{Lehman and Stanley}{Lehman and
  Stanley}{2011a}]%
        {novelty_search}
\bibfield{author}{\bibinfo{person}{Joel Lehman} {and} \bibinfo{person}{Kenneth
  Stanley}.} \bibinfo{year}{2011}\natexlab{a}.
\newblock \showarticletitle{Abandoning Objectives: Evolution Through the Search
  for Novelty Alone}.
\newblock \bibinfo{journal}{\emph{Evolutionary computation}}
  \bibinfo{volume}{19} (\bibinfo{date}{06} \bibinfo{year}{2011}),
  \bibinfo{pages}{189--223}.
\newblock
\urldef\tempurl%
\url{https://doi.org/10.1162/EVCO_a_00025}
\showDOI{\tempurl}


\bibitem[\protect\citeauthoryear{Lehman and Stanley}{Lehman and
  Stanley}{2011b}]%
        {lehman2011evolving}
\bibfield{author}{\bibinfo{person}{Joel Lehman} {and}
  \bibinfo{person}{Kenneth~O Stanley}.} \bibinfo{year}{2011}\natexlab{b}.
\newblock \showarticletitle{Evolving a diversity of virtual creatures through
  novelty search and local competition}. In
  \bibinfo{booktitle}{\emph{Proceedings of the 13th annual conference on
  Genetic and evolutionary computation}}. \bibinfo{pages}{211--218}.
\newblock


\bibitem[\protect\citeauthoryear{Li, Chen, Li, Ma, and Vit{\'a}nyi}{Li
  et~al\mbox{.}}{2004}]%
        {li2004similarity}
\bibfield{author}{\bibinfo{person}{Ming Li}, \bibinfo{person}{Xin Chen},
  \bibinfo{person}{Xin Li}, \bibinfo{person}{Bin Ma}, {and}
  \bibinfo{person}{Paul~MB Vit{\'a}nyi}.} \bibinfo{year}{2004}\natexlab{}.
\newblock \showarticletitle{The similarity metric}.
\newblock \bibinfo{journal}{\emph{IEEE Transactions on Information Theory}}
  \bibinfo{volume}{50}, \bibinfo{number}{12} (\bibinfo{year}{2004}),
  \bibinfo{pages}{3250--3264}.
\newblock


\bibitem[\protect\citeauthoryear{Liapis, Yannakakis, and Togelius}{Liapis
  et~al\mbox{.}}{2013}]%
        {liapis_2013c}
\bibfield{author}{\bibinfo{person}{Antonios Liapis}, \bibinfo{person}{Georgios
  Yannakakis}, {and} \bibinfo{person}{Julian Togelius}.}
  \bibinfo{year}{2013}\natexlab{}.
\newblock \showarticletitle{Enhancements to constrained novelty search:
  two-population novelty search for generating game content}. In
  \bibinfo{booktitle}{\emph{GECCO 2013 - Proceedings of the 2013 Genetic and
  Evolutionary Computation Conference}}. \bibinfo{pages}{343--350}.
\newblock
\urldef\tempurl%
\url{https://doi.org/10.1145/2463372.2463416}
\showDOI{\tempurl}


\bibitem[\protect\citeauthoryear{Liapis, Yannakakis, and Togelius}{Liapis
  et~al\mbox{.}}{2015}]%
        {constrained_novelty}
\bibfield{author}{\bibinfo{person}{Antonios Liapis},
  \bibinfo{person}{Georgios~N. Yannakakis}, {and} \bibinfo{person}{Julian
  Togelius}.} \bibinfo{year}{2015}\natexlab{}.
\newblock \showarticletitle{Constrained Novelty Search: A Study on Game Content
  Generation}.
\newblock \bibinfo{journal}{\emph{Evolutionary computation}}
  \bibinfo{volume}{23}, \bibinfo{number}{1} (\bibinfo{date}{March}
  \bibinfo{year}{2015}), \bibinfo{pages}{101–129}.
\newblock
\showISSN{1063-6560}
\urldef\tempurl%
\url{https://doi.org/10.1162/EVCO_a_00123}
\showDOI{\tempurl}


\bibitem[\protect\citeauthoryear{Liu, Snodgrass, Khalifa, Risi, Yannakakis, and
  Togelius}{Liu et~al\mbox{.}}{2020}]%
        {liu2020deep}
\bibfield{author}{\bibinfo{person}{Jialin Liu}, \bibinfo{person}{Sam
  Snodgrass}, \bibinfo{person}{Ahmed Khalifa}, \bibinfo{person}{Sebastian
  Risi}, \bibinfo{person}{Georgios~N Yannakakis}, {and} \bibinfo{person}{Julian
  Togelius}.} \bibinfo{year}{2020}\natexlab{}.
\newblock \showarticletitle{Deep learning for procedural content generation}.
\newblock \bibinfo{journal}{\emph{Neural Computing and Applications}}
  (\bibinfo{year}{2020}), \bibinfo{pages}{1--19}.
\newblock


\bibitem[\protect\citeauthoryear{Liu, Yao, Zhao, and Higuchi}{Liu
  et~al\mbox{.}}{2001}]%
        {liu2001scaling}
\bibfield{author}{\bibinfo{person}{Yong Liu}, \bibinfo{person}{Xin Yao},
  \bibinfo{person}{Qiangfu Zhao}, {and} \bibinfo{person}{Tetsuya Higuchi}.}
  \bibinfo{year}{2001}\natexlab{}.
\newblock \showarticletitle{Scaling up fast evolutionary programming with
  cooperative coevolution}. In \bibinfo{booktitle}{\emph{Proceedings of the
  2001 Congress on Evolutionary Computation (IEEE Cat. No. 01TH8546)}},
  Vol.~\bibinfo{volume}{2}. Ieee, \bibinfo{pages}{1101--1108}.
\newblock


\bibitem[\protect\citeauthoryear{Lowell, Grabkovsky, and Birger}{Lowell
  et~al\mbox{.}}{2011}]%
        {hyperneat_vs_neat}
\bibfield{author}{\bibinfo{person}{Jessica Lowell}, \bibinfo{person}{Sergey
  Grabkovsky}, {and} \bibinfo{person}{Kir Birger}.}
  \bibinfo{year}{2011}\natexlab{}.
\newblock \showarticletitle{Comparison of {NEAT} and HyperNEAT Performance on a
  Strategic Decision-Making Problem}. In \bibinfo{booktitle}{\emph{Fifth
  International Conference on Genetic and Evolutionary Computing, {ICGEC} 2011,
  Kinmen, Taiwan / Xiamen, China, August 29 - September 1, 2011}},
  \bibfield{editor}{\bibinfo{person}{Junzo Watada}, \bibinfo{person}{Pau{-}Choo
  Chung}, \bibinfo{person}{Jim{-}Min Lin}, \bibinfo{person}{Chin{-}Shiuh
  Shieh}, {and} \bibinfo{person}{Jeng{-}Shyang Pan}} (Eds.).
  \bibinfo{publisher}{{IEEE} Computer Society}, \bibinfo{pages}{102--105}.
\newblock
\urldef\tempurl%
\url{https://doi.org/10.1109/ICGEC.2011.33}
\showDOI{\tempurl}


\bibitem[\protect\citeauthoryear{Mann and Whitney}{Mann and Whitney}{1947}]%
        {mann_whitney_test}
\bibfield{author}{\bibinfo{person}{H.~B. Mann} {and} \bibinfo{person}{D.~R.
  Whitney}.} \bibinfo{year}{1947}\natexlab{}.
\newblock \showarticletitle{{On a Test of Whether one of Two Random Variables
  is Stochastically Larger than the Other}}.
\newblock \bibinfo{journal}{\emph{The Annals of Mathematical Statistics}}
  \bibinfo{volume}{18}, \bibinfo{number}{1} (\bibinfo{year}{1947}),
  \bibinfo{pages}{50 -- 60}.
\newblock
\urldef\tempurl%
\url{https://doi.org/10.1214/aoms/1177730491}
\showDOI{\tempurl}


\bibitem[\protect\citeauthoryear{Maung and Crawfis}{Maung and Crawfis}{2015}]%
        {maung2015applying_wangs}
\bibfield{author}{\bibinfo{person}{David Maung} {and} \bibinfo{person}{Roger
  Crawfis}.} \bibinfo{year}{2015}\natexlab{}.
\newblock \showarticletitle{Applying formal picture languages to procedural
  content generation}. In \bibinfo{booktitle}{\emph{2015 Computer Games: AI,
  Animation, Mobile, Multimedia, Educational and Serious Games (CGAMES)}}.
  IEEE, \bibinfo{pages}{58--64}.
\newblock


\bibitem[\protect\citeauthoryear{McIntyre, Kallada, Miguel, and Feher~de
  Silva}{McIntyre et~al\mbox{.}}{[n.\,d.]}]%
        {neat_python_cite}
\bibfield{author}{\bibinfo{person}{Alan McIntyre}, \bibinfo{person}{Matt
  Kallada}, \bibinfo{person}{Cesar~G. Miguel}, {and} \bibinfo{person}{Carolina
  Feher~de Silva}.} \bibinfo{year}{[n.\,d.]}\natexlab{}.
\newblock \bibinfo{booktitle}{\emph{{neat-python}}}.
\newblock
\urldef\tempurl%
\url{https://github.com/CodeReclaimers/neat-python}
\showURL{%
\tempurl}


\bibitem[\protect\citeauthoryear{Monga and Evans}{Monga and Evans}{2006}]%
        {monga2006perceptual}
\bibfield{author}{\bibinfo{person}{Vishal Monga} {and} \bibinfo{person}{Brian~L
  Evans}.} \bibinfo{year}{2006}\natexlab{}.
\newblock \showarticletitle{Perceptual image hashing via feature points:
  performance evaluation and tradeoffs}.
\newblock \bibinfo{journal}{\emph{IEEE transactions on Image Processing}}
  \bibinfo{volume}{15}, \bibinfo{number}{11} (\bibinfo{year}{2006}),
  \bibinfo{pages}{3452--3465}.
\newblock


\bibitem[\protect\citeauthoryear{Mouret and Clune}{Mouret and Clune}{2015}]%
        {map_elites}
\bibfield{author}{\bibinfo{person}{Jean{-}Baptiste Mouret} {and}
  \bibinfo{person}{Jeff Clune}.} \bibinfo{year}{2015}\natexlab{}.
\newblock \showarticletitle{Illuminating search spaces by mapping elites}.
\newblock \bibinfo{journal}{\emph{CoRR}}  \bibinfo{volume}{abs/1504.04909}
  (\bibinfo{year}{2015}).
\newblock
\showeprint[arXiv]{1504.04909}
\urldef\tempurl%
\url{http://arxiv.org/abs/1504.04909}
\showURL{%
\tempurl}


\bibitem[\protect\citeauthoryear{Pinelle, Wong, and Stach}{Pinelle
  et~al\mbox{.}}{2008}]%
        {pinelle2008heuristic}
\bibfield{author}{\bibinfo{person}{David Pinelle}, \bibinfo{person}{Nelson
  Wong}, {and} \bibinfo{person}{Tadeusz Stach}.}
  \bibinfo{year}{2008}\natexlab{}.
\newblock \showarticletitle{Heuristic evaluation for games: usability
  principles for video game design}. In \bibinfo{booktitle}{\emph{Proceedings
  of the SIGCHI conference on human factors in computing systems}}.
  \bibinfo{pages}{1453--1462}.
\newblock


\bibitem[\protect\citeauthoryear{Pretorius, Biljon, van Niekerk, Eloff,
  Reynard, James, Rosman, Kamper, and Kroon}{Pretorius et~al\mbox{.}}{2019}]%
        {dropout_statistical_analysis}
\bibfield{author}{\bibinfo{person}{Arnu Pretorius}, \bibinfo{person}{Elan~Van
  Biljon}, \bibinfo{person}{Benjamin van Niekerk}, \bibinfo{person}{Ryan
  Eloff}, \bibinfo{person}{Matthew Reynard}, \bibinfo{person}{Steven~D. James},
  \bibinfo{person}{Benjamin Rosman}, \bibinfo{person}{Herman Kamper}, {and}
  \bibinfo{person}{Steve Kroon}.} \bibinfo{year}{2019}\natexlab{}.
\newblock \showarticletitle{If dropout limits trainable depth, does critical
  initialisation still matter? {A} large-scale statistical analysis on ReLU
  networks}.
\newblock \bibinfo{journal}{\emph{CoRR}}  \bibinfo{volume}{abs/1910.05725}
  (\bibinfo{year}{2019}).
\newblock
\showeprint[arXiv]{1910.05725}
\urldef\tempurl%
\url{http://arxiv.org/abs/1910.05725}
\showURL{%
\tempurl}


\bibitem[\protect\citeauthoryear{Preuss, Liapis, and Togelius}{Preuss
  et~al\mbox{.}}{2014}]%
        {searching_for_good_and_diverse_game_levels}
\bibfield{author}{\bibinfo{person}{Mike Preuss}, \bibinfo{person}{Antonios
  Liapis}, {and} \bibinfo{person}{Julian Togelius}.}
  \bibinfo{year}{2014}\natexlab{}.
\newblock \showarticletitle{Searching for good and diverse game levels}. In
  \bibinfo{booktitle}{\emph{2014 {IEEE} Conference on Computational
  Intelligence and Games, {CIG} 2014, Dortmund, Germany, August 26-29, 2014}}.
  \bibinfo{publisher}{{IEEE}}, \bibinfo{pages}{1--8}.
\newblock
\urldef\tempurl%
\url{https://doi.org/10.1109/CIG.2014.6932908}
\showDOI{\tempurl}


\bibitem[\protect\citeauthoryear{Schrum, Volz, and Risi}{Schrum
  et~al\mbox{.}}{2020}]%
        {CPPN2GAN}
\bibfield{author}{\bibinfo{person}{Jacob Schrum}, \bibinfo{person}{Vanessa
  Volz}, {and} \bibinfo{person}{Sebastian Risi}.}
  \bibinfo{year}{2020}\natexlab{}.
\newblock \showarticletitle{{CPPN2GAN}: Combining Compositional Pattern
  Producing Networks and GANs for Large-Scale Pattern Generation}. In
  \bibinfo{booktitle}{\emph{Proceedings of the 2020 Genetic and Evolutionary
  Computation Conference}} (Canc\'{u}n, Mexico) \emph{(\bibinfo{series}{GECCO
  '20})}. \bibinfo{publisher}{Association for Computing Machinery},
  \bibinfo{address}{New York, NY, USA}, \bibinfo{pages}{139–147}.
\newblock
\showISBNx{9781450371285}
\urldef\tempurl%
\url{https://doi.org/10.1145/3377930.3389822}
\showDOI{\tempurl}


\bibitem[\protect\citeauthoryear{Shaker, Nicolau, Yannakakis, Togelius, and
  O'neill}{Shaker et~al\mbox{.}}{2012}]%
        {shaker2012evolving}
\bibfield{author}{\bibinfo{person}{Noor Shaker}, \bibinfo{person}{Miguel
  Nicolau}, \bibinfo{person}{Georgios~N Yannakakis}, \bibinfo{person}{Julian
  Togelius}, {and} \bibinfo{person}{Michael O'neill}.}
  \bibinfo{year}{2012}\natexlab{}.
\newblock \showarticletitle{Evolving levels for super mario bros using
  grammatical evolution}. In \bibinfo{booktitle}{\emph{2012 IEEE Conference on
  Computational Intelligence and Games (CIG)}}. IEEE,
  \bibinfo{pages}{304--311}.
\newblock


\bibitem[\protect\citeauthoryear{Shannon}{Shannon}{1948}]%
        {shannon1948mathematical}
\bibfield{author}{\bibinfo{person}{Claude~E Shannon}.}
  \bibinfo{year}{1948}\natexlab{}.
\newblock \showarticletitle{A mathematical theory of communication}.
\newblock \bibinfo{journal}{\emph{The Bell system technical journal}}
  \bibinfo{volume}{27}, \bibinfo{number}{3} (\bibinfo{year}{1948}),
  \bibinfo{pages}{379--423}.
\newblock


\bibitem[\protect\citeauthoryear{Shaphiro and Wilk}{Shaphiro and Wilk}{1965}]%
        {shaphiro1965analysis}
\bibfield{author}{\bibinfo{person}{S Shaphiro} {and} \bibinfo{person}{M Wilk}.}
  \bibinfo{year}{1965}\natexlab{}.
\newblock \showarticletitle{An analysis of variance test for normality}.
\newblock \bibinfo{journal}{\emph{Biometrika}} \bibinfo{volume}{52},
  \bibinfo{number}{3} (\bibinfo{year}{1965}), \bibinfo{pages}{591--611}.
\newblock


\bibitem[\protect\citeauthoryear{Shu, Liu, and Yannakakis}{Shu
  et~al\mbox{.}}{2021}]%
        {edrl}
\bibfield{author}{\bibinfo{person}{Tianye Shu}, \bibinfo{person}{Jialin Liu},
  {and} \bibinfo{person}{Georgios~N. Yannakakis}.}
  \bibinfo{year}{2021}\natexlab{}.
\newblock \showarticletitle{Experience-Driven PCG via Reinforcement Learning: A
  Super Mario Bros Study}. In \bibinfo{booktitle}{\emph{2021 IEEE Conference on
  Games (CoG)}}. IEEE.
\newblock


\bibitem[\protect\citeauthoryear{Smith, Whitehead, Mateas, Treanor, March, and
  Cha}{Smith et~al\mbox{.}}{2010}]%
        {smith2010launchpad_leniency}
\bibfield{author}{\bibinfo{person}{Gillian Smith}, \bibinfo{person}{Jim
  Whitehead}, \bibinfo{person}{Michael Mateas}, \bibinfo{person}{Mike Treanor},
  \bibinfo{person}{Jameka March}, {and} \bibinfo{person}{Mee Cha}.}
  \bibinfo{year}{2010}\natexlab{}.
\newblock \showarticletitle{Launchpad: A rhythm-based level generator for 2-d
  platformers}.
\newblock \bibinfo{journal}{\emph{IEEE Transactions on computational
  intelligence and AI in games}} \bibinfo{volume}{3}, \bibinfo{number}{1}
  (\bibinfo{year}{2010}), \bibinfo{pages}{1--16}.
\newblock


\bibitem[\protect\citeauthoryear{Spears et~al\mbox{.}}{Spears
  et~al\mbox{.}}{1995}]%
        {spears1995adapting}
\bibfield{author}{\bibinfo{person}{William~M Spears} {et~al\mbox{.}}}
  \bibinfo{year}{1995}\natexlab{}.
\newblock \showarticletitle{Adapting crossover in evolutionary algorithms.}. In
  \bibinfo{booktitle}{\emph{Evolutionary programming}}.
  \bibinfo{pages}{367--384}.
\newblock


\bibitem[\protect\citeauthoryear{Stanley}{Stanley}{2007}]%
        {cppn}
\bibfield{author}{\bibinfo{person}{Kenneth~O. Stanley}.}
  \bibinfo{year}{2007}\natexlab{}.
\newblock \showarticletitle{Compositional Pattern Producing Networks: A Novel
  Abstraction of Development}.
\newblock \bibinfo{journal}{\emph{Genetic Programming and Evolvable Machines}}
  \bibinfo{volume}{8}, \bibinfo{number}{2} (\bibinfo{date}{June}
  \bibinfo{year}{2007}), \bibinfo{pages}{131–162}.
\newblock
\showISSN{1389-2576}
\urldef\tempurl%
\url{https://doi.org/10.1007/s10710-007-9028-8}
\showDOI{\tempurl}


\bibitem[\protect\citeauthoryear{Stanley, D'Ambrosio, and Gauci}{Stanley
  et~al\mbox{.}}{2009}]%
        {hyperneat}
\bibfield{author}{\bibinfo{person}{Kenneth~O. Stanley},
  \bibinfo{person}{David~B. D'Ambrosio}, {and} \bibinfo{person}{Jason Gauci}.}
  \bibinfo{year}{2009}\natexlab{}.
\newblock \showarticletitle{A Hypercube-Based Encoding for Evolving Large-Scale
  Neural Networks}.
\newblock \bibinfo{journal}{\emph{Artif. Life}} \bibinfo{volume}{15},
  \bibinfo{number}{2} (\bibinfo{year}{2009}), \bibinfo{pages}{185--212}.
\newblock
\urldef\tempurl%
\url{https://doi.org/10.1162/artl.2009.15.2.15202}
\showDOI{\tempurl}


\bibitem[\protect\citeauthoryear{Stanley and Miikkulainen}{Stanley and
  Miikkulainen}{2002}]%
        {neat}
\bibfield{author}{\bibinfo{person}{Kenneth~O. Stanley} {and}
  \bibinfo{person}{Risto Miikkulainen}.} \bibinfo{year}{2002}\natexlab{}.
\newblock \showarticletitle{Evolving Neural Networks through Augmenting
  Topologies}.
\newblock \bibinfo{journal}{\emph{Evolutionary Computation}}
  \bibinfo{volume}{10}, \bibinfo{number}{2} (\bibinfo{date}{June}
  \bibinfo{year}{2002}), \bibinfo{pages}{99–127}.
\newblock
\showISSN{1063-6560}
\urldef\tempurl%
\url{https://doi.org/10.1162/106365602320169811}
\showDOI{\tempurl}


\bibitem[\protect\citeauthoryear{Steam}{Steam}{2021}]%
        {steam_hw_survey}
\bibfield{author}{\bibinfo{person}{Steam}.} \bibinfo{year}{2021}\natexlab{}.
\newblock \bibinfo{title}{Steam Hardware \& Software Survey: September 2021}.
\newblock
  \bibinfo{howpublished}{\url{https://store.steampowered.com/hwsurvey/}}.
\newblock
\newblock
\shownote{[Online; accessed 25-October-2021]}.


\bibitem[\protect\citeauthoryear{Sturtevant and Ota}{Sturtevant and
  Ota}{2018}]%
        {sturtevant2018exhaustive}
\bibfield{author}{\bibinfo{person}{Nathan Sturtevant} {and}
  \bibinfo{person}{Matheus Ota}.} \bibinfo{year}{2018}\natexlab{}.
\newblock \showarticletitle{Exhaustive and semi-exhaustive procedural content
  generation}. In \bibinfo{booktitle}{\emph{Proceedings of the AAAI Conference
  on Artificial Intelligence and Interactive Digital Entertainment}},
  Vol.~\bibinfo{volume}{14}.
\newblock


\bibitem[\protect\citeauthoryear{Summerville, Snodgrass, Guzdial, Holmg{\aa}rd,
  Hoover, Isaksen, Nealen, and Togelius}{Summerville et~al\mbox{.}}{2018}]%
        {summerville2018procedural_pcgml}
\bibfield{author}{\bibinfo{person}{Adam Summerville}, \bibinfo{person}{Sam
  Snodgrass}, \bibinfo{person}{Matthew Guzdial}, \bibinfo{person}{Christoffer
  Holmg{\aa}rd}, \bibinfo{person}{Amy~K Hoover}, \bibinfo{person}{Aaron
  Isaksen}, \bibinfo{person}{Andy Nealen}, {and} \bibinfo{person}{Julian
  Togelius}.} \bibinfo{year}{2018}\natexlab{}.
\newblock \showarticletitle{{Procedural Content Generation via Machine Learning
  (PCGML)}}.
\newblock \bibinfo{journal}{\emph{IEEE Transactions on Games}}
  \bibinfo{volume}{10}, \bibinfo{number}{3} (\bibinfo{year}{2018}),
  \bibinfo{pages}{257--270}.
\newblock
\urldef\tempurl%
\url{https://doi.org/10.1109/TG.2018.2846639}
\showDOI{\tempurl}


\bibitem[\protect\citeauthoryear{Sutton and Barto}{Sutton and Barto}{2018}]%
        {sutton1998_rlbook}
\bibfield{author}{\bibinfo{person}{Richard~S. Sutton} {and}
  \bibinfo{person}{Andrew~G. Barto}.} \bibinfo{year}{2018}\natexlab{}.
\newblock \bibinfo{booktitle}{\emph{Reinforcement Learning: An Introduction}
  (\bibinfo{edition}{second} ed.)}.
\newblock \bibinfo{publisher}{The MIT Press}.
\newblock
\urldef\tempurl%
\url{http://incompleteideas.net/book/the-book-2nd.html}
\showURL{%
\tempurl}


\bibitem[\protect\citeauthoryear{Syswerda}{Syswerda}{1993}]%
        {syswerda1993simulated}
\bibfield{author}{\bibinfo{person}{Gilbert Syswerda}.}
  \bibinfo{year}{1993}\natexlab{}.
\newblock \showarticletitle{Simulated crossover in genetic algorithms}.
\newblock In \bibinfo{booktitle}{\emph{Foundations of genetic algorithms}}.
  Vol.~\bibinfo{volume}{2}. \bibinfo{publisher}{Elsevier},
  \bibinfo{pages}{239--255}.
\newblock


\bibitem[\protect\citeauthoryear{Tamaki, Kita, and Kobayashi}{Tamaki
  et~al\mbox{.}}{1996}]%
        {tamaki1996multi}
\bibfield{author}{\bibinfo{person}{Hisashi Tamaki}, \bibinfo{person}{Hajime
  Kita}, {and} \bibinfo{person}{Shigenobu Kobayashi}.}
  \bibinfo{year}{1996}\natexlab{}.
\newblock \showarticletitle{Multi-Objective Optimization by Genetic Algorithms:
  {A} Review}. In \bibinfo{booktitle}{\emph{Proceedings of 1996 {IEEE}
  International Conference on Evolutionary Computation, Nayoya University,
  Japan, May 20-22, 1996}}, \bibfield{editor}{\bibinfo{person}{Toshio Fukuda}
  {and} \bibinfo{person}{Takeshi Furuhashi}} (Eds.).
  \bibinfo{publisher}{{IEEE}}, \bibinfo{address}{Japan},
  \bibinfo{pages}{517--522}.
\newblock
\urldef\tempurl%
\url{https://doi.org/10.1109/ICEC.1996.542653}
\showDOI{\tempurl}


\bibitem[\protect\citeauthoryear{Togelius, Champandard, Lanzi, Mateas, Paiva,
  Preuss, and Stanley}{Togelius et~al\mbox{.}}{2013a}]%
        {togelius2013procedural_challenges_generic}
\bibfield{author}{\bibinfo{person}{Julian Togelius}, \bibinfo{person}{Alex~J.
  Champandard}, \bibinfo{person}{Pier~Luca Lanzi}, \bibinfo{person}{Michael
  Mateas}, \bibinfo{person}{Ana Paiva}, \bibinfo{person}{Mike Preuss}, {and}
  \bibinfo{person}{Kenneth~O. Stanley}.} \bibinfo{year}{2013}\natexlab{a}.
\newblock \showarticletitle{Procedural Content Generation: Goals, Challenges
  and Actionable Steps}.
\newblock In \bibinfo{booktitle}{\emph{Artificial and Computational
  Intelligence in Games}}, \bibfield{editor}{\bibinfo{person}{Simon~M. Lucas},
  \bibinfo{person}{Michael Mateas}, \bibinfo{person}{Mike Preuss},
  \bibinfo{person}{Pieter Spronck}, {and} \bibinfo{person}{Julian Togelius}}
  (Eds.). \bibinfo{series}{Dagstuhl Follow-Ups}, Vol.~\bibinfo{volume}{6}.
  \bibinfo{publisher}{Schloss Dagstuhl - Leibniz-Zentrum f{\"{u}}r Informatik},
  \bibinfo{pages}{61--75}.
\newblock
\urldef\tempurl%
\url{https://doi.org/10.4230/DFU.Vol6.12191.61}
\showDOI{\tempurl}


\bibitem[\protect\citeauthoryear{Togelius, Karakovskiy, and
  Baumgarten}{Togelius et~al\mbox{.}}{2010}]%
        {togelius20102009_with_baumgarten}
\bibfield{author}{\bibinfo{person}{Julian Togelius}, \bibinfo{person}{Sergey
  Karakovskiy}, {and} \bibinfo{person}{Robin Baumgarten}.}
  \bibinfo{year}{2010}\natexlab{}.
\newblock \showarticletitle{The 2009 mario ai competition}. In
  \bibinfo{booktitle}{\emph{IEEE Congress on Evolutionary Computation}}. IEEE,
  \bibinfo{pages}{1--8}.
\newblock


\bibitem[\protect\citeauthoryear{Togelius, Shaker, Karakovskiy, and
  Yannakakis}{Togelius et~al\mbox{.}}{2013b}]%
        {togelius2013mario}
\bibfield{author}{\bibinfo{person}{Julian Togelius}, \bibinfo{person}{Noor
  Shaker}, \bibinfo{person}{Sergey Karakovskiy}, {and}
  \bibinfo{person}{Georgios~N Yannakakis}.} \bibinfo{year}{2013}\natexlab{b}.
\newblock \showarticletitle{The mario ai championship 2009-2012}.
\newblock \bibinfo{journal}{\emph{AI Magazine}} \bibinfo{volume}{34},
  \bibinfo{number}{3} (\bibinfo{year}{2013}), \bibinfo{pages}{89--92}.
\newblock


\bibitem[\protect\citeauthoryear{Togelius, Yannakakis, Stanley, and
  Browne}{Togelius et~al\mbox{.}}{2011}]%
        {togelius2011searchbased}
\bibfield{author}{\bibinfo{person}{Julian Togelius},
  \bibinfo{person}{Georgios~N Yannakakis}, \bibinfo{person}{Kenneth~O Stanley},
  {and} \bibinfo{person}{Cameron Browne}.} \bibinfo{year}{2011}\natexlab{}.
\newblock \showarticletitle{Search-based procedural content generation: A
  taxonomy and survey}.
\newblock \bibinfo{journal}{\emph{IEEE Transactions on Computational
  Intelligence and AI in Games}} \bibinfo{volume}{3}, \bibinfo{number}{3}
  (\bibinfo{year}{2011}), \bibinfo{pages}{172--186}.
\newblock


\bibitem[\protect\citeauthoryear{Volz, Schrum, Liu, Lucas, Smith, and
  Risi}{Volz et~al\mbox{.}}{2018}]%
        {volz2018evolving}
\bibfield{author}{\bibinfo{person}{Vanessa Volz}, \bibinfo{person}{Jacob
  Schrum}, \bibinfo{person}{Jialin Liu}, \bibinfo{person}{Simon~M Lucas},
  \bibinfo{person}{Adam Smith}, {and} \bibinfo{person}{Sebastian Risi}.}
  \bibinfo{year}{2018}\natexlab{}.
\newblock \showarticletitle{Evolving mario levels in the latent space of a deep
  convolutional generative adversarial network}. In
  \bibinfo{booktitle}{\emph{Proceedings of the Genetic and Evolutionary
  Computation Conference}}. \bibinfo{pages}{221--228}.
\newblock


\bibitem[\protect\citeauthoryear{Yao and Liu}{Yao and Liu}{1998}]%
        {yao1998scaling}
\bibfield{author}{\bibinfo{person}{Xin Yao} {and} \bibinfo{person}{Yong Liu}.}
  \bibinfo{year}{1998}\natexlab{}.
\newblock \showarticletitle{Scaling up evolutionary programming algorithms}. In
  \bibinfo{booktitle}{\emph{International Conference on Evolutionary
  Programming}}. Springer, \bibinfo{pages}{103--112}.
\newblock


\end{thebibliography}
